%% file: main.tex
\documentclass[]{fairmeta}

\usepackage{microtype}
\usepackage{graphicx}
\usepackage{subcaption}
\usepackage{csquotes}
\usepackage{afterpage}

\usepackage{amsmath}
\usepackage{amssymb}
\usepackage{mathtools}
\usepackage{amsthm}
\usepackage{xspace}
\usepackage{colortbl}
\usepackage{multirow}
\usepackage{enumitem}

\usepackage{nicefrac}

\usepackage{algorithm}
\usepackage{algorithmic}
\usepackage{amsmath}
\usepackage{amssymb} 
\usepackage{booktabs}
\usepackage{caption}
\usepackage{enumitem}
\usepackage{graphicx}
\usepackage{url}
\usepackage[table]{xcolor}
\usepackage{multirow}

\usepackage{arydshln}

\lstdefinestyle{mystyle}{
    backgroundcolor=\color{white},   
    commentstyle=\color{green},
    keywordstyle=\color{blue},
    stringstyle=\color{red},
    basicstyle=\ttfamily\footnotesize,
    breakatwhitespace=false,         
    breaklines=true,                 
    captionpos=b,                    
    keepspaces=false,                 
    showspaces=false,                
    showstringspaces=false,
    showtabs=false,                  
    tabsize=2,
    frame=single,
    xleftmargin=0pt  
}
\lstdefinestyle{rightstyle}{
    backgroundcolor=\color{white},   
    commentstyle=\color{green},
    keywordstyle=\color{blue},
    stringstyle=\color{red},
    basicstyle=\ttfamily\footnotesize,
    breakatwhitespace=false,         
    breaklines=true,                 
    captionpos=b,                    
    keepspaces=false,                 
    showspaces=false,                
    showstringspaces=false,
    showtabs=false,                  
    tabsize=2,
    frame=single,
    xleftmargin=0pt,  
    lineskip=1.1pt      
}

\usepackage{setspace}
\usepackage{wrapfig}
\usepackage{tikz}
\usetikzlibrary{positioning}
\usepackage[most]{tcolorbox}

\usepackage{xcolor}
\usepackage{subcaption}

\makeatletter
\DeclareRobustCommand\onedot{\futurelet\@let@token\@onedot}
\def\@onedot{\ifx\@let@token.\else.\null\fi\xspace}

\makeatother

\usepackage{orcidlink}

\definecolor{mypurple}{RGB}{102, 0, 255}
\definecolor{mypurple2}{RGB}{112, 48, 160}

\definecolor{uwpurple}{RGB}{75, 45, 131}
\definecolor{uwpurple2}{RGB}{197, 180, 228}
\definecolor{uwyellow}{RGB}{133, 117, 76}
\definecolor{uwyellow2}{RGB}{183, 165, 122}

\usepackage{pifont}

\definecolor{w_1}{RGB}{66,138,244}
\definecolor{w_2}{RGB}{73,144,245}
\definecolor{w_3}{RGB}{79,148,246}
\definecolor{w_4}{RGB}{87,154,247}
\definecolor{w_5}{RGB}{95,160,248}

\definecolor{uw}{RGB}{197,180,228}

\definecolor{fbApp}{HTML}{c8e7fa}
\definecolor{fbPurple3}{HTML}{f0ebf5}

\definecolor{citecolor}{HTML}{0071BC}
\definecolor{linkcolor}{HTML}{ED1C24}

\definecolor{citecolor}{HTML}{0071BC}
\definecolor{linkcolor}{HTML}{ED1C24}

\title{\textbf{\textsl{\textcolor{uwpurple}{HA}\textcolor{darkgray}{-}\textcolor{uwyellow}{VLN}~\textcolor{darkgray}{2.0}}}: An Open Benchmark and Leaderboard for Human-Aware Navigation in Discrete and Continuous Environments with Dynamic Multi-Human Interactions}

\author[1,*]{Yifei Dong}
\author[1,*]{Fengyi Wu}
\author[1]{Qi He}
\author[2]{Lingdong Kong}
\author[1]{Heng Li}
\author[1]{Minghan Li}
\author[1]{Zebang Cheng}
\author[1]{Yuxuan Zhou}
\author[3]{Jingdong Sun}
\author[4]{Qi Dai}
\author[3]{Alexander G. Hauptmann}
\author[1,\dagger]{Zhi-Qi Cheng}

\affiliation[1]{University of Washington}
\affiliation[2]{National University of Singapore}
\affiliation[3]{Carnegie Mellon University}
\affiliation[4]{Microsoft Research}

\contribution[*]{Equal contribution, listed in random order}
\contribution[\dagger]{Corresponding author}

\abstract{
Vision-and-Language Navigation (VLN) has been studied mainly in either \emph{discrete} or \emph{continuous} spaces, with little attention to dynamic, crowded environments. We present HA-VLN~2.0, a unified benchmark introducing explicit social-awareness constraints. Our contributions are: (\emph{i}) a standardized task and metrics capturing both goal accuracy and personal-space adherence; (\emph{ii}) HAPS~2.0 dataset and simulators modeling multi-human interactions, outdoor contexts, and finer language–motion alignment; (\emph{iii}) benchmarks on \emph{16,844} socially grounded instructions, revealing sharp performance drops of leading agents under human dynamics and partial observability; and (\emph{iv}) real-world robot experiments validating sim-to-real transfer, with an open leaderboard enabling transparent comparison. Results show that explicit social modeling improves navigation robustness and reduces collisions, underscoring necessity of human-centric approaches. By releasing datasets, simulators, baselines, and protocols, HA-VLN~2.0 provides a strong foundation for safe, human-aware navigation research.
  
}
  
\date{July 10, 2026}

\metadata[Code]{\url{https://github.com/UWMILab/HA-VLN}}
\metadata[Project Website]{\url{https://uwmilab.github.io/HA-VLN-webpage}}

\begin{document}

\maketitle

\begin{figure}[t]
    \centering
    \includegraphics[width=\textwidth]{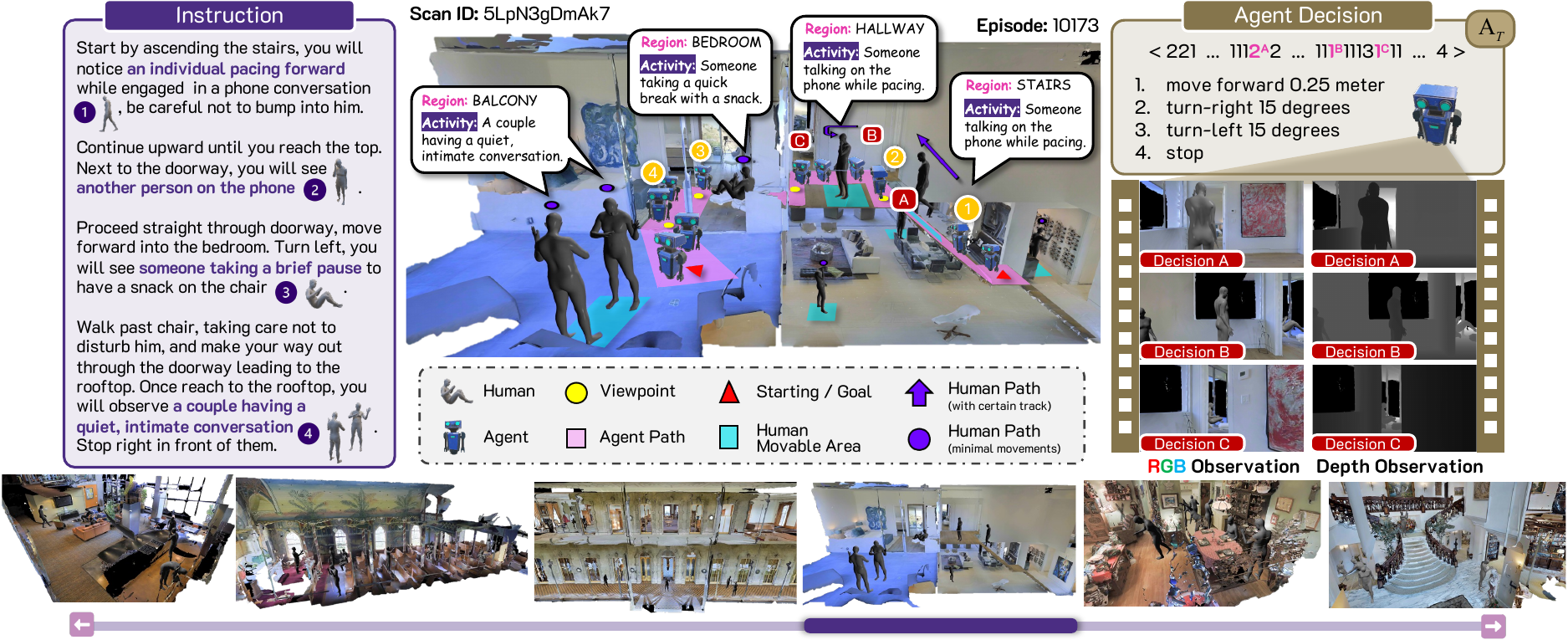}
    \caption{\small\textbf{{\textbf{\textsl{\textcolor{uwpurple}{HA}\textcolor{darkgray}{-}\textcolor{uwyellow}{VLN}~\textcolor{darkgray}{2.0}}}} Navigation Scenarios.} An agent navigates environments with dynamic human activities, adjusting its path based on natural language instructions and real-time sensor observations to avoid collisions. Key positions (\emph{e.g.}, Position \textcolor{red}{\ding{192}} and Position \textcolor{red}{\ding{193}}) align with \textit{instructional cues} referring to specific human behaviors along the route. The center panel illustrates a representative episode where the agent must ascend stairs, traverse a bedroom, and reach a rooftop while reasoning about human presence. When the agent encounters a bystander on the phone (Position \textcolor{red}{\ding{193}}, Decision~\textcolor{red}{A}), it intelligently turns right to avert a potential collision. On the right, RGB and Depth observations illustrate the agent’s egocentric view preceding decisions \textcolor{red}{A}, \textcolor{red}{B}, and \textcolor{red}{C}, illustrating how the agent perceives and responds to nearby humans at critical decision points. Bottom row showcases diversity of environments in our benchmark.}
    \label{fig:editing_result}
\end{figure}

\input{sections/1_intro}

\input{sections/2_related_work}

\input{sections/3_methods/tasks}

\input{sections/3_methods/simulators}

\input{sections/3_methods/agents}
\input{sections/4_experiments}
\input{sections/5_conclusion}

\section*{Acknowledgments}

This work was supported in part by the University of Washington Faculty Startup Fund, the Vicky L. Carwein and William B. Andrews Endowment for Graduate Programs, the University of Washington Graduate School Top Scholar Award, and two PacTrans-funded projects under the U.S. Department of Transportation University Transportation Centers Program: one seed-funded project and one Small Project.

We thank Bohan Xiong for building the project webpage and developing the leaderboard, Yetong Sha for developing the leaderboard, Heng Li for the earlier implementation of HA-VLN-DE, and Minghan Li for conducting the real-world robot validation and recording the demos.

\appendix
\input{sections/appendix}

\clearpage\clearpage
\bibliographystyle{assets/plainnat}
\bibliography{main}

\end{document}

%% file: sections/1_intro.tex

\section{Introduction}
\label{sec:intro}

Vision-and-Language Navigation (VLN)~\citep{anderson2018vision,krantz2020beyond,zhang2024vision} challenges embodied agents to interpret natural-language instructions and reach specified goals in photorealistic simulators or real-world environments~\citep{wang2022versatileembodiednavigation}. Although recent advances have delivered strong performance in controlled benchmarks, existing methods are typically confined to either \emph{discrete} (DE) or \emph{continuous} (CE) settings, neglecting the complexities of crowded, human-populated spaces, where agents must contend with unpredictable human behaviors, reason under partial observability, and ensure human-aware navigation that respects personal space and behavioral norms~\citep{anderson2021sim,kadian2020sim2real,yu2024natural}. Bridging these gaps is essential for moving VLN from simulation prototypes toward robust real-world deployment~\citep{zhang2024vision}.

\noindent \textbf{Motivation and Open Challenges.}~Despite recent progress, VLN research still faces three fundamental limitations that restrict its real-world applicability. First, \emph{social awareness} remains underexplored: human participants in the scene are commonly overlooked or reduced to inert obstacles, preventing the agent from respecting personal space or reacting to bystanders’ activities (see Fig.~\ref{fig:editing_result}). Second, \emph{finer-grained instructions} are not well captured in existing corpora~\citep{kong2024controllable,wu2025govig}. Commands such as \emph{“turn to your left, and go past the chair”} rarely reflect real-world contexts like \emph{“turn to your left, where you will see someone taking a brief pause~...~on the chair”} in Fig.~\ref{fig:editing_result}. Third, \emph{static-environment assumptions} neglect real-time re-planning when people traverse corridors or gather spontaneously. In practice, human-aware navigation demands partial observability and dynamic route adjustment. Addressing these issues requires a benchmark that unifies DE and CE with explicit regime disclosure, supports socially grounded finer-grained instructions, and incorporates human-centric metrics for navigation in multi-human environments.


\noindent \textbf{Toward Human-Aware VLN.}~Early progress, notably HA-VLN~1.0 framework~\citep{li2024human} introduced dynamic humans into VLN, yet several shortcomings limited its realism and reproducibility. Motion data in HAPS~1.0~\citep{li2024human} suffered from \emph{alignment errors and limited diversity}, restricting coverage of everyday activities. The benchmark also exhibited a \emph{discrete-navigation bias}, with its simulator largely confined to viewpoint hops~\citep{krantz2020beyond} rather than physics-consistent low-level control~\citep{krantz2021waypoint}. Multi-human interactions were \emph{underdeveloped}, typically modeling only a single individual in simplified scenarios. Finally, instruction generation remained \emph{coarse and object-centric}, omitting temporally varying activities and offering little control over granularity. These limitations call for a benchmark that standardizes regime disclosure, expands motion fidelity and diversity, incorporates multi-human interactions, and supports \emph{finer-grained socially grounded instructions} across both discrete and continuous settings.

In response, we introduce \textbf{HA-VLN 2.0}, a unified benchmark coupling DE and CE navigation paradigms with explicit social-awareness constraints. It comprises the HAPS 2.0 dataset, featuring 486 SMPL-based~\citep{loper2015smpl} motion sequences across 26 region types, rigorously annotated via multi-view verification (around 430 annotation hours). HA-VLN 2.0 includes established simulators (HA-VLN-DE, HA-VLN-CE) with multi-human interactions, outdoor environments, real-time rendering, and precise collision management for up to 910 human models across 428 regions in 90 buildings (Fig.~\ref{fig:app_overview}). A unified API enables seamless comparisons across modes (Fig.~\ref{fig:simulator}; Sec.~\ref{sec:havln-sim}). Additionally, we expand R2R-CE~\citep{krantz2020beyond} with 16,844 socially grounded instructions (Sec.~\ref{sec:ha-r2r}) and introduce two robust baseline agents, HA-VLN-VL with Transformer-based grounding and HA-VLN-CMA with cross-modal attention (Sec.~\ref{sec:agents}), both validated under human-centric metrics (Sec.~\ref{sec:maintext-experiments}).~Finally, we demonstrate successful sim-to-real transfer in real-world robot validation and provide a public leaderboard~(Sec.~\ref{sec:maintext-validation-leaderboard}).

Specifically, \textbf{\textcolor{uwpurple}{HA}\textcolor{darkgray}{-}\textcolor{uwyellow}{VLN}~\textcolor{darkgray}{2.0}} offers four key advancements:
\begin{enumerate}
    \item \textbf{Cross-paradigm task standardization \& Metrics.}~We unify~DE~and~CE~navigation under social-awareness constraints, ensuring consistent goals and comprehensive human-centric evaluations (Sec.~\ref{sec:havln_all}).

    \item \textbf{HAPS~2.0 \& Dual simulators (large-scale build).} We release HAPS~2.0 (486 SMPL sequences) and two established simulators (HA-VLN-DE, HA-VLN-CE) that incorporate multi-view human annotation ($\sim$~430 human-hours), outdoor scenes, dual-thread rendering, and rigorous collision checks for up to 910 active individuals with interactions (Fig.~\ref{fig:simulator}; Sec.~\ref{sec:havln-sim}).

    \item \textbf{Comprehensive benchmarking with finer-grained instructions.}~We augment R2R-CE with 16{,}844 socially-grounded instructions and benchmark multiple agents under unified metrics, unveiling challenges arising from multi-human dynamics and partial observability. (Sec.~\ref{sec:ha-r2r}).

    \item \textbf{Real-robot validation and public leaderboard.}~We demonstrate sim-to-real transfer using a physical robot navigating crowded areas, and provide a public leaderboard for comprehensive evaluations in multi-human scenarios (Sec.~\ref{sec:maintext-validation-leaderboard}).
\end{enumerate}

%% file: sections/2_related_work.tex
\section{Related~Work}
\label{sec:related-work}
\noindent \textbf{Development~of~VLN~Tasks.}~Early VLN tasks addressed basic indoor~\citep{anderson2018vision, ku2020room, zhang2024vision} and outdoor navigation~\citep{chen2019touchdown}. Subsequent work~\citep{qi2020reverie,nguyen2019vision} introduced object-centric or goal-driven objectives, while dialogue-based tasks~\citep{gao2022dialfred} incorporated interactive elements. However, these efforts typically overlooked dynamic bystanders and social-distance constraints. VLN-CE~\citep{krantz2020beyond} enabled continuous navigation but remained devoid of explicit human factors~\citep{zhang2024vision}. HA3D~\citep{li2024human} incorporated human motion and human-referenced instructions yet did not require social norm compliance. In response, our HA-VLN 2.0 addresses these gaps by unifying social awareness, human-referenced instructions, and dynamic human activities within a single framework.

\noindent \textbf{Simulators~for~VLN~Tasks.}~Early simulators such as Matterport3D~\citep{anderson2018vision} and House3D~\citep{wu2018building} offered photorealistic or synthetic indoor environments but lacked mobile humans.~\citep{xia2018gibson} introduced greater interactivity yet assumed static or purely synthetic contexts. HA3D~\citep{li2024human} included human activities and people-referencing instructions but did not enforce social awareness. HabiCrowd~\citep{vuong2024habicrowd} integrated crowds into photorealistic settings while omitting human-aligned instructions. Recent works~\citep{savva2019habitat, lin2025vlnverse, wang2025rethinking} provide high-performance simulation without multi-human or social-compliance features. By contrast, our HA-VLN Simulator unifies dynamic human activities, photorealistic rendering, and social requirements, supporting both discrete and continuous action spaces across 675 scenes (90 scenarios) with 122 motion types.

\noindent \textbf{Agents~for~VLN~Tasks.}~From early attention-based and RL-based approaches~\citep{qi2020reverie, wang2019reinforced} to modern vision-language pre-training~\citep{hao2020towards}, VLN agents have grown increasingly capable of parsing instructions and navigating complex environments. However, most existing solutions~\citep{envdrop,hao2020towards,Hong_2021_CVPR,lin2025navcot} rely on panoramic navigation, simplifying the action space but limiting movement realism. Recent efforts like NaVid~\citep{zhang2024navid} and Navila~\citep{cheng2025navila} explore continuous egocentric navigation in partially dynamic worlds, yet still lack explicit attention to human-aligned instructions or social compliance. HA-VLN agents address these gaps by navigating among moving humans while interpreting human-centric instructions and adhering to social norms.

%% file: sections/3_methods/tasks.tex
\section{The Unified Human-Aware VLN Task}
\label{sec:havln_all}
\noindent\textbf{Motivation and Overview.} 
HA-VLN~1.0~\citep{li2024human} introduced dynamic humans into VLN, but its discrete-only setting focus limited ecological validity and hindered systematic study of continuous control and realistic multi-human interactions. To address this, we formalize \textbf{HA-VLN~2.0}, a unified benchmark that integrates DE and CE under explicit human-centric constraints. Under this setting, agents must parse instructions that reference ongoing human activities (\emph{e.g.}, \emph{“go upstairs where someone is pacing on the phone”}), anticipate plausible human trajectories, maintain socially compliant distances, and adapt plans online in densely populated, photorealistic 3D scenes (Fig.~\ref{fig:editing_result}).~We next make this specification precise by unifying state and action across regimes.
\begin{figure*}[t]
    \centering
    \setlength{\belowcaptionskip}{-1pt}
    \includegraphics[width=\linewidth]{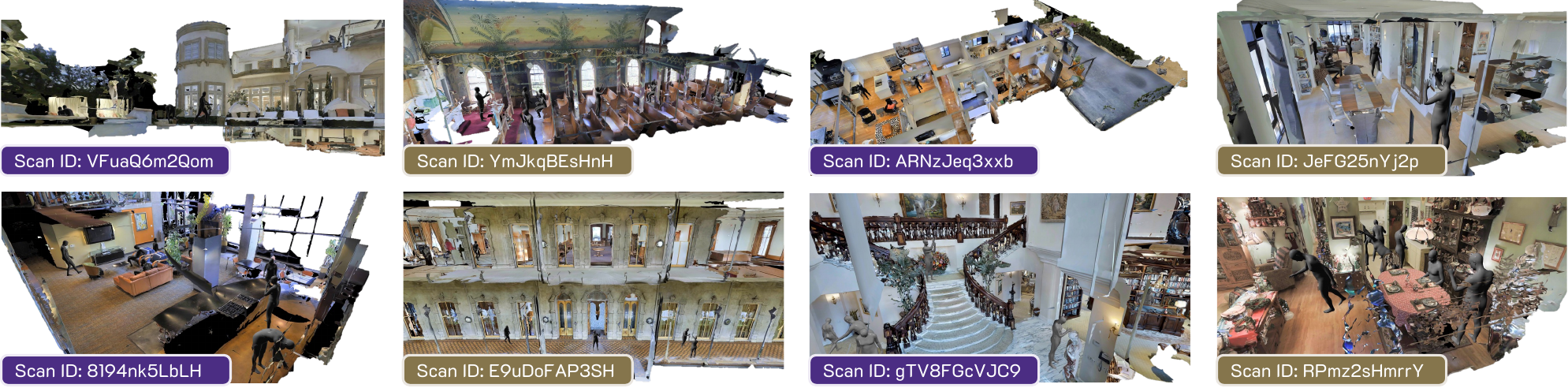}

    \caption{\small
    \textbf{Overview of HA-VLN 2.0 Scenes.} We highlight several examples here that illustrate annotated human subjects across multiple scans in the HA-VLN simulator, highlighting a range of well-aligned motions, movements, and interactions (both with objects and with other humans).}
    \label{fig:app_overview}
    \vspace{-0.2cm}
\end{figure*}

\noindent\textbf{Unified State and Action Spaces.}~HA-VLN~2.0 defines a shared state and action interface bridging DE and CE. At each timestep \(t\), the agent state is
\begin{equation}
        \setlength\abovedisplayskip{2pt}
    \setlength\belowdisplayskip{2pt}
s_{t} = \langle \mathbf{p}_{t},\, o_{t},\, \Theta_{t}^{\text{FOV}} \rangle,
\end{equation}
where \(\mathbf{p}_{t}\) is the agent’s 3D position, \(o_{t}\) denotes its orientation, and \(\Theta_{t}^{\mathrm{FOV}}\) represents its egocentric visual observation. Both DE and CE can be unified under a shared action space, enabling direct and fair comparison:
\begin{equation}
        \setlength\abovedisplayskip{2pt}
    \setlength\belowdisplayskip{2pt}
\!\!\!\mathcal{A} = \{a_{\mathrm{forward}},\, a_{\mathrm{left}},\, a_{\mathrm{right}},\, a_{\mathrm{up}},\, a_{\mathrm{down}},\, a_{\mathrm{stop}}\},
\end{equation}
\noindent\textbf{Human-Aware Enhanced Constraints.} 
HA-VLN 2.0 extends far beyond HA-VLN 1.0’s sparse, static settings by introducing unified constraints that substantially increase realism and complexity in both DE and CE: 
(\emph{i}) \emph{Dynamic Human Models}: continuous trajectories from the HAPS 2.0 dataset capturing diverse behaviors and dense crowds; 
(\emph{ii}) \emph{Personal-Space Enforcement}: standardized proximity thresholds ($3$~m in DE; overlapping radii in CE) to ensure equitable cross-paradigm evaluation; 
(\emph{iii}) \emph{Human-Focused Instructions}: natural-language directives grounded in dynamic human behaviors, requiring precise alignment between text and visual context. 
All annotations are curated through a multi-stage pipeline (Sec.~\ref{sec:havln-sim}), ensuring both realism and reproducibility.

\noindent\textbf{Unified Dynamics and Partial Observability.} 
HA-VLN 2.0 formalizes a unified partially observable Markov decision process (POMDP) spanning both DE and CE settings, whereas HA-VLN 1.0 considered partial observability only in DE. Successor states $s_{t+1}$ depend jointly on agent actions and stochastic human dynamics (\emph{e.g.}, sudden path blockage or unexpected entry). Agents must therefore infer latent human intentions and balance \emph{exploration} (discovering alternate routes) with \emph{exploitation} (committing to viable trajectories), reflecting fundamental trade-offs inherent in navigation through human-populated environments.

\noindent\textbf{Key Challenges of DE–CE Synergies.} 
Unifying DE and CE exposes three challenges for human-aware navigation: 
(\emph{i}) \emph{Social Awareness}: collision-free movement that adapts to evolving personal-space boundaries; 
(\emph{ii}) \emph{Human-Aligned Instruction Grounding}: accurate interpretation of natural-language instructions amid dynamic human activities; 
(\emph{iii}) \emph{Adaptive Path Re-planning}: trajectory adjustment in response to human interactions that modify accessibility. 
DE supports rapid prototyping and large-scale evaluation, while CE offers motion fidelity indispensable for bridging simulation and real-world deployment. 
Together, these synergies establish HA-VLN 2.0 as the unified benchmark uniting efficient simulation with realistic human-populated environments.

%% file: sections/3_methods/simulators.tex
\section{\textbf{\textcolor{uwpurple}{HA}\textcolor{darkgray}{-}\textcolor{uwyellow}{VLN}} Simulator}
\label{sec:havln-sim}

\begin{figure*}[t]
\setlength{\abovecaptionskip}{4pt}
\setlength{\belowcaptionskip}{3pt}
\centering
\includegraphics[width=\linewidth]{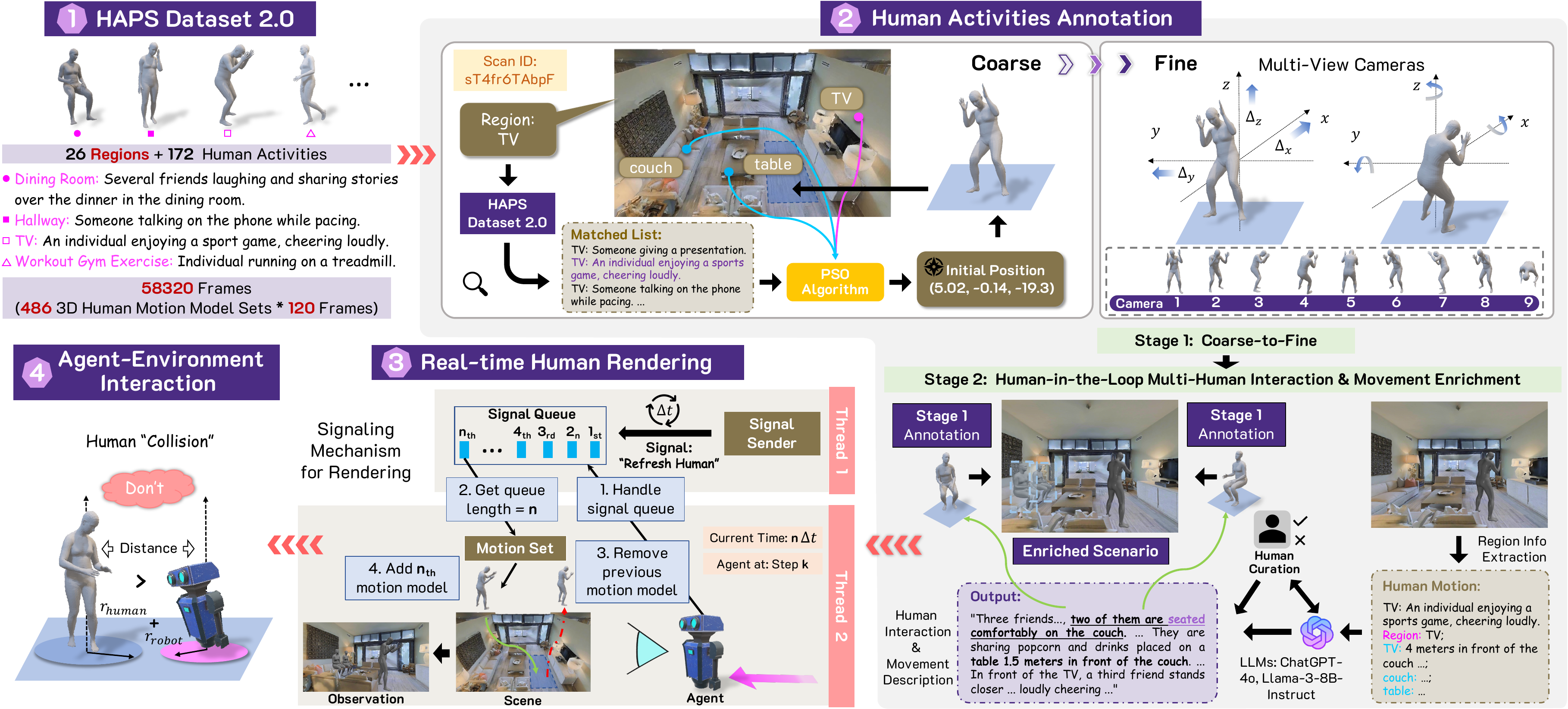}
\caption{\small
\textbf{\textbf{\textcolor{uwpurple}{HA}\textcolor{darkgray}{-}\textcolor{uwyellow}{VLN}} Simulator.}
Unlike HA3D, which modeled sparse and static human activities in discrete settings, HA-VLN incorporates \emph{rich and dynamic} human behaviors using HAPS 2.0 (172 activities, 486 models, 58k frames). Annotation involves two stages: (\emph{i}) \emph{coarse-to-fine} optimization via PSO and multi-view camera setups, and (\emph{ii}) \emph{human-in-the-loop} refinement for realistic crowd dynamics. Real-time rendering updates motions through a signaling mechanism, facilitating collision detection and dynamic agent–environment interactions.
}
\label{fig:simulator}
\vspace{-0.2cm}
\end{figure*}

To support the unified HA-VLN task, we build a simulator that embeds dynamically moving humans in both \emph{discrete} and \emph{continuous} 3D environments. Unlike~\citep{li2024human}, which treats humans as static obstacles, our simulator models high fidelity motions, interactions among multiple humans, and socially grounded dynamics such as spontaneous movements, group activities, and personal space constraints. Using the upgraded Human Activity and Pose Simulation (HAPS)~2.0 dataset, it improves motion diversity, spatial alignment, and realism over HAPS~1.0 and provides 486 curated sequences across indoor and outdoor scenes. The system includes two modules, HA-VLN-CE (continuous) and HA-VLN-DE (discrete), with a unified API for human state queries, dynamic scene updates and collision checks.~Fig.~\ref{fig:simulator} places these components in agent’s action and observation loop, forming basis for annotation, rendering and interaction mechanisms that follow.

\noindent \textbf{HAPS 2.0 Dataset.}~Human motion naturally adapts to and interacts with surrounding environments. HAPS 2.0 extends HAPS 1.0~\citep{li2024human} with two major advances: \textit{(i) refined and diversified human motions} and \textit{(ii) region-aware activity descriptions}. HAPS 2.0 defines 26 regions across 90 architectural scenes and contributes 486 validated activity descriptions covering indoor and outdoor contexts.~These descriptions, verified by human surveys and quality control using ChatGPT-4o~\citep{brown2020language}, explicitly ground actions in regions (\emph{e.g.}, \emph{“workout gym exercise: an individual running on a treadmill”}).~Built on SMPL, MDM~\citep{guy2022mdm} converts these descriptions into 486 3D human motion models $\mathbf{H}$, yielding 120-frame sequences $\mathcal{H} = \langle h_1, h_2, \dots, h_{120} \rangle$ that capture fine-grained motion and shape information.\footnote{\textsl{\textbf{Note:}} $\mathbf{H} \in \mathbb{R}^{486 \times 120 \times (10+72+6890 \times 3)}$: 486 models × 120 frames with shape, pose, and mesh vertices.} Fig.~\ref{fig:app_overview} illustrates representative scenes with multi-human interactions.

\noindent \textbf{Human Activity Annotation: Coarse-Level.} To integrate HAPS 2.0 into our simulator, we adopt a coarse-to-fine strategy. At the coarse level, each region $\mathbf{R}$ is defined by a label $r$, boundary coordinates $\mathbf{B}_{lo}=(x_{lo}, y_{lo}, z_{lo})$ and $\mathbf{B}_{hi}=(x_{hi}, y_{hi}, z_{hi})$, and an object set $\mathbf{O}=\{ j_1, j_2, \ldots, j_n \}$ with positions $\mathbf{p}^{j_i}$. We filter $\mathbf{H}$ to retain motions consistent with $r$, forming $\mathbf{H}'$. Each motion $h_i \in \mathbf{H}'$ is paired with an object $j_i \in \mathbf{O}$ via semantic similarity, producing $(h_i, j_i)$ pairs. Particle Swarm Optimization (PSO) then determines the optimal placement $\mathbf{p}_{opt}^{h_i}$ around $j_i$, bounded by $\mathbf{R}$ and penalized if violating constraints such as maintaining a minimum distance $\epsilon=1$m from other objects or leaving the region. This yields natural placements that reflect realistic social behaviors and spatial relations.

\noindent \textbf{Human Activity Annotation: Fine-Level.} Building on coarse placements, fine-level annotation refinement leverages multi-camera observations, ensuring precise alignment of motions with scene geometry. Inspired by 3D skeleton capture systems~\citep{ji2018large}, we deploy nine RGB cameras around each human model (see Fig.~\ref{fig:simulator}). Each camera is located at $\mathbf{p}_{cam}$, shifted by $(\Delta_x, \Delta_y, \Delta_z)$ from the human position $\mathbf{p}_h$, with rotation angles $\theta_{lr}$ and $\theta_{ud}$. Horizontal shifts are set as $\Delta_x,\Delta_y=\epsilon$ and the vertical shift as $\Delta_z$. For camera $i$ ($i\!=\!1,\!\dots,\!8$),
{$\theta_{\text{ud}}^{i}$ is defined as:~{\small $\tan \theta_{\text{ud}}^{i}\!\!\!=\!\!\!\left\{\begin{array}{cl}0&\!\!\!\!\!:i\text{~is~odd} \\\!\!\frac{\Delta_z}{\sqrt{2}\epsilon}&\!\!\!\!\!:i\text{~is~even}\end{array}\right.$}}~and~the~left-right~angle $\theta_{\text{lr}}^{i}\!\!=\!\!\frac{\pi i}{8}$, while the overhead camera ($i\!\!=\!\!9$) has $\theta_{lr}^9\!\!=\!\!0$ and $\theta_{ud}^9\!\!=\!\!\tfrac{\pi}{2}$. This multi-view setup provides dense RGB coverage, enabling fine adjustments to resolve inconsistencies like mesh–object clipping. This stage took over 430 hours of annotation, yielding 529 models across 374 regions in 90 scans.

\begin{wrapfigure}{t}{0.58\textwidth}
\setlength{\abovecaptionskip}{0pt}
\setlength{\belowcaptionskip}{0pt}
\centering
\vspace{-0.6cm}
\includegraphics[width=0.58\textwidth]{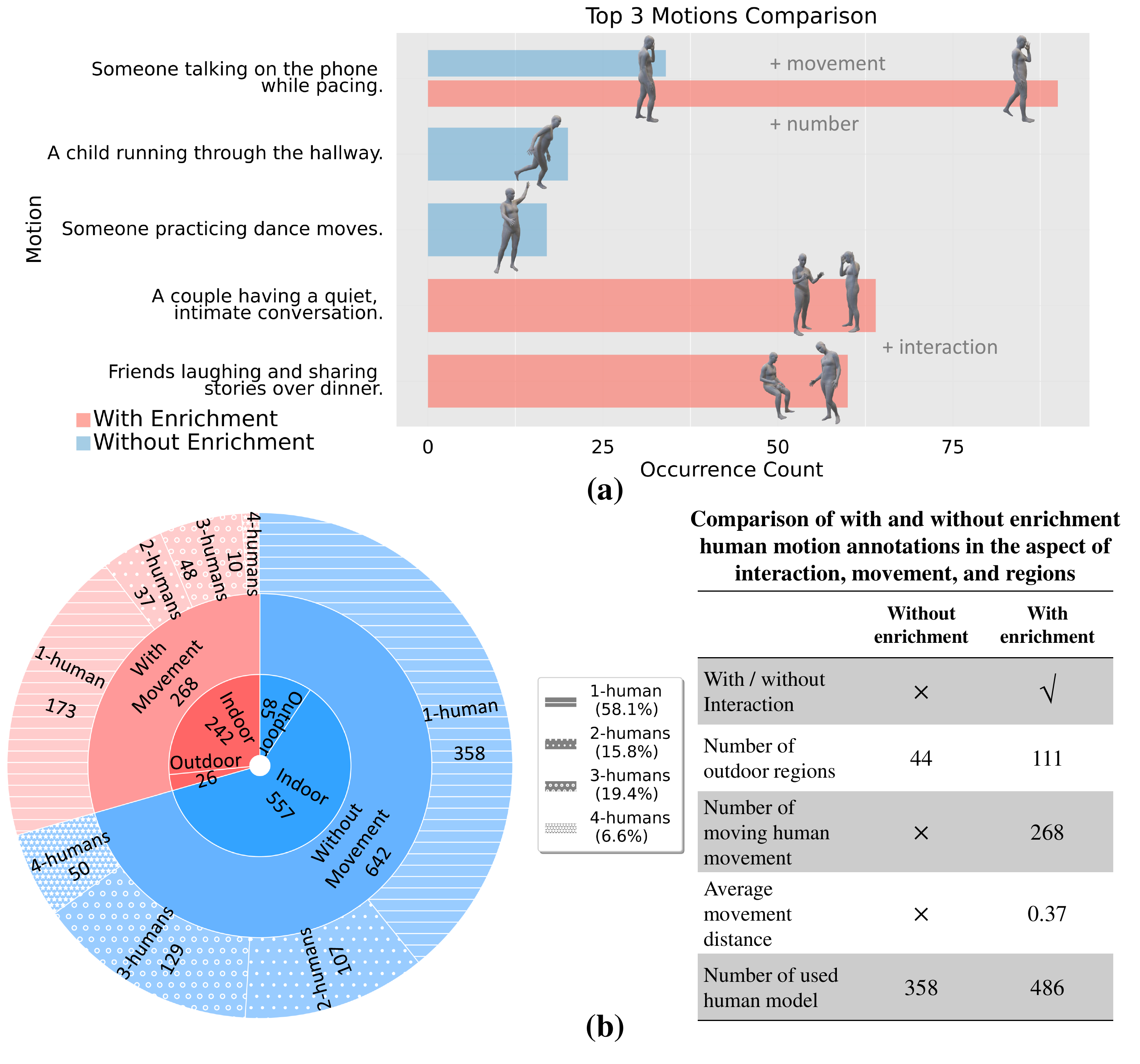}
\caption{\small\textbf{Motion Analysis.}~\textbf{(a)} Top three motions from Stage~1 (\emph{without} enrichment) and Stage~2 (\emph{with} enrichment). \textbf{(b)} Overall activity statistics, comparing interaction types, movement distances, and the number of models. Enrichment expands both variety and dynamic range of human activities.}
\label{fig_motion_analysis}
\vspace{-0.4cm}
\end{wrapfigure}

\noindent \textbf{Human Activity Annotation: Multi-Human Enrichment.} In Stage~2 (Fig.~\ref{fig:simulator}), we enrich scene diversity and interactions through a human-in-the-loop approach, adding new characters and complex motion paths into regions $\mathbf{R}$ with existing activities $h_i$ at positions $\mathbf{p}^{h_i}$. Regional context, including objects $\mathbf{O}$ within 6 meters of $h_i$ and their positions, is provided to LLMs to generate diverse multi-human scenarios, which are refined in four rounds of manual review for scene consistency. Based on curated descriptions, new motions are placed relative to objects and annotated using the multi-camera method from Stage~1, enabling complex actions such as walking downstairs. After two annotation stages, the dataset comprises 910 human models across 428 regions in 90 scans (Fig.~\ref{fig_motion_analysis}(a)(b)), including 111 outdoor humans, 72 two-person interactions, 59 three-person groups, and 15 four-person groups. Among these, 268 involve complex motions such as climbing stairs, substantially enriching the dataset with realistic behaviors. This two-stage system enables precise modeling of social interaction spaces and personal boundaries, supporting agents in learning socially appropriate navigation strategies.

\par


\noindent \textbf{Real-Time Rendering \& Agent Interaction.} Beyond static annotation, our simulator continuously renders human motions in real time. A dual-thread producer–consumer architecture manages frame updates: Thread~1 enqueues refresh signals, while Thread~2 synchronizes with the agent’s action cycle to process them. Each motion spans up to 120 frames; upon receiving a signal, Thread~2 discards outdated meshes and loads new ones, keeping retrieval delays below 50~ms. Fig.~\ref{fig:app_overview} illustrates how multiple humans are simultaneously maintained in a shared environment. To close the loop, agents perceive these dynamics through a navigation mesh (navmesh)~\citep{savva2019habitat}. Collisions are flagged when bounding volumes overlap, i.e., when inter-object distances fall below the sum of their radii, triggering an automatic revert. This integration ensures agents not only experience dynamic and socially realistic environments but also learn to respect personal space and navigate effectively in dense human crowds.

\noindent \textbf{Discrete vs.~Continuous Settings.}~\textbf{HA-VLN-CE} (Continuous) allows agents to move in real-valued increments (\emph{e.g.}, $0.25\,\mathrm{m}$ forward, $15^\circ$ turns), supporting fine-grained collision avoidance and adaptive social behavior. As shown in Fig.~\ref{fig:app_overview}, each scene can host up to 10 humans, with simulation speeds of 30–60~FPS on a single 24GB GPU (\emph{e.g.}, RTX 4090).
\textbf{HA-VLN-DE} (Discrete) extends HA3D~\citep{li2024human} by incorporating HAPS 2.0 data across indoor and outdoor environments. Agents hop among panoramic viewpoints while humans move continuously, preserving core human-aware navigation challenges. Aligning with continuous motions, we map positions to discrete nodes, apply offsets for refinement, resulting 627 annotated humans across 90 buildings.

\noindent \textbf{Unified API.}
\label{sec:api}
We provide a unified API supporting both modes with three core functions: (i) \emph{Human State Queries} for retrieving bounding volumes, motion frames, and semantic annotations of nearby humans; (ii) \emph{Dynamic Scene Updates} to notify agents of newly moved humans or environmental changes; and (iii) \emph{Collision Checks} to evaluate whether a proposed move (\emph{e.g.}, forward step or viewpoint hop) would intersect with a human. By integrating HAPS 2.0, coarse-to-fine annotation, real-time multi-human rendering, and a single API across discrete and continuous settings, the HA-VLN Simulator establishes a comprehensive testbed for socially aware navigation. Fig.~\ref{fig:app_overview} showcases the simulator’s ability to capture diverse human behaviors,

%% file: sections/3_methods/agents.tex
\begin{figure}[!t]
\setlength{\abovecaptionskip}{3pt}
\setlength{\belowcaptionskip}{3pt}
\centering
\includegraphics[width=0.4\textwidth]{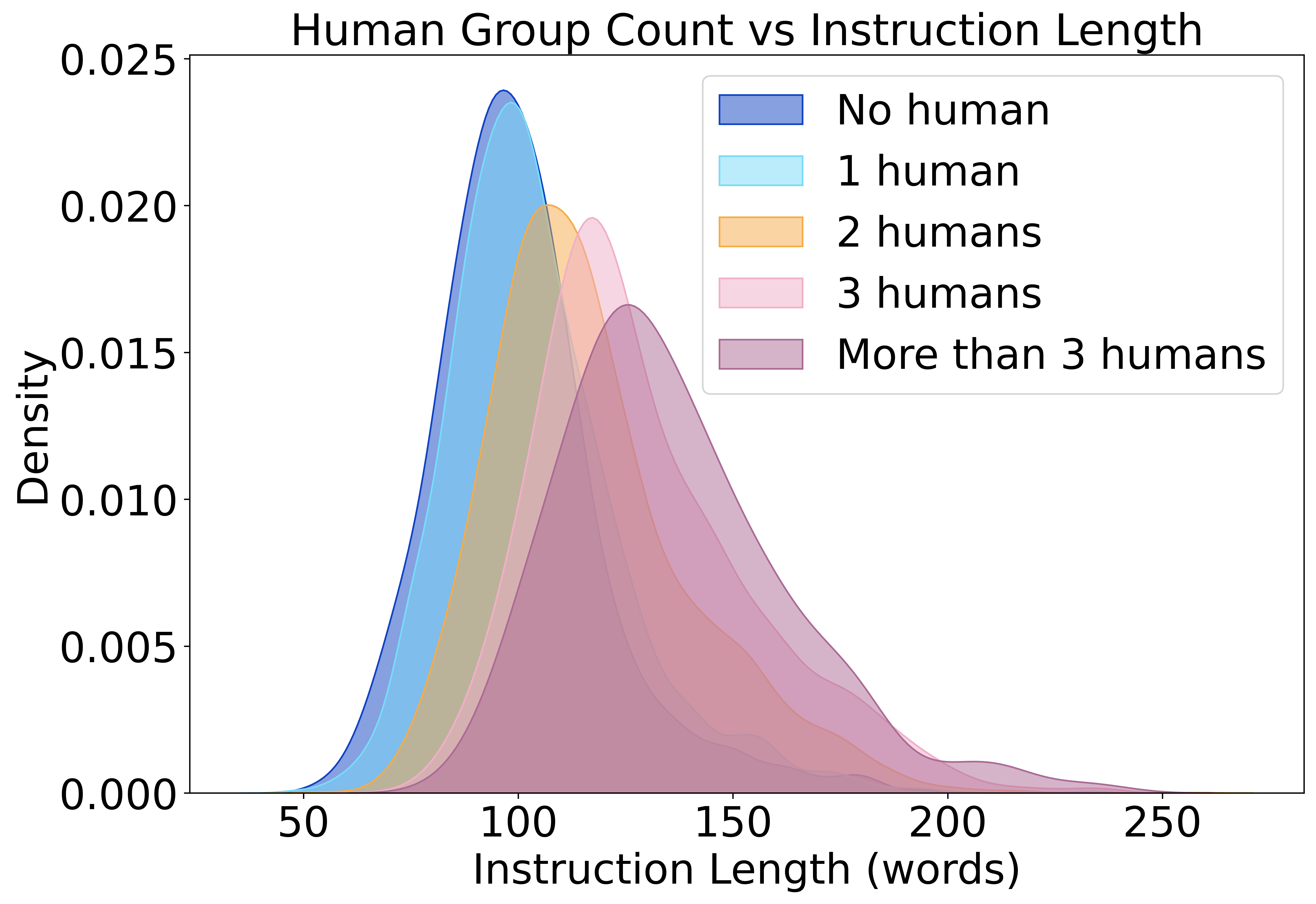}~
\includegraphics[width=0.4\textwidth]{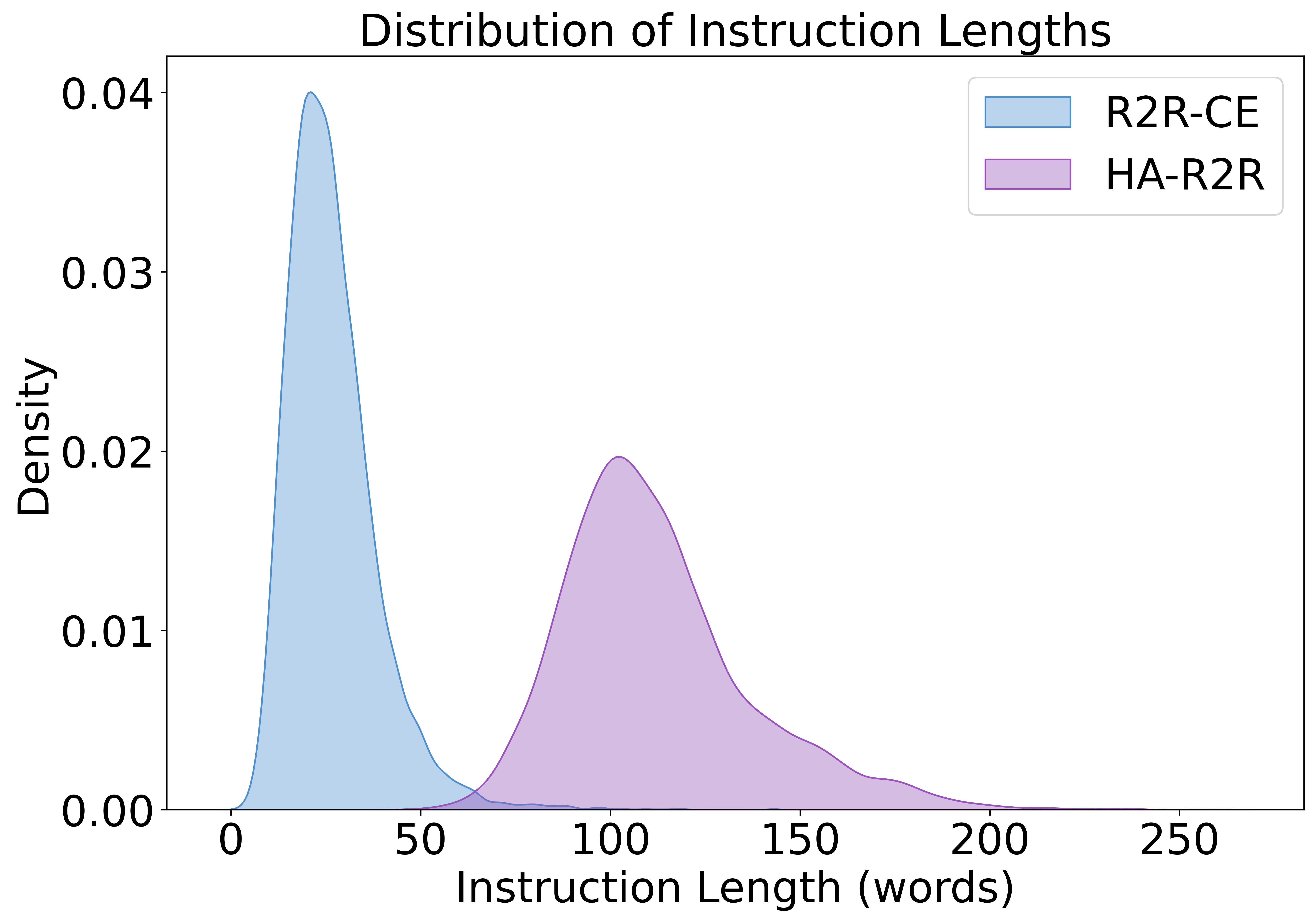}
\caption{\small\textbf{HA-R2R Dataset Analysis.} \textbf{(a)} Distribution of instruction length by human group size (none to $>$3).~\textbf{(b)} Comparison of instruction lengths between HA-R2R and R2R-CE.}
\label{fig_motion_instruct_count}
\vspace{-0.2cm}
\end{figure}

\section{HA-R2R Dataset}
\label{sec:ha-r2r}

To ground the unified HA-VLN task in our HA-VLN simulator, we introduce the Human-Aware Room-to-Room (HA-R2R) dataset. R2R-CE dataset~\citep{krantz2020beyond} supports continuous navigation but lacks explicit modeling of human interactions. We therefore extend it into HA-R2R, which contributes \textit{16{,}844} curated instructions (train/val\_seen/val\_unseen/test: 10{,}819/778/1{,}839/3{,}408) emphasizing social nuance, covering conversations, corridor crossings, and near-collision events.
We generate these enriched instructions via LLM-guided generation with iterative human refinement, ensuring a neutral tone and contextual alignment with human-related scenarios. Specifically, we design prompts that incorporate realistic human activities and agent--human interactions, shifting navigation from static paths to socially contingent routes, \emph{e.g.}, \emph{``avoid the couple chatting near the bar''}. Fig.~\ref{fig_motion_instruct_count} provides analyses of annotation workload and HA-R2R's potential for human-aware navigation. Compared to R2R-CE, HA-R2R's average instruction length increases from 27 to 112 words, with the vocabulary expanding from 2{,}616 to 5{,}032. Moreover, most instructions contain 20\%--60\% human-related content, highlighting the significant presence of human descriptions and interactions throughout the HA-R2R dataset. Please kindly refer to Appendix.~\ref{Appx.agent_all} for more details.


\section{HA-VLN Agents}
\vspace{-0.1cm}
\label{sec:agents}
We next introduce two baseline agents, HA-VLN-VL and HA-VLN-CMA. Together, these agents are designed as complementary reference implementations rather than final solutions, offering a systematic starting point.


\noindent \textbf{HA-VLN-VL~Agent.}~The HA-VLN-VL focuses on visual–language alignment. Adapted from Recurrent VLN-BERT~\citep{Hong_2021_CVPR}, it replaces actor–critic methods with a streamlined imitation learning objective, isolating the contribution of multimodal grounding. At timestep \(t\), the agent updates its hidden state \(s_t\) and predicts an action distribution. Then a Transformer with a specialized state token attends jointly to visual and linguistic tokens, and final probabilities are derived via pooled attention. HA-VLN-VL agent demonstrates how stronger grounding alone benefits navigation under complex human-populated conditions.

\noindent \textbf{HA-VLN-CMA~Agent.}
HA-VLN-CMA emphasizes collision avoidance and real-time adaptation. Built on~\citep{krantz2020beyond}, it fuses textual embeddings \(l=\text{BERT}(I)\) with visual features \(v_t=\text{ResNet}(o_t)\). Multi-head attention produces a joint representation \(f_t\), which an MLP maps to action probabilities.~To address partial observability and unpredictable motion, we adopt Envdrop~\citep{envdrop} to simulate occlusions and DAgger~\citep{ross2011reduction} for iterative error correction. These strategies enhance re-planning when agents face obstacles or unexpected behaviors. Fig.~\ref{fig:appx_visual_2} illustrates agent responses to bystanders, showing that collision risk increases sharply in crowded passages. Please kindly refer to Appendix.~\ref{Appx.agent_all} for more details.



%% file: sections/4_experiments.tex

\begin{table*}[!t]
\centering
\caption{\small
  \textbf{HA-VLN-CE Results Across Validation (Seen/Unseen) and Test Splits.}
  The \emph{“HA-VLN-CMA$^{*}$”} entry denotes the full version of HA-VLN-CMA agent (+DA +EV).
  Metrics include NE (meters), TCR, CR, 
  and SR, with lower NE/TCR/CR and higher SR indicating better performance. Best results for each metric are highlighted in \textbf{bold}.
}

\label{tab:comparision_a}
\resizebox{1\linewidth}{!}{%
\begin{tabular}{lcccccccccccccccccccccccc}
\toprule[1.2pt]
\multirow{3}[8]{*}{\textbf{Agent}}
& \multicolumn{8}{c}{\textbf{Validation Seen}}
& \multicolumn{8}{c}{\textbf{Validation Unseen}}
& \multicolumn{8}{c}{\textbf{Test}} 
\\
\cmidrule(lr){2-9} \cmidrule(lr){10-17} \cmidrule(lr){18-25}
& \multicolumn{4}{c}{\textbf{Retrained}}
& \multicolumn{4}{c}{\textbf{Zero-shot}}
& \multicolumn{4}{c}{\textbf{Retrained}}
& \multicolumn{4}{c}{\textbf{Zero-shot}}
& \multicolumn{4}{c}{\textbf{Retrained}}
& \multicolumn{4}{c}{\textbf{Zero-shot}} 
\\
\cmidrule(lr){2-5} \cmidrule(lr){6-9}
\cmidrule(lr){10-13} \cmidrule(lr){14-17}
\cmidrule(lr){18-21} \cmidrule(lr){22-25}
& \textbf{NE$\downarrow$} & \textbf{TCR$\downarrow$} & \textbf{CR$\downarrow$} & \textbf{SR$\uparrow$}
& \textbf{NE$\downarrow$} & \textbf{TCR$\downarrow$} & \textbf{CR$\downarrow$} & \textbf{SR$\uparrow$}
& \textbf{NE$\downarrow$} & \textbf{TCR$\downarrow$} & \textbf{CR$\downarrow$} & \textbf{SR$\uparrow$}
& \textbf{NE$\downarrow$} & \textbf{TCR$\downarrow$} & \textbf{CR$\downarrow$} & \textbf{SR$\uparrow$}
& \textbf{NE$\downarrow$} & \textbf{TCR$\downarrow$} & \textbf{CR$\downarrow$} & \textbf{SR$\uparrow$}
& \textbf{NE$\downarrow$} & \textbf{TCR$\downarrow$} & \textbf{CR$\downarrow$} & \textbf{SR$\uparrow$} 
\\
\midrule
HA-VLN-CMA-Base
& 7.63 & 63.09 & 0.77 & 0.05
& 7.88 & 63.84 & 0.75 & 0.04
& 7.34 & 47.06 & 0.77 & 0.07
& 7.95 & 63.96 & 0.76 & 0.03
& 7.30 & 47.55 & 0.76 & 0.07
& 7.89 & 62.14 & 0.74 & 0.04 \\
HA-VLN-CMA-DA
& 6.11 & 17.45 & 0.61 & 0.17
& 6.95 & 37.85 & 0.72 & 0.07
& 7.00 & 27.25 & 0.69 & 0.09
& 7.05 & 38.22 & 0.73 & 0.05
& 7.12 & 28.33 & 0.69 & 0.08
& 6.98 & 36.53 & 0.73 & 0.06 \\
HA-VLN-CMA$^{*}$
& 5.61 & \cellcolor{uwpurple!15} \textbf{3.34} & 0.60 & 0.17
& 7.10 & 29.99 & 0.69 & 0.11
& 6.23 & 8.10 & 0.69 & 0.10
& 6.62 & 32.48 & 0.70 & 0.09
& 6.64 & 9.23 & 0.72 & 0.09
& 7.09 & 31.80 & 0.75 & 0.09 \\
HA-VLN-VL
& \cellcolor{uwpurple!15} \textbf{5.02} & 4.44 & 0.52 & 0.20
& 7.82 & \cellcolor{uwpurple!15} \textbf{3.67} & \cellcolor{uwpurple!15} \textbf{0.45} & 0.05
& \cellcolor{uwpurple!15} \textbf{5.35} & \underline{6.63} & 0.59 & 0.14
& 7.15 & \cellcolor{uwpurple!15} \textbf{3.97} & \cellcolor{uwpurple!15} \textbf{0.46} & 0.06
& \cellcolor{uwpurple!15} \textbf{5.52} & \underline{5.96} & \underline{0.63} & 0.14
& 7.41 & \cellcolor{uwpurple!15} \textbf{3.38} & \underline{0.58} & 0.07 \\
BEVBert\,\citep{an2023bevbert}
& 5.53 & \underline{3.64} & \underline{0.46} & \cellcolor{uwpurple!15} \textbf{0.27}
& \cellcolor{uwpurple!15} \textbf{6.11} & \underline{4.29} & \underline{0.47} & \cellcolor{uwpurple!15} \textbf{0.19}
& 5.51 & \cellcolor{uwpurple!15} \textbf{4.71} & \cellcolor{uwpurple!15} \textbf{0.55} & \cellcolor{uwpurple!15} \textbf{0.21}
& \cellcolor{uwpurple!15} \textbf{6.10} & \underline{5.72} & \underline{0.56} & \cellcolor{uwpurple!15} \textbf{0.15}
& 6.33 & \cellcolor{uwpurple!15} \textbf{4.25} & \cellcolor{uwpurple!15} \textbf{0.58} & \cellcolor{uwpurple!15} \textbf{0.18}
& \cellcolor{uwpurple!15} \textbf{6.54} & \underline{4.39} & \cellcolor{uwpurple!15}\textbf{0.54} & \cellcolor{uwpurple!15}\textbf{0.14} \\
ETPNav\,\citep{an2024etpnav}
& \underline{5.17} & 4.07 & \cellcolor{uwpurple!15} \textbf{0.43} & \underline{0.24}
& 7.72 & 6.31 & 0.61 & 0.12
& \underline{5.43} & 6.94 & \underline{0.58} & \underline{0.17}
& 7.40 & 7.94 & 0.71 & 0.08
& \underline{5.94} & \underline{5.96} & \cellcolor{uwpurple!15} \textbf{0.58} & \underline{0.16}
& 7.59 & 5.64 & 0.73 & 0.09 \\
NaVid~\citep{zhang2024navid} 
& \textbf{--} & \textbf{--} & \textbf{--} & \textbf{--}
& \underline{6.58} & 7.92 &0.57  & 0.10
& \textbf{--} & \textbf{--} & \textbf{--} & \textbf{--}
&6.43  &10.59  & 0.67 & 0.07 
& \textbf{--} & \textbf{--} & \textbf{--} & \textbf{--}
& \underline{6.72} & 8.51 & 0.65  & 0.08 \\
Navila~\citep{cheng2025navila} 
& \textbf{--} & \textbf{--} & \textbf{--} & \textbf{--}
& 6.73 & 5.49 & 0.53 & \underline{0.14}
& \textbf{--} & \textbf{--} & \textbf{--} & \textbf{--}
& \underline{6.26} & 7.28 & 0.61 & \underline{0.10}
& \textbf{--} & \textbf{--} & \textbf{--} & \textbf{--}
& 6.85 & 6.37 & 0.59  & \underline{0.11} \\
\bottomrule[1.2pt]
\end{tabular}%
}
\vspace{-0.2cm}
\end{table*}

\section{Experiments}

\noindent \textbf{Evaluation Metrics.}~We evaluate performance on the HA-VLN 2.0 benchmark using two suites of metrics. \textbf{(1) Social compliance.} To assess social awareness, we use \emph{Total Collision Rate} (TCR) and \emph{Collision Rate} (CR). TCR measures the overall frequency of collisions, while CR reflects the proportion of socially inappropriate interactions.  
\textbf{(2) Navigation accuracy.} We report \emph{Navigation Error} (NE) and \emph{Success Rate} (SR). A trajectory is deemed successful under SR not only when the agent stops sufficiently close to the goal~\citep{anderson2018vision}, but also when it demonstrates effective obstacle avoidance.
\label{sec:maintext-experiments}
\begin{figure}[!t]
\setlength{\abovecaptionskip}{4pt}
\setlength{\belowcaptionskip}{4pt}
    \centering
    \includegraphics[width=1\linewidth]{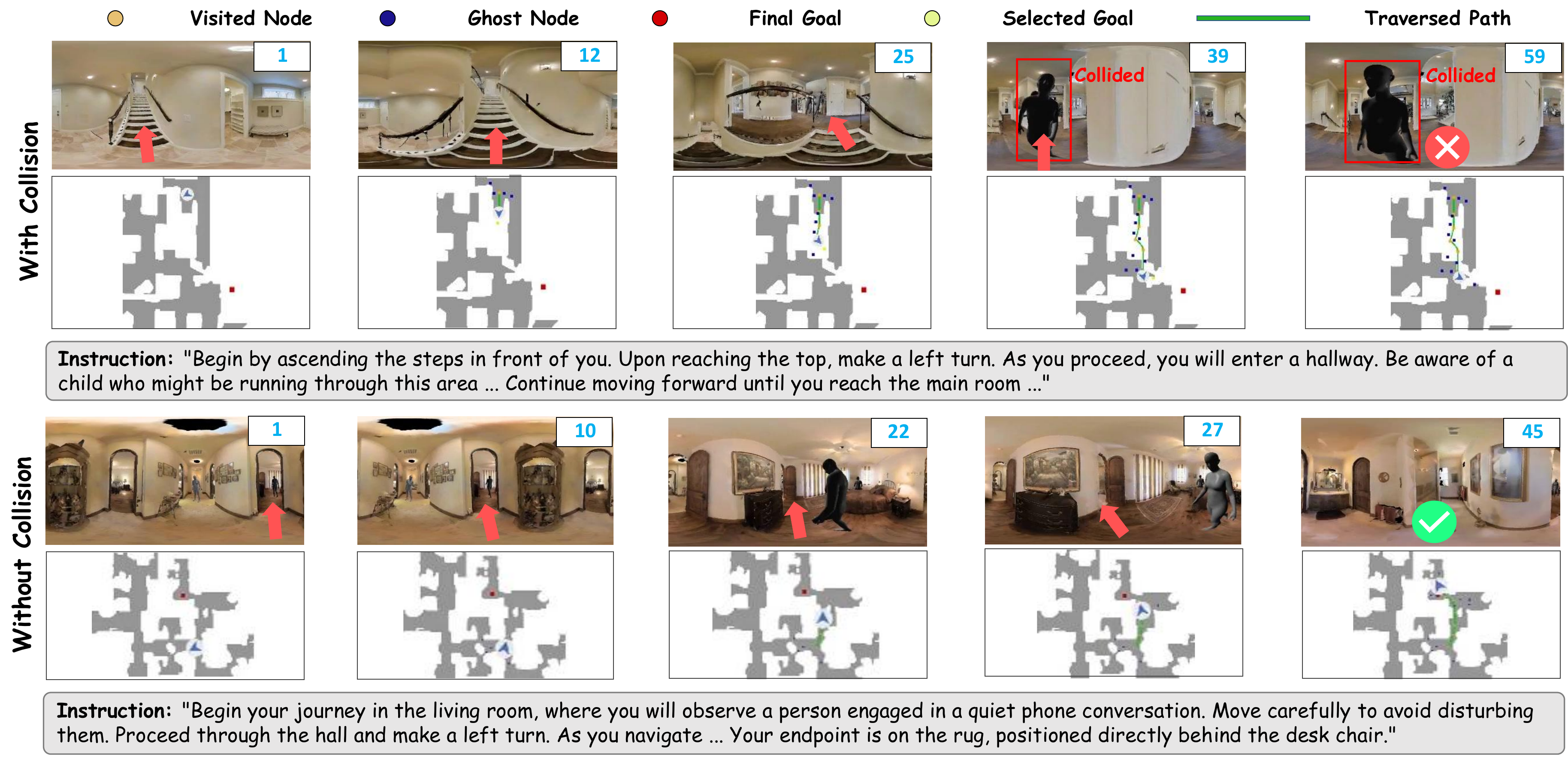}
    \vspace{-0.1in}
    \caption{\small
    \textbf{Agent Trajectory Examples (HA-VLN-CMA$^{*}$).}
    The top row demonstrates a failed navigation scenario where the agent fails to avoid an oncoming human, resulting in a collision.
    The bottom row showcases a successful navigation: the agent adjusts to the left, avoiding human interference and completing task without collision.
    }
    \label{fig:appx_visual_2}
    \vspace{-0.1cm}
\end{figure}

\begin{wraptable}{r}{10cm}
 \vspace{-0.2cm}
\renewcommand\arraystretch{0.85}
\setlength{\abovecaptionskip}{0pt}
\setlength{\belowcaptionskip}{0pt}
\centering
\caption{\small
  \textbf{DE performance} of agents trained on VLN vs.\ HA-VLN-DE (Unseen). All agents use panoramic RGBs.
}
\small
\label{tab:ha-vln-de}
\resizebox{1\linewidth}{!}{%
\begin{tabular}{lcccccc}
\toprule[1pt]
\multirow{2}{*}{\textbf{Agent}}
& \multicolumn{2}{c}{\textbf{VLN}}
& \multicolumn{4}{c}{\textbf{HA-VLN-DE}} \\
\cmidrule(lr){2-3}\cmidrule(lr){4-7}
& \textbf{NE}$\downarrow$ & \textbf{SR}$\uparrow$
& \textbf{NE}$\downarrow$ & \textbf{TCR}$\downarrow$ & \textbf{CR}$\downarrow$ & \textbf{SR}$\uparrow$ \\
\midrule
Speaker-Follower~\citep{fried2018speaker} & 6.62 & 0.35 & 7.44 & 0.32 & \cellcolor{uwpurple!15}\textbf{0.72} & 0.21 \\
Rec (PREVALENT)~\citep{Hong_2021_CVPR}   & \cellcolor{uwpurple!15}\textbf{3.93} & \cellcolor{uwpurple!15}\textbf{0.63} & \underline{6.12} & \underline{0.29} & 0.81 & 0.33 \\
Rec (OSCAR)~\citep{Hong_2021_CVPR}        & 4.29 & 0.59 & \underline{6.12} & \cellcolor{uwpurple!15}\textbf{0.28} & \underline{0.78} & \underline{0.34} \\
Airbert~\citep{guhur2021airbert}          & \underline{4.01} & \underline{0.62} & \cellcolor{uwpurple!15}\textbf{5.54} & 0.30 & 0.83 & \cellcolor{uwpurple!15}\textbf{0.36} \\
NavCoT~\citep{lin2025navcot}         & 6.26 &0.40 & 6.83& 0.36 & 0.85& 0.23\\
\bottomrule[1pt]
\end{tabular}
}
\vspace{-0.4cm}
\end{wraptable}
\noindent \textbf{Experimental Setup.}~We evaluate agents in two settings: \textbf{(1)} We assess the performance of HA-VLN 2.0 agents alongside several top agents on the \textbf{HA-VLN 2.0 benchmark}, utilizing both \textbf{HA-VLN-CE (continuous)} and \textbf{HA-VLN-DE (discrete)} (Sec.~\ref{sec.exp-benchmark-havln2.0}). We conduct extensive analysis and ablation studies examining key factors including continuous versus discrete settings, cross-domain generalization capabilities, human presence and interaction enrichment, step size variations, and sensor modality configurations. These analyses investigate their respective impacts on human-aware navigation performance and reveal complementary knowledge between DE and CE approaches. \textbf{(2)} We deploy and evaluate HA-VLN 2.0 agents in real-world scenarios across diverse layouts (offices, living rooms, hallways, and lobbies) with free-moving human volunteers (Sec.~\ref{sec:maintext-validation-leaderboard}).

\vspace{-0.1cm}

\begin{table*}[!t]
\setlength{\abovecaptionskip}{1pt}
\setlength{\belowcaptionskip}{2pt}
\renewcommand{\arraystretch}{0.95}
  \centering
  \caption{\small
    \textbf{Cross Domain Evaluation of BEVBert (CE) vs.\ Rec (PREVALENT) (DE).}
    Each model is trained/validated under different simulators (HA-VLN-CE/HA-VLN-DE vs.~VLN-CE/VLN-DE) and different instruction sets (HA-R2R~vs.~R2R-CE/R2R).  
    The purple cells (\cellcolor{uwpurple!15}) indicate performance changes when models are trained on R2R-CE/R2R but validated on HA-R2R.
  }
  \label{tab:havln-cross eval}
  \begin{subtable}[t]{0.49\textwidth}
 \resizebox{1\linewidth}{!}{%
    \begin{tabular}{ccccccc}
      \toprule[1.2pt]
      \multirow{2}{*}{\textbf{Env}} 
      & \multicolumn{2}{c}{\textbf{Training}} 
      & \multicolumn{2}{c}{\textbf{Validation}} 
      & \multicolumn{2}{c}{\textbf{Val (Unseen)}} \\ 
      \cmidrule(lr){2-3} \cmidrule(lr){4-5} \cmidrule(lr){6-7}
      & \textbf{Simulator} & \textbf{Instr.} 
      & \textbf{Simulator} & \textbf{Instr.} 
      & \textbf{NE}$\downarrow$ & \textbf{SR}$\uparrow$ \\
      \midrule
      \multirow{6}{*}{\textbf{CE}} 
     & VLN-CE&R2R-CE&\multirow{2}{*}{VLN-CE}&R2R-CE&4.57&0.37\\
     & HA-VLN-CE &HA-R2R&&R2R-CE&\cellcolor{uwpurple!15}5.11&\cellcolor{uwpurple!15}0.35\\\cdashline{2-7}
      &\multirow{4}{*}{HA-VLN-CE}&HA-R2R&&R2R-CE&4.35&0.27\\
      &&R2R-CE&&R2R-CE&4.13&0.29\\
        & & HA-R2R  &  & HA-R2R  & 5.51 & 0.21 \\  
      & & R2R-CE & \multirow{-4}{*}{HA-VLN-CE}& HA-R2R & \cellcolor{uwpurple!15}6.23 \scriptsize($\uparrow$13.1\%) 
        & \cellcolor{uwpurple!15}0.15 \scriptsize($\downarrow$28.6\%) \\
        \bottomrule[1.2pt]
\end{tabular}}
\end{subtable}
\begin{subtable}[t]{0.495\textwidth}
 \resizebox{1\linewidth}{!}{%
    \begin{tabular}{ccccccc}
      \toprule[1.2pt]
      \multirow{2}{*}{\textbf{Env}} 
      & \multicolumn{2}{c}{\textbf{Training}} 
      & \multicolumn{2}{c}{\textbf{Validation}} 
      & \multicolumn{2}{c}{\textbf{Val (Unseen)}} \\ 
      \cmidrule(lr){2-3} \cmidrule(lr){4-5} \cmidrule(lr){6-7}
      & \textbf{Simulator} & \textbf{Instr.} 
      & \textbf{Simulator} & \textbf{Instr.} 
      & \textbf{NE}$\downarrow$ & \textbf{SR}$\uparrow$ \\      \midrule
      \multirow{6}{*}{\textbf{DE}}&VLN-DE & R2R& \multirow{2}{*}{VLN-DE}& R2R&3.93 &0.48\\
      & HA-VLN-DE & R2R& & R2R&\cellcolor{uwpurple!15}4.62&\cellcolor{uwpurple!15}0.45\\\cdashline{2-7}
      & \multirow{4}{*}{HA-VLN-DE} & HA-R2R& \multirow{4}{*}{HA-VLN-DE}& R2R&5.86&0.36\\
      & & R2R& & R2R&5.21&0.33\\
      && HA-R2R  &  & HA-R2R  & 5.01 & 0.39 \\
      & & R2R & & HA-R2R & \cellcolor{uwpurple!15}6.11 \scriptsize($\uparrow$22.0\%) & \cellcolor{uwpurple!15}0.24 \scriptsize($\downarrow$38.5\%) \\
      \bottomrule[1.2pt]
    \end{tabular}
      }
    \end{subtable}
    \vspace{-0.15cm}
\end{table*}
\begin{table*}[!t]
\setlength{\abovecaptionskip}{0cm}
\setlength{\belowcaptionskip}{0cm}
  \caption{\small
    \textbf{Left: (a).~Impact of Human Presence (hp) and Interaction Enrichment (enrich).}
    We evaluate without hp (replace human with cylinders) and without enrich (skip interaction \& movement enrichment in Sec.~\ref{sec:havln-sim}) on both CE and DE settings. 
    \small
    \textbf{Right: (b).~Impact of Step Size on Navigation.} Here the collision is detected only at endpoint of a step, thus increasing step size transitions from finer-grained control to more discrete (teleport-potential) steps (default step size for CE is 0.25m). 
    We show results for both BEVBert~\citep{an2023bevbert} and ETPNav~\citep{an2024etpnav} on seen/unseen. Best results for each metric are highlighted in \textbf{bold}.}
\begin{subtable}[t]{0.52\textwidth}
  \resizebox{\linewidth}{!}{%
    \begin{tabular}{ccclrrrr}
      \toprule[1.2pt]
      \textbf{hp} & \textbf{enrich} & \textbf{Env} & \textbf{Agent} 
      & \textbf{NE}$\downarrow$ & \textbf{TCR}$\downarrow$ & \textbf{CR}$\downarrow$ & \textbf{SR}$\uparrow$ \\
      \midrule
      \multirow{3}{*}{\textcolor{uwpurple}{\ding{51}}} 
      & \multirow{3}{*}{\textcolor{uwpurple}{\ding{51}}} 
      & CE & BEVBert 
      & 6.10 & 5.72 & 0.56 & 0.15 \\
      & & CE & ETPNav
      & 7.40 & 7.94 & 0.71 & 0.08 \\
      & & DE & Rec (PRE)
      & 7.31 & 0.31 & 0.79 & 0.22 \\
      \midrule
      \multirow{3}{*}{\textcolor{uwpurple}{\ding{51}}} 
      & \multirow{3}{*}{\textcolor{red}{\ding{55}}} 
      & CE & BEVBert
      & \cellcolor{uwpurple!15}6.32 \scriptsize($\uparrow$3.6\%)
      & \cellcolor{uwpurple!15}5.11 \scriptsize($\downarrow$10.7\%)
      & \cellcolor{uwpurple!15}0.46 \scriptsize($\downarrow$17.9\%)
      & \cellcolor{uwpurple!15}0.17 \scriptsize($\uparrow$13.3\%) \\
      & & CE & ETPNav
      & \cellcolor{uwpurple!15}7.35 \scriptsize($\downarrow$0.6\%)
      & \cellcolor{uwpurple!15}6.12 \scriptsize($\downarrow$22.9\%)
      & \cellcolor{uwpurple!15}0.63 \scriptsize($\downarrow$11.3\%)
      & \cellcolor{uwpurple!15}0.10 \scriptsize($\uparrow$25.0\%) \\
      & & DE & Rec (PRE)
      & \cellcolor{uwpurple!15}7.52 \scriptsize($\uparrow$2.9\%)
      & \cellcolor{uwpurple!15}0.27 \scriptsize($\downarrow$12.9\%)
      & \cellcolor{uwpurple!15}0.64 \scriptsize($\downarrow$19.0\%)
      & \cellcolor{uwpurple!15}0.27 \scriptsize($\uparrow$22.7\%) \\
      \midrule
      \multirow{3}{*}{\textcolor{red}{\ding{55}}} 
      & \multirow{3}{*}{\textcolor{red}{\ding{55}}} 
      & CE & BEVBert
      & \cellcolor{uwpurple!15}6.13 \scriptsize($\uparrow$0.5\%)
      & \cellcolor{uwpurple!15}3.25 \scriptsize($\downarrow$43.2\%)
      & \cellcolor{uwpurple!15}0.35 \scriptsize($\downarrow$37.5\%)
      & \cellcolor{uwpurple!15}0.19 \scriptsize($\uparrow$26.7\%) \\
      & & CE & ETPNav
      & \cellcolor{uwpurple!15}7.75 \scriptsize($\uparrow$4.7\%)
      & \cellcolor{uwpurple!15}4.47 \scriptsize($\downarrow$43.7\%)
      & \cellcolor{uwpurple!15}0.53 \scriptsize($\downarrow$25.4\%)
      & \cellcolor{uwpurple!15}0.14 \scriptsize($\uparrow$75.0\%) \\
      & & DE & Rec (PRE)
      & \cellcolor{uwpurple!15}7.33 \scriptsize($\uparrow$0.3\%)
      & \cellcolor{uwpurple!15}0.19 \scriptsize($\downarrow$38.7\%)
      & \cellcolor{uwpurple!15}0.42 \scriptsize($\downarrow$46.8\%)
      & \cellcolor{uwpurple!15}0.26 \scriptsize($\uparrow$18.2\%) \\
      \bottomrule[1.2pt]
    \end{tabular}
  }
\end{subtable}
\hspace{\fill}
\begin{subtable}[t]{0.47\textwidth}
  \resizebox{\linewidth}{!}{%
    \begin{tabular}{lcccccccccc}
      \toprule[1.2pt]
      \multirow{2}{*}{\textbf{Agent}}
      & \multirow{2}{*}{\textbf{Step Size}}
      & \multicolumn{4}{c}{\textbf{Validation (Seen)}}
      & \multicolumn{4}{c}{\textbf{Validation (Unseen)}} \\
      \cmidrule(lr){3-6} \cmidrule(lr){7-10}
      & & \textbf{NE}$\downarrow$ & \textbf{TCR}$\downarrow$ & \textbf{CR}$\downarrow$ & \textbf{SR}$\uparrow$
        & \textbf{NE}$\downarrow$ & \textbf{TCR}$\downarrow$ & \textbf{CR}$\downarrow$ & \textbf{SR}$\uparrow$ \\
      \midrule
      \multirow{5}{*}{\textbf{BEVBert}}
        & 0.10 
          & 5.65 & 8.43 & 0.50 & 0.23
          & \textbf{5.41} & 12.60 & 0.54 & 0.22 \\
        & \cellcolor{uwpurple!15} 0.25 (CE Default)
          & \textbf{5.53} & 3.64 & 0.46 & 0.27
          & 5.51 & 4.71 & 0.55 & 0.21 \\
        & 0.40
          & 5.60 & 1.77 & 0.39 & 0.28
          & 5.63 & 2.63 & 0.44 & 0.25 \\
        & 1.00
          & 5.82 & 0.42 & 0.21 & \textbf{0.29}
          & 5.54 & 0.63 & 0.26 & \textbf{0.26} \\
        & 2.25
          & 7.66 & \textbf{0.09} & \textbf{0.10} & 0.03
          & 7.23 & \textbf{0.10} & \textbf{0.10} & 0.03 \\
      \midrule
      \multirow{5}{*}{\textbf{ETPNav}}
        & 0.10
          & 5.15 & 11.70 & 0.54 & 0.20
          & 5.47 & 18.66 & 0.64 & 0.16 \\
        & \cellcolor{uwpurple!15}0.25 (CE Default)
          & 5.17 & 4.07 & 0.43 & 0.24
          & 5.43 & 6.94 & 0.58 & 0.17 \\
        & 0.40
          & \textbf{5.11} & 2.43 & 0.36 & \textbf{0.26}
          & \textbf{5.32} & 3.77 & 0.46 & \textbf{0.21} \\
        & 1.00
          & 6.67 & 0.49 & 0.25 & 0.24
          & 6.76 & 0.79 & 0.32 & 0.17 \\
        & 2.25
          & 7.61 & \textbf{0.10} & \textbf{0.10} & 0.02
          & 7.21 & \textbf{0.13} & \textbf{0.12} & 0.03 \\
      \bottomrule[1.2pt]
    \end{tabular}
  }
\end{subtable}
 \vspace{-0.2cm}
\label{tab-main-3}
\end{table*}

\subsection{Benchmarking Agents on HA-VLN 2.0}
\label{sec.exp-benchmark-havln2.0}
\noindent \textbf{HA-VLN-CE.}~We systematically benchmark several SOTA navigation agents, BEVBert~\citep{an2023bevbert}, ETPNav~\citep{an2024etpnav}, NaVid~\citep{zhang2024navid}, and Navila~\citep{cheng2025navila}, together with our HA-VLN-CMA and HA-VLN-VL agents in Table~\ref{tab:comparision_a}. Each approach is trained/evaluated under two configurations:  
\textbf{Retrained}, where agents are trained/evaluated solely on HA-VLN-CE benchmark (HA-VLN-CE simulator + HA-R2R instruction dataset), and \textbf{Zero-shot}, where agents are trained solely on VLN-CE benchmark (VLN-CE simulator + R2R-CE) and evaluated on our benchmark.
Table~\ref{tab:comparision_a} shows pronounced gains when models incorporate HA-VLN-CE benchmark. For instance, BEVBert’s SR increases from 0.19 to 0.27 in seen split and from 0.15 to 0.21 in unseen. In contrast, Table~\ref{tab:havln-cross eval} shows that BEVBert trained on our benchmark performs comparably to the VLN-CE-trained one on VLN-CE benchmark (SR: 0.35 vs.\ 0.37). This bidirectional evaluation suggests that explicit references to dynamic crowd behavior enhance real-world navigational readiness and confirm the robustness of HA-VLN-CE.
Fig.~\ref{fig:appx_visual_2} presents navigation visualizations of the HA-VLN-CMA$^{*}$ agent on HA-VLN-CE benchmark, including one successful and one failed example. These examples demonstrate that dynamic human activities indeed increase the difficulty of navigation, while also making the scenarios more realistic and reflective of real-world challenges.
\begin{wraptable}{r}{10cm}
\vspace{-0.3cm}
  \centering
  \caption{\small
    \textbf{Ablation on RGB/Depth Inputs.}
    We compare BEVBert~\citep{an2023bevbert} and ETPNav~\citep{an2024etpnav} on unseen validations. 
    \textcolor{uwpurple}{\ding{51}} denotes the sensor is enabled, while \textcolor{red}{\ding{55}} is disabled.
    Purple cells highlight performance changes (in \%) upon removing/adding a modality.
  }
  \label{tab:apx_depth}
  \vspace{-0.05in}
  \resizebox{1\linewidth}{!}{%
  \begin{tabular}{lcccccc}
    \toprule[1.2pt]
    \multirow{2}{*}{\textbf{Agent}} 
      & \multirow{2}{*}{\textbf{RGB}} 
      & \multirow{2}{*}{\textbf{Depth}} 
      & \multicolumn{4}{c}{\textbf{Validation (Unseen)}} \\
    \cmidrule(lr){4-7}
    & & 
    & \textbf{NE}$\downarrow$ & \textbf{TCR}$\downarrow$ & \textbf{CR}$\downarrow$ & \textbf{SR}$\uparrow$ \\
    \midrule
    \multirow{3}{*}{\textbf{BEVBert}\,\citep{an2023bevbert}}
      & \textcolor{uwpurple}{\ding{51}} & \textcolor{red}{\ding{55}}
      & \cellcolor{uwpurple!15}5.79 \scriptsize($\uparrow$5.1\%)
      & \cellcolor{uwpurple!15}4.97 \scriptsize($\uparrow$5.5\%)
      & \cellcolor{uwpurple!15}0.53 \scriptsize($\uparrow$3.6\%)
      & \cellcolor{uwpurple!15}0.15 \scriptsize($\downarrow$28.6\%) \\
    & \textcolor{red}{\ding{55}} & \textcolor{uwpurple}{\ding{51}}
      & \cellcolor{uwpurple!15}5.50 \scriptsize($\uparrow$0.2\%)
      & \cellcolor{uwpurple!15}4.73 \scriptsize($\uparrow$0.4\%)
      & \cellcolor{uwpurple!15}0.53 \scriptsize($\uparrow$3.6\%)
      & \cellcolor{uwpurple!15}0.20 \scriptsize($\downarrow$4.8\%) \\
    & \textcolor{uwpurple}{\ding{51}} & \textcolor{uwpurple}{\ding{51}}
      & 5.51 & 4.71 & 0.55 & 0.21 \\
    \midrule
    \multirow{3}{*}{\textbf{ETPNav}\,\citep{an2024etpnav}}
      & \textcolor{uwpurple}{\ding{51}} & \textcolor{red}{\ding{55}}
      & \cellcolor{uwpurple!15}6.38 \scriptsize($\uparrow$17.5\%)
      & \cellcolor{uwpurple!15}7.44 \scriptsize($\uparrow$7.2\%)
      & \cellcolor{uwpurple!15}0.65 \scriptsize($\uparrow$12.1\%)
      & \cellcolor{uwpurple!15}0.13 \scriptsize($\downarrow$23.5\%) \\
    & \textcolor{red}{\ding{55}} & \textcolor{uwpurple}{\ding{51}}
      & \cellcolor{uwpurple!15}5.94 \scriptsize($\uparrow$9.4\%)
      & \cellcolor{uwpurple!15}7.23 \scriptsize($\uparrow$4.2\%)
      & \cellcolor{uwpurple!15}0.65 \scriptsize($\uparrow$12.1\%)
      & \cellcolor{uwpurple!15}0.16 \scriptsize($\downarrow$5.9\%) \\
    & \textcolor{uwpurple}{\ding{51}} & \textcolor{uwpurple}{\ding{51}}
      & 5.43 & 6.94 & 0.58 & 0.17 \\
    \bottomrule[1.2pt]
  \end{tabular}
  }
\vspace{-0.3cm}
\end{wraptable}
\noindent \textbf{HA-VLN-DE.}~Table~\ref{tab:ha-vln-de} compares top discrete agents on both VLN and HA-VLN-DE benchmarks, showing that discrete agents can achieve moderate SR yet suffer high collisions in crowded scenes. For example, while Airbert~\citep{guhur2021airbert} achieves a moderate SR at 0.36, it can incur a CR of up to 0.83, illustrating persistent collision risks. Results showcase that adaptive collision-avoidance strategies remain essential in discrete settings. Approaches that overlook human dynamics often fail when multiple bystanders converge, particularly in tight junctions or doorways.

\noindent \textbf{Analysis \& Ablation~Studies.}
\textit{\textbf{(1)~Cross-domain Generalization.}} Table~\ref{tab:havln-cross eval} reveals that HA-R2R-trained agents achieve comparable SR to R2R-CE-trained agents (0.27 vs.~0.29) on R2R-CE validation set, while they outperform by +28.6\% SR on the HA-R2R validation set, showcasing HA-R2R improves in-domain performance while maintaining cross-domain robustness.
\textit{\textbf{(2)~Human~Presence and Interaction Enrichment.}}  
Table~\ref{tab-main-3} (a) shows in human presence ablations, replacing humans with cylinders drops TCR by around 36\% and raises SR by around 10\%, while removing human interaction enrichment drops TCR by up to 22\% and raises SR by up to 25\%, confirming humans are not merely treated as generic moving obstacles during navigation.
\textit{\textbf{(3)~Step~Size.}}  
Table~\ref{tab-main-3} (b) indicates a degree of knowledge complementarity between DE and CE navigation when collisions are detected only at the endpoint of a step. Specifically, increasing the step size (from 0.1\,m to 1.0\,m), approximating DE-style navigation, can improve performance.  
We also conducted an additional experiment in which a 1.0\,m step was treated as four 0.25\,m sub-steps, and a 2.25\,m step as nine 0.25\,m sub-steps, with collisions checked after each sub-step.  
When evaluated on BEVBert in the val\_unseen split, the agents failed to navigate effectively with both 1.0\,m and 2.25\,m step sizes, with SR dropping close to zero. These results highlight the need to account for the potentially “teleport-like” movement behaviors in DE when considering complementarity.  
\textit{\textbf{(4)~Sensor~Modalities.}}  
Table~\ref{tab:apx_depth} confirms that either adding depth or RGB consistently lowers collisions and raises SR, reflecting the importance of 3D cues for navigating around moving bystanders.

\begin{figure}[!t]
\renewcommand{\arraystretch}{0.95}
\setlength{\abovecaptionskip}{0pt}
\setlength{\belowcaptionskip}{0pt}
    \centering
    \includegraphics[width=0.999\linewidth]{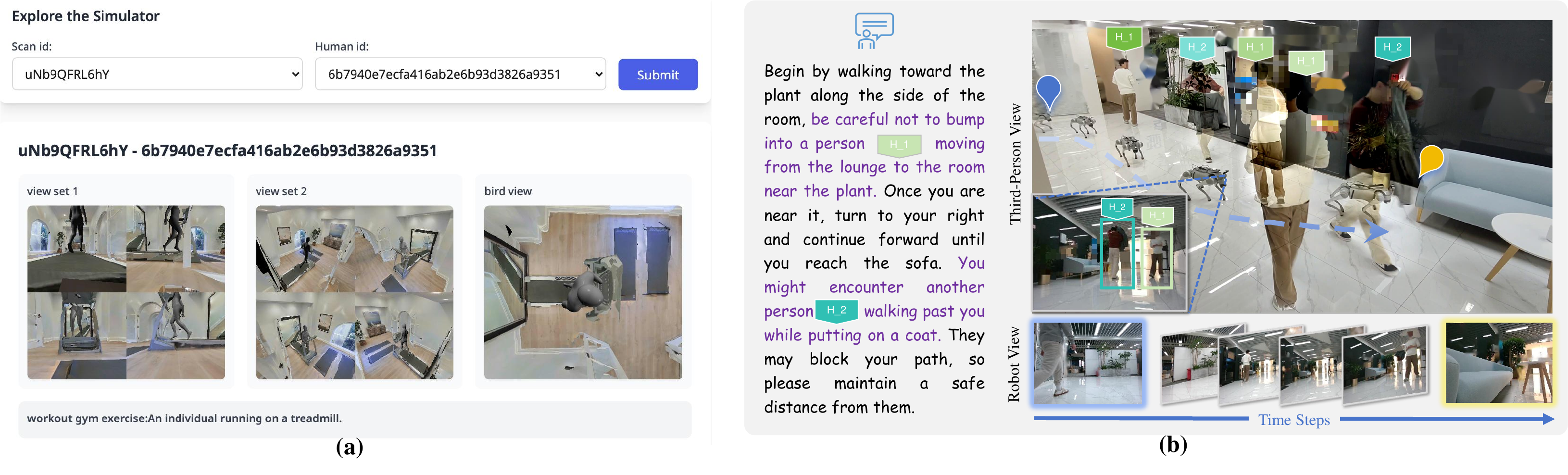}
    \caption{\textbf{(a).~Interactive interface} we provide to explore 910 annotated human models and surrounding environments in the HA-VLN 2.0 simulator from nine views.~\textbf{(b).~Human-aware navigation with multiple bystanders.}  
    \emph{Left:} Instruction provided to the robot.  
    \emph{Right:} A third-person view illustrates the robot’s trajectory among dynamic bystanders, and selected robot observations from the onboard camera.}
    \label{fig:interactive interface plus real world exp}
    \vspace{-0.2cm}
\end{figure}

\subsection{Leaderboard~\& Real-World Validation}
\label{sec:maintext-validation-leaderboard}

\noindent \textbf{HA-R2R~Test~Dataset~\& Leaderboard.}  
Building on R2R-CE, we present HA-R2R, featuring 16,844 instructions across 90 building scans with 910 annotated human models (Secs.~\ref{sec:havln-sim} \&~\ref{sec:ha-r2r}).~While retaining path continuity from R2R-CE, we introduce refined goals to emphasize social awareness. The test partition of HA-R2R contains 3,408 instructions across 18 withheld buildings and intentionally emphasizes multi-human routes. To assess performance on this challenging test split, we host leaderboards for HA-VLN 2.0 benchmarks, evaluating both collision-related (TCR, CR) and navigation metrics (NE, SR). We also prepare an interactive demo interface shown in Figure~\ref{fig:interactive interface plus real world exp} (a) and also on our webpage, where participants can explore the simulator from nine different views to examine all the annotated human motions and surrounding environments. Submissions may include agent code or trajectories, providing reproducible, server-side evaluations and setting a new benchmark for human-centric, dynamic VLN research.

\noindent \textbf{Real-World Validation~\& Setup.}  
We deploy our trained agents on a \emph{Unitree Go2-EDU} quadruped, equipped with Intel~Realsense~D435i RGB-D camera, MID360~3D LiDAR, and IMU for onboard perception and control. As Fig.~\ref{fig:interactive interface plus real world exp} (b) illustrates, experiments are conducted in four indoor spaces (office, living room, hallway, lobby), each populated by 2--4 free-moving volunteers. The agent navigates safely in moderately congested conditions but faces challenges in tight corridors or sudden crowd convergence, highlighting the need for robust re-planning under partial observability. Please kindly refer to Appendix.~\ref{Appx.exp-all} for more experimental details.


%% file: sections/5_conclusion.tex
\section{Conclusion}

We present \textsc{\textbf{\textcolor{uwpurple}{HA}\textcolor{darkgray}{-}\textcolor{uwyellow}{VLN}~\textcolor{darkgray}{2.0}}}, a unified framework standardizing discrete and continuous VLN under explicit human-centric constraints.~By integrating multi-human interactions and high-fidelity simulators, our \emph{HA-R2R} dataset emphasizes human-aware instructions. Experiments reveal that social awareness, multi-human interactions, and partial observability increase task complexity, reducing the performance of advanced agents. Our approach balances safety and efficiency, with real-world tests confirming sim-to-real transfer and a public leaderboard enabling standardized evaluation. We release all resources to advance safe navigation research in dynamic, human-populated environments.

%% file: sections/appendix.tex
\section*{Appendix}
\noindent This supplementary material provides expanded details and results that complement the main paper.  
Section~\ref{Appx.related-work-all} offers a comprehensive literature survey focusing on three key research challenges.  
Section~\ref{Sec:appendixB} describes our dataset construction, annotation protocols, real-time rendering methods, API design, and additional insights on annotation data.  
Section~\ref{Appx.agent_all} presents an in-depth overview of the HA-R2R dataset and the proposed navigation agents.  
Finally, Section~\ref{Appx.exp-all} includes detailed evaluation metrics, additional numerical results, visualized navigation outcomes, and real-world robot validation studies, each supplemented with thorough analysis.
For further resources, access project page \url{https://uwmilab.github.io/HA-VLN-webpage}.  

\section{Related~Work}
\label{Appx.related-work-all}
This appendix surveys the evolution of Vision-and-Language Navigation~(VLN) tasks, simulators, and agent designs, with particular attention to how \emph{Human-Aware~VLN (HA-VLN) 2.0} advances the state of the art. We focus on three key aspects deemed critical for bridging the Sim2Real gap: \emph{(1)~Social Awareness}, \emph{(2)~Human-Aligned Instructions and Visual Cues}, and \emph{(3)~Dynamic Environments with Human Activities and Interactions}. In Table~\ref{tab:task_compare}, we summarize how prior work compares under these dimensions.

\subsection{Development~of~VLN~Tasks}
\label{Appx.related-work-task}
Early VLN tasks focused on basic indoor navigation, exemplified by Room-to-Room~(R2R)~\citep{anderson2018vision, fried2018speaker, gu2022vision, ku2020room}, and outdoor tasks like TOUCHDOWN~\citep{chen2019touchdown} and MARCO~\citep{macmahon2006walk}. Later efforts such as REVERIE~\citep{qi2020reverie} and VNLA~\citep{nguyen2019vision} introduced object-centric or goal-driven navigation. While these approaches expanded the range of tasks, they typically overlooked real human behavior and social contexts. Dialogue-based tasks (e.g., DialFRED~\citep{gao2022dialfred}, CVDN~\citep{thomason2020vision}) incorporated interactive elements but did not account for dynamically moving bystanders or social-distance constraints.
Initiatives like VLN-CE~\citep{krantz2020beyond} moved closer to real-world conditions by enabling continuous navigation, yet remained devoid of explicit human factors~\citep{jain2019stay, ku2020room, nguyen2019vision, thomason2020vision}. HA3D~\citep{li2024human} addressed human motion and included human-oriented instructions but did not require agents to conform to social norms (maintaining safe distances or refraining from disturbing ongoing activities).
Our proposed \emph{HA-VLN 2.0} addresses these gaps by embedding all three essential elements, socially compliant navigation, human-referenced instructions, and dynamic human activities, into a single framework. Agents must plan routes among unpredictable bystanders, interpret language mentioning people and their behaviors, and uphold social standards. This integrated setup results in a benchmark that closely aligns with real-world navigation demands.

\subsection{Simulators~for~VLN~Tasks}
\label{Appx.related-work-sim}
A reliable simulator is essential for developing and evaluating VLN agents. Early simulators like Matterport3D~\citep{anderson2018vision} and House3D~\citep{wu2018building} offered photorealistic or synthetic indoor environments but lacked mobile humans.~Others, such as AI2-THOR~\citep{kolve2017ai2} and Gibson~\citep{xia2018gibson}, introduced more interactive elements yet typically assumed static or purely synthetic contexts, thus limiting their applicability for studying social compliance. Google Street View, used in some outdoor navigation tasks, presents static imagery with occasional humans in the scene but lacks dynamic or interactive elements.~HA3D~\citep{li2024human} moved a step further by including human activities and instructions referencing people, though it did not mandate socially compliant navigation.~HabiCrowd~\citep{vuong2024habicrowd} integrated crowds into photorealistic domains, improving visual diversity but omitting human-aligned instructions. Similarly, recent works~\citep{savva2019habitat, lin2025vlnverse, wang2025rethinking} provide high-performance simulation without extensive multi-human or social-compliance features. By contrast, our \emph{HA-VLN Simulator} unifies dynamic human activities, photorealistic rendering, and social-compliance requirements. Agents perceive and react to evolving bystander behaviors like avoiding collisions or maintaining personal space using both discrete and continuous navigation.~Specifically, we introduce 675 scenes (across 90 scenarios), 122 motion types, and a cohesive framework that supports instruction-driven dynamic human interactions. By supporting both discrete and continuous action spaces, HA-VLN further broadens its potential for addressing diverse navigation goals~\citep{dong2025unified, dong2026language} and real-world deployment challenges.


\subsection{Agents~for~VLN~Tasks}
\label{Appx.related-work-agent}
From early attention-based and RL-based approaches~\citep{ma2019self, qi2020reverie, wang2019reinforced} to modern vision-language pre-training~\citep{lu2019vilbert, hao2020towards, li2020oscar}, VLN agents have grown increasingly adept at parsing instructions and navigating complex environments.~However, most existing solutions, including EnvDrop~\citep{envdrop}, PREVALENT~\citep{hao2020towards} and VLN-BERT~\citep{Hong_2021_CVPR}, rely on panoramic navigation, streamlining the action space but limiting realism of their movement.~Recent efforts like NavGPT~\citep{zhou2024navgpt} and NaVid~\citep{zhang2024navid} explore continuous, egocentric navigation in partially dynamic worlds, yet they still lack explicit attention to \emph{human-aligned} instructions or \emph{social compliance}.~In particular, these agents may not recognize the need to maintain safe distances, avoid disturbing activities, or adapt routes with active bystanders.
HA-VLN agents address these gaps by navigating among multiple, moving humans and adhering to social norms.~They interpret fine-grained, human-centric instructions and leverage visual cues that reflect real-world interactions, ensuring collision-free, respectful travel.~This fusion of social compliance and human dynamics sets HA-VLN apart, aligning agent behavior more closely with real-world challenges~\citep{dong2025securing, dong2025large}.

\begin{table*}[htbp]
\tiny
\centering
\caption{\small
Comparison of VLN tasks, simulators, and agents based on 
\textit{(1) Socially Compliant Navigation}, 
\textit{(2) Human-aligned Instructions and Visual Cues}, 
and \textit{(3) Dynamic Environments with Human Activities}.
}
\vspace{-0.1in}
\label{tab:task_compare}
\begin{tabular}{p{0.8cm}cccp{7.3cm}}
\toprule
& \begin{tabular}[c]{@{}c@{}}\textbf{Socially Compliant}\\ \textbf{Navigation}\end{tabular}
& \begin{tabular}[c]{@{}c@{}}\textbf{Human-aligned Instructions}\\\textbf{and Visual Cues}\end{tabular}
& \begin{tabular}[c]{@{}c@{}}\textbf{Dynamic}\\\textbf{Environments}\end{tabular}
& \multicolumn{1}{c}{\textbf{Prior Work}} \\
\midrule
\multirow{8}{*}{\textbf{Tasks}} 
& \multirow{5}{*}{\textcolor[rgb]{1,0,0}{\texttimes}} 
  & \multirow{5}{*}{\textcolor[rgb]{1,0,0}{\texttimes}} 
  & \multirow{5}{*}{\textcolor[rgb]{1,0,0}{\texttimes}} 
  & MARCO~\citep{macmahon2006walk}, DRIF~\citep{blukis2018mapping}, VLN-R2R~\citep{anderson2018vision}, TOUCHDOWN~\citep{chen2019touchdown}, REVERIE~\citep{qi2020reverie}, DialFRED~\citep{gao2022dialfred}\\
& & & & VNLA~\citep{nguyen2019vision}, CVDN~\citep{thomason2020vision}, R4R~\citep{jain2019stay}, RxR~\citep{ku2020room}, EQA~\citep{das2018embodied}, IQA~\citep{gordon2018iqa}\\
& \textcolor[rgb]{1,0,0}{\texttimes} & \textcolor[rgb]{1,0,0}{\texttimes} & \textcolor[rgb]{0,0,1}{\ding{51}} & VLN-CE~\citep{krantz2020beyond}, LH-VLN \citep{song2025towards}\\
& \textcolor[rgb]{1,0,0}{\texttimes} & \textcolor[rgb]{0,0,1}{\ding{51}} & \textcolor[rgb]{0,0,1}{\ding{51}} & HA3D~\citep{li2024human}, VLN-PE \citep{wang2025rethinking}\\
& \textcolor[rgb]{0,0,1}{\ding{51}} & \textcolor[rgb]{0,0,1}{\ding{51}} & \textcolor[rgb]{0,0,1}{\ding{51}} & \textbf{HA-VLN (Ours)}\\
\midrule
\multirow{5}{*}{\textbf{Simulators}}
& \textcolor[rgb]{1,0,0}{\texttimes} 
  & \textcolor[rgb]{1,0,0}{\texttimes} 
  & \textcolor[rgb]{1,0,0}{\texttimes} 
  & Matterport3D~\citep{anderson2018vision}, House3D~\citep{wu2018building}, AI2-THOR~\citep{kolve2017ai2}, Gibson GANI~\citep{xia2018gibson}\\
& \textcolor[rgb]{1,0,0}{\texttimes} & \textcolor[rgb]{1,0,0}{\texttimes} & \textcolor[rgb]{0,0,1}{\ding{51}} & Habitat~\citep{savva2019habitat}, Google Street, ViZDoom~\citep{kempka2016vizdoom}\\
& \textcolor[rgb]{1,0,0}{\texttimes} & \textcolor[rgb]{0,0,1}{\ding{51}} & \textcolor[rgb]{0,0,1}{\ding{51}} & HA3D~\citep{li2024human}, VLN-PE \citep{wang2025rethinking}, VLNVerse~\citep{lin2025vlnverse}\\
& \textcolor[rgb]{0,0,1}{\ding{51}} & \textcolor[rgb]{0,0,1}{\ding{51}} & \textcolor[rgb]{0,0,1}{\ding{51}} & \textbf{HA-VLN (Ours)}, Habitat3.0~\citep{puig2023habitat}\\
\midrule
\multirow{7}{*}{\textbf{Agents}}
& \multirow{4}{*}{\textcolor[rgb]{1,0,0}{\texttimes}} 
  & \multirow{4}{*}{\textcolor[rgb]{1,0,0}{\texttimes}} 
  & \multirow{4}{*}{\textcolor[rgb]{1,0,0}{\texttimes}} 
  & EnvDrop~\citep{envdrop}, AuxRN~\citep{zhu2020vision}, PREVALENT~\citep{hao2020towards}, RelGraph~\citep{hong2020language}, HAMT~\citep{chen2021history}, NavCoT~\citep{lin2025navcot}\\
& & & & Rec-VLNBERT~\citep{Hong_2021_CVPR}, EnvEdit~\citep{li2022envedit}, Airbert~\citep{guhur2021airbert}, Lily~\citep{lin2023learning}, ScaleVLN~\citep{wang2023scaling}, ETPNav \citep{an2024etpnav}, BEVBert \citep{an2023bevbert}\\
& \textcolor[rgb]{0,0,1}{\ding{51}} & \textcolor[rgb]{1,0,0}{\texttimes} & \textcolor[rgb]{0,0,1}{\ding{51}} & Student Force~\citep{anderson2018vision}, NavGPT~\citep{zhou2024navgpt}, NaVid~\citep{zhang2024navid}, Navila~\citep{cheng2025navila} \\
& \textcolor[rgb]{0,0,1}{\ding{51}} & \textcolor[rgb]{0,0,1}{\ding{51}} & \textcolor[rgb]{0,0,1}{\ding{51}} & \textbf{HA-VLN Agent (Ours)}\\
\bottomrule
\end{tabular}%
\vspace{-0.1in}
\end{table*}

\begin{table*}[t]
\setlength{\abovecaptionskip}{1pt}
\setlength{\belowcaptionskip}{-1pt}
\caption{\small
\textbf{Comparison of HAPS~1.0 vs.\ HAPS~2.0.}
We show the total number of motion categories, average \emph{accuracy} and \emph{compatibility} scores (both on a 1--10 scale), number of failure cases (e.g., motion-description mismatches), and annotation time. HAPS~2.0 features more diverse motions, improved motion-env alignment, 
and reduced failures with higher annotation effort.}
    \centering
    \resizebox{1\linewidth}{!}{
    \begin{tabular}{c|c|c|c|c|c}
    \toprule[1.2pt]
         Datasets & Motions $\uparrow$ & Accuracy (1-10) $\uparrow$ & Compatibility (1-10) $\uparrow$ & Failure Cases $\downarrow$ & Annotation Time (hours)  \\\hline
         HAPS 1.0~\citep{li2024human} & 435 & 6.3 & 5.9 & 120 & 320 (verified by \citep{li2024human}) \\
         HAPS 2.0 (ours)   & 486     &  8.5    & 8.1    & 0    & 430+    \\
             \bottomrule[1.2pt]
    \end{tabular}}
    \vspace{-0.1in}
    \label{tab:comparison_metrics}
\end{table*}

\section{Simulator~Details}
\label{Sec:appendixB}

\subsection{HAPS~Dataset~2.0}
\label{Appx.coarse2fine-all}
\noindent We develop HAPS~2.0 to address the shortcomings of its predecessor~\citep{li2024human}, particularly in terms of mismatches between textual descriptions and motion data, and limited diversity of region--motion associations.

\noindent \textbf{Motion--Description~Alignment.}~The original HAPS dataset contains 435 motion categories, each defined by a region (e.g., \emph{hallway}) and a textual description (e.g., “Someone talking on the phone while pacing”). However, more than half of these pairs do not match accurately. We therefore conduct a two-round manual verification, where multiple volunteers determine whether each pair is valid. Motions that fail both rounds are removed, yielding 172 precisely aligned motions.

\noindent \textbf{Diversifying~Region--Motion~Relationships.}~In the initial dataset, each region was tied to only a few rigidly defined motions (e.g., \emph{hallway} mostly involves “pacing on a phone,” \emph{stairs} focuses on “sliding down a banister” or “decorating the stairway”). Such narrow mappings cause biases and limit the realism of agent navigation. To remedy this, we reorganize region--motion associations, adapting the same motion to fit various environments, including both indoor and outdoor scenes. For instance, “talking on the phone” is re-contextualized to reflect whether someone is pacing upstairs or moving around a meeting room. This broader approach offers more faithful representations of human behavior and reduces environmental biases, thus improving real-world applicability.

\noindent \textbf{HAPS~2.0~vs.~HAPS~1.0.}~Table~\ref{tab:comparison_metrics} quantitatively contrasts HAPS~2.0 with HAPS~1.0. We recruit 26 volunteers to evaluate every motion in both datasets on two 1--10 scales (\emph{motion~accuracy}, \emph{motion--environment~compatibility}). A motion is deemed a failure if it scores under 3 in either category or below 5 in both. As shown, HAPS~2.0 achieves higher accuracy (8.5~vs.~6.3), better compatibility (8.1~vs.~5.9), and zero failures (0~vs.~120). It also increases motion diversity (486~vs.~435) and overall annotation effort (430+~vs.~320 hours). Moreover, HAPS~2.0 refines annotation workflows and simulator design for enhanced generalization.

Altogether, HAPS~2.0 includes 26 distinct regions across 90 architectural scenes, covering 486 human activities in both indoor and outdoor contexts. Fig.~\ref{fig:app_overview} illustrates these improvements. By offering more accurate, flexible, and diverse depictions of human actions, HAPS~2.0 provides a robust foundation for research in human motion analysis, social navigation, and beyond.

During dataset construction, we checked that the included visual and textual data do not contain names, faces, or other information that could uniquely identify individuals. Offensive or harmful content was not observed.

\subsection{Coarse~Annotation~Using~PSO}
\label{Appx.coarse2fine-pso}
\noindent We adopt a coarse-to-fine strategy for positioning human motions in 3D scans.~Initially, we define each region by boundary coordinates 
\(\mathbf{B}_{\text{lo}}=(x_{\text{lo}}, y_{\text{lo}}, z_{\text{lo}})\),
\(\mathbf{B}_{\text{hi}}=(x_{\text{hi}}, y_{\text{hi}}, z_{\text{hi}})\),
and compile an object list 
\(\mathbf{O} = \{j_1, j_2, \dots, j_n\}\) 
with positions \(\mathbf{p}^{j_i}\).
We then use Particle~Swarm~Optimization~(PSO)~\citep{kennedy1995particle} (more details are provided in Algorithm~\ref{alg:human_motion_coarse_PSO}) to locate each motion \(h_i\) at an optimal position \(\mathbf{p}^{opt}\).  

\noindent \textbf{Safe Distance Constraint.}~We set \(\epsilon=1\,\mathrm{m}\) as the minimum clearance between humans and objects, ensuring a realistic layout while leaving space for agent passage.  

\noindent \textbf{Adaptive Penalties.}~Our fitness function applies penalties to placements that violate constraints (e.g., intersecting walls or overlapping humans). This strategy discourages infeasible poses and promotes plausible scene geometry alignments.  
The resulting coarse alignment establishes a starting point, after which we apply finer manual or semi-automated adjustments to refine multi-human interactions and ensure consistent coverage of diverse motion types.

\begin{algorithm}[!t]
\caption{Coarse Annotation via PSO}
\label{alg:human_motion_coarse_PSO}
\begin{algorithmic}[1]
\REQUIRE Region $\mathbf{R}$~$\!\leftarrow\!\langle\mathbf{r}, \mathbf{B}_{\text{lo}}, \mathbf{B}_{\text{hi}}\rangle$, where $\mathbf{r}$ is region label and boundary coordinates $\mathbf{B}_{\text{lo}}\!=\!(x_{\text{lo}}, y_{\text{lo}}, z_{\text{lo}})$ and $\mathbf{B}_{\text{hi}}\!=\!(x_{\text{hi}}, y_{\text{hi}}, z_{\text{hi}})$; object list $\mathbf{O} \leftarrow \{ j_1, j_2, \ldots, j_n \}$ with positions $\mathbf{p}_{j_i} \leftarrow (x_{j_i}, y_{j_i}, z_{j_i})$; human motion set $\mathbf{H}$; minimum safe distance $\epsilon \leftarrow 1~\text{m}$; height offset $\Delta_z \leftarrow 0.75~\text{m}$.

\ENSURE Final positions $\mathbf{p}^{h} \leftarrow (x_h, y_h, z_h)$ for each human motion $h \in \mathbf{H}$.

\WHILE{not all human motions placed}
    \STATE Filter human motions $\mathbf{H}' \subseteq \mathbf{H}$ matching $\mathbf{r}$;
    \STATE Match objects $\mathbf{O}$ with human motions $\mathbf{H}'$ based on semantic similarity to form pairs $(h_i, j_i)$;
    
    \FOR{each pair $(h_i, j_i)$}
        \STATE Define search space $\mathbf{S} \leftarrow \left\langle x_{lo}, x_{hi} \right\rangle \times \left\langle z_{lo}, z_{hi} \right\rangle \times \left\langle y_{lo}, y_{hi} \right\rangle $ around object $j_i$;
        \STATE Initialize PSO with particles randomly positioned within $\mathbf{S}$;
        \STATE Convergence criteria $\leftarrow$ minimal fitness change;
        
        \REPEAT
            \FOR{each particle $p$ in the swarm}
                \STATE Compute position $\mathbf{p}^{h}$ of particle $p$;
                \STATE Compute fitness $f(p)$;
                \STATE $f(p) \leftarrow d(\mathbf{p}^{h}, \mathbf{p}^{j_i}) + P_{\text{constraints}}(p)$;
                \STATE where $d(\mathbf{p}^h, \mathbf{p}^{j_i})$ is the Euclidean distance, and $P_{\text{constraints}}(p)$ is the penalty for constraint violations;
                
                \STATE \textbf{Constraints}:
        
                \STATE $d(\mathbf{p}^{h}, \mathbf{p}^{j_i}) \leq 1~\text{m}$; \hfill (Proximity to target object)
                
                \STATE $d(\mathbf{p}^{h}, \mathbf{p}^{j_u}) \geq \epsilon$, $\forall j_u \in \mathbf{O}, j_u \neq j_i$; \hfill (Maintain safe distance from other objects)
                
                \STATE $\mathbf{p}^{h} \in \mathbf{R}$; \hfill (Within region boundaries)
                
                \STATE Optional: $z_h \geq z_{j_i} + \Delta_z$; \hfill (Height offset)
                
            \ENDFOR
            \STATE Update particle velocities and positions using PSO update equations;
        \UNTIL{convergence criteria met}
        
        \STATE Assign best particle position $\mathbf{p}^{h}$ to $h_i$;
        
        \IF{no feasible solution found}
            \STATE Adjust PSO parameters and retry;
        \ENDIF
    \ENDFOR
\ENDWHILE
\end{algorithmic}
\end{algorithm}

\begin{figure*}[t]
  \centering
  \setlength{\abovecaptionskip}{1pt}
  \setlength{\belowcaptionskip}{-1pt}
  \includegraphics[width=\linewidth]{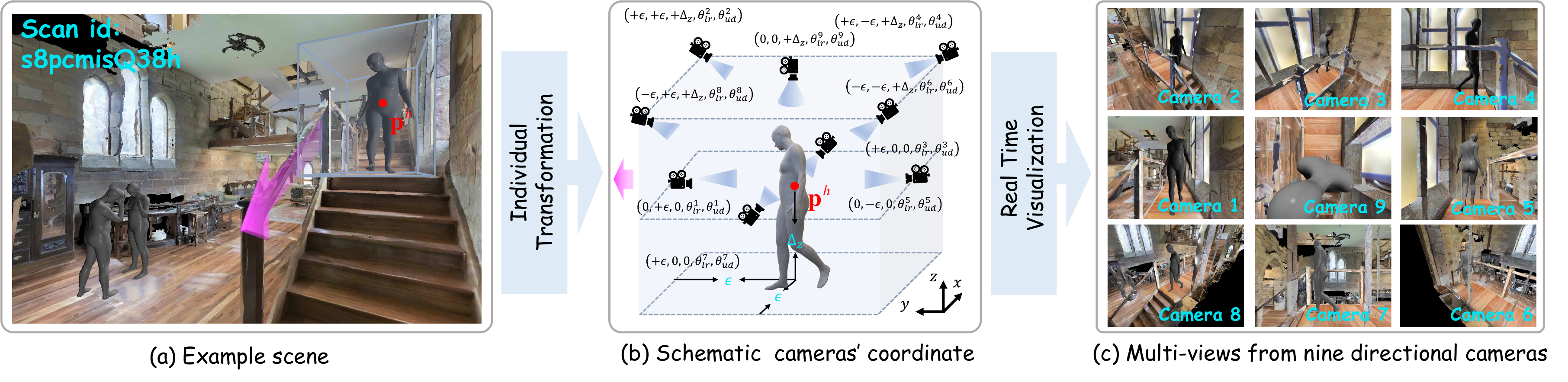}
    \vspace{-0.1in}
  \caption{\small
  \textbf{Multi-View Camera Setup.}
  \textbf{(a)}~A sample scene overview. 
  \textbf{(b)}~Schematic illustrating the nine camera placements around the human figure, noting key coordinates and rotations.
  \textbf{(c)}~Example snapshots from the nine directional cameras, each providing a distinct viewpoint for accurate motion annotation.
  }
  \label{app:fine_annonation}
  \vspace{-0.1in}
\end{figure*}

\begin{figure*}[t]
  \centering
  \setlength{\abovecaptionskip}{-1pt}
  \setlength{\belowcaptionskip}{-1pt}
  \includegraphics[width=\linewidth]{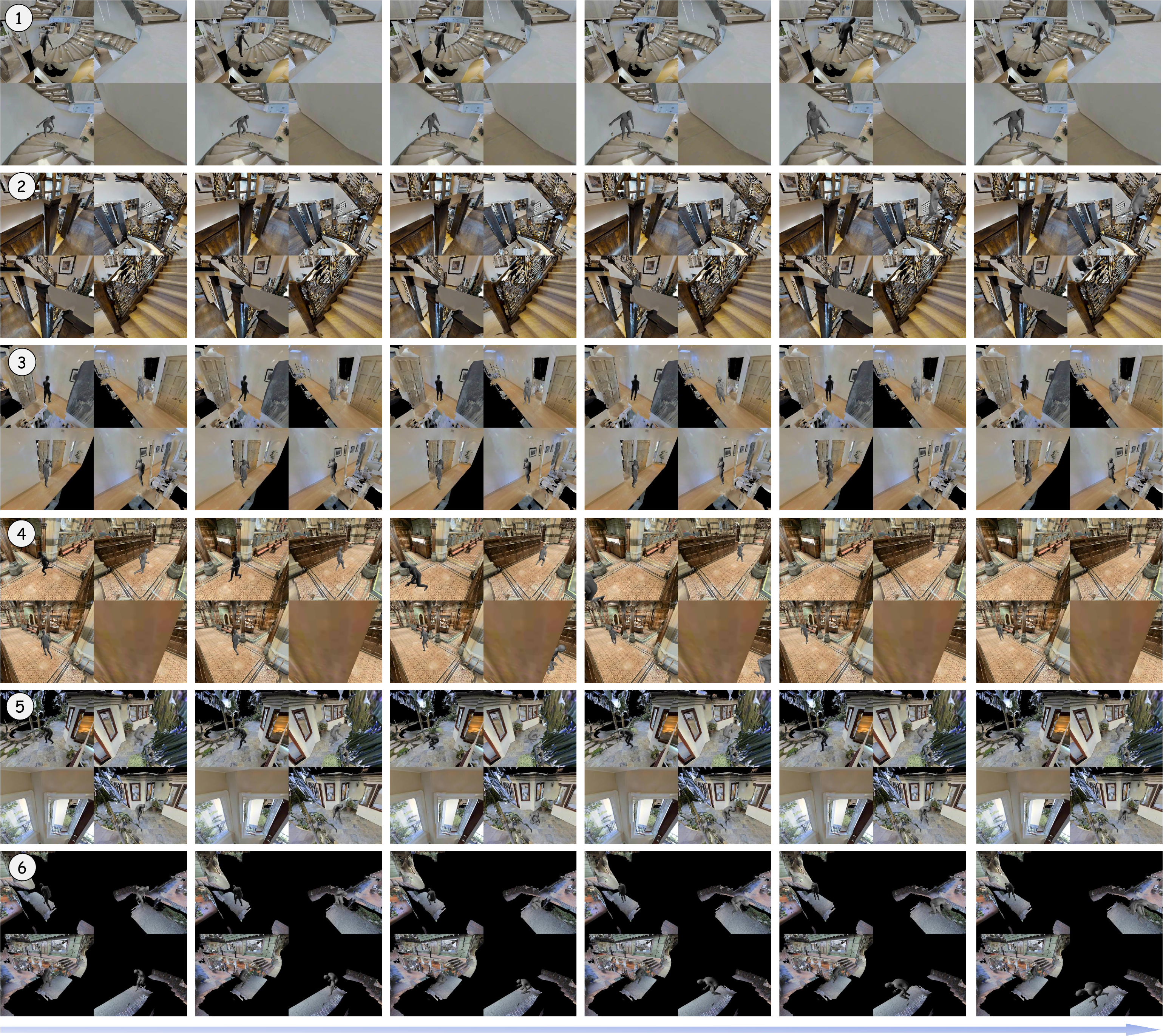}
  \vspace{-0.1in}
  \caption{\small
    \textbf{Movement Examples.}
    We present representative frames from a single set of human motions, each annotated with its corresponding movement. 
    Activities include ascending stairs, running, and pacing. 
    We highlight four camera views (Cameras~2,~4,~6,~8) within multi-camera setup to provide a comprehensive perspective of human behaviors. 
    \textit{(Zoom in for details.)}
  }
  \label{fig:app_move_exp}
  \vspace{-0.3cm}
\end{figure*}

\subsection{Fine~Annotation~With~Multi-Camera}
\label{Appx.coarse2fine-multicamera-setup}
\noindent To refine the coarse placements of human motions, we draw inspiration from 3D skeleton-capture methods~\citep{ji2018large, petrovich2021action} and deploy nine RGB cameras, each positioned around the motion site. As in Fig.~\ref{app:fine_annonation}, this arrangement provides a comprehensive multi-view perspective, revealing potential collisions or misalignments between human figure and surrounding objects.

\noindent \textbf{Camera~Positions~\&~Angles.}
For each camera \(i\) \((i=1,2,\ldots,8)\), we set its 3D location \(\mathbf{p}^{\text{cam}}\) to shift by \(\Delta_x\), \(\Delta_y\), and \(\Delta_z\) from the base position \(\mathbf{p}^{\text{h}}\). Horizontal rotation \(\theta_{\text{lr}}^{i}\) is uniformly spaced at \(\frac{\pi i}{8}\), while vertical rotation \(\theta_{\text{ud}}^{i}\) depends on whether \(i\) is odd or even:
\begin{equation}
\tan \theta_{\text{ud}}^{i} = 
\begin{cases}
0, & \text{if } i \text{ is odd},\\
\frac{\Delta_z}{\sqrt{2}\,\epsilon}, & \text{if } i \text{ is even}.
\end{cases}
\end{equation}
For the ninth camera (overhead view), \(\theta_{\text{lr}}^{9}=0\) and \(\theta_{\text{ud}}^{9}=\frac{\pi}{2}\). These settings are ideal for general views and can be further adjusted in constrained spaces (e.g., narrow closets) or scenes requiring specialized viewpoints.

\subsection{Fine~Annotation~Protocol}
\label{Appx.coarse2fine-protocol}
\noindent We adopt the following six-step procedure to fine-tune a human’s position and orientation:

\begin{enumerate}
[leftmargin=1.1em,itemsep=2pt,topsep=2pt]
    \item \emph{Initial~View.} Generate an overall preview of the human figure at \(\mathbf{p}^{\text{h}}\) (Fig.~\ref{app:fine_annonation}(a)).
    
    \item \emph{Multi-Camera~Observations.}~Collect images from nine cameras~(Figs.~\ref{app:fine_annonation}(b)--(c)).~Adjust camera angles or offsets as necessary, particularly in tight scenes like bathrooms or closets.
    
    \item \emph{Vertical~Collision~Checks.} Inspect overhead Camera~9 to detect vertical overlaps (e.g., arms interpenetrating a table). If collisions exist, identify the nearest side camera to determine how best to shift the figure.
    
    \item \emph{Horizontal~Translation.} Modify \(\Delta_x\) and \(\Delta_y\) accordingly—if a nearby camera (e.g., Camera~1) reveals front-facing overlaps, shift \(\mathbf{p}^{\text{h}}\) by adding or subtracting based on Camera~1’s perspective.
    
    \item \emph{Side~Cameras~Review.} Examine Cameras~2--8 to catch lingering overhang or collisions. Adjust the figure’s position proportionally, typically referencing a standard human height of 1.5\,m to gauge whether shifts remain plausible.
    
    \item \emph{Finalize~Output.} Upon confirming a collision-free layout, automatically generate final video render and corresponding JSON metadata files.
\end{enumerate}

\noindent This multi-camera process systematically eliminates misalignments, ensuring each human model remains properly integrated within the environment. The result is a more realistic portrayal of multi-human interactions and improved fidelity for downstream tasks. The annotators were volunteers recruited through university mailing lists and research group announcements. Participation was entirely voluntary, with no monetary compensation provided. Volunteers contributed time to support research, and their involvement was recognized in accordance with established ethical guidelines.

\begin{figure*}[!t]
    \begin{minipage}[!htbp]{.45\textwidth} 
        \begin{lstlisting}[caption={Format of our LLM prompt for multi-human interactions \& movements enrichment.}, label={list:dataset_prompt}, numbers=none]
"system": "You are an expert in 3D scene understanding, specializing in generating detailed multi-person interaction scenarios in complex environments. I will provide you with information about the current human motion, object positions, and the region context. Your task is to infer possible additional human interactions and describe their actions in relation to the objects and scene."

"example": ...  
//examples in Listing 2

"user": Input:
[human_motion]: \"...\";
[Region]: \"...\";
[Object_1]: \"... \";
[Object_2]: \"...\";
[Object_3]: \"...\";

Based on these inputs, provide a detailed, objective description of additional people in the scene, their locations, and their interactions with objects and each other. Ensure the interactions are plausible based on the given layout and actions.

Output: ...  //generation
        \end{lstlisting}
    \end{minipage}%
    \hspace{.035\textwidth}
    \begin{minipage}[!htbp]{.5\textwidth}  
        \begin{lstlisting}[style=rightstyle, caption={Few-shot example for multi-human interactions \& movements enrichment.}, label={list:few-shot-example}, numbers=none]
# Example 1
Input:
[human_motion]: "tv:An individual enjoying a sports game, cheering loudly.\"
[region]: "\tv.\"
[TV]: "\4 meters in front of the couch, placed on a white media console against the wall.\"
[couch]: "\Against the wall, 4 meters away facing the TV.\"
[table]: "\1.5 meters in front of the couch, holding some snacks and drinks.\"

Output: 
"Three friends are gathered in the living room, watching a movie together. Two of them are seated comfortably on the couch, which is positioned against the wall, facing the TV. They are sharing popcorn and drinks placed on a table 1.5 meters in front of the couch. The atmosphere is relaxed, with one friend fully focused on the movie, casually eating popcorn. Another friend seated next to him. In front of the TV, a third friend stands closer, about 2 meters from the couch, is more animated, loudly cheering as they switch their attention to a sports game playing on a different screen."
        \end{lstlisting}
    \end{minipage}
\end{figure*}

\subsection{Multi-Human~Interaction~\&~Movement~Enrichment}
\label{Appx.human-loop-enrichment}
\noindent To diversify scenes and amplify interactivity, we place additional characters into regions already featuring human motion annotations. This enables more complex interactions and varied motion trajectories. Manual insertion of extra characters, however, is time-consuming and prone to subjective bias, limiting data reliability and diversity.

\noindent \textbf{Human-in-the-Loop~Method.}
We employ LLMs such as ChatGPT-4o and LLaMA-3-8B-Instruct to propose plausible multi-human scenarios. Each prompt integrates details about existing human motions, object positions, and regional context, guiding the LLMs to generate rich, multi-character interactions. Our prompt design uses a \emph{system prompt} and \emph{few-shot examples} (Listings~\ref{list:dataset_prompt} and~\ref{list:few-shot-example}) to ensure clarity and detail. For instance, we collect each human’s position and identify objects within 6\,m, describing relative distances and orientations. The LLMs then construct additional human activities suited to the scene, merging them into cohesive multi-person narratives.

\noindent \textbf{Iterative~Annotation~Workflow.}
After the LLMs produce candidate interactions, we merge outputs from ChatGPT-4o and LLaMA-3-8B-Instruct, then manually refine and validate them over four rounds~\citep{ding2024data, cheng2024shield}. This process corrects inconsistencies and ensures contextual alignment. We subsequently place new human motions according to the generated descriptions, leveraging our multi-camera technique (Sec.~\ref{Appx.coarse2fine-multicamera-setup}) for precise annotation of complex activities (e.g., stair-walking, see Fig.~\ref{fig:app_move_exp}).

\noindent \textbf{Examples~of~Enriched~Interactions.}
Fig.~\ref{fig:simulator} demonstrates how additional humans can populate a living room: \textit{“two people sit on the couch, sharing popcorn on a small table,”} while \textit{“a third friend stands in front of the TV, cheering.”} Such enriched scenes capture realistic multi-human behaviors—from casual gatherings to active cheering—offering agents a broader range of social cues for navigation and interaction.

\begin{algorithm}[!t]
\caption{Real-time Human Rendering in Simulation}
\label{alg:realistic_human_rendering}
\begin{algorithmic}[1]

\REQUIRE Simulation environment $\mathcal{E}$; Human motion data $\mathbf{H}$; Signal queue $\mathcal{Q}$ with maximum size $M \leftarrow 120$; Total frames $N \leftarrow 120$; Frame interval $\Delta t$.

\ENSURE Continuous real-time rendering of $\mathbf{H}$ within $\mathcal{E}$.

\STATE Initialize simulator $\mathcal{E}$, object template manager $\mathcal{T}$ in $\mathcal{E}$, human motion data $\mathbf{H}$ and signal queue $\mathcal{Q}$;
\STATE Initialize total signals sent and processed to $0$;
\STATE \textcolor{blue!40!white}{// \textbf{Thread 1: Signal sender thread}}
    \WHILE{true}
        \IF{not $\mathcal{Q}.\texttt{full}()$}
            \STATE Enqueue signal ``\texttt{REFRESH\_HUMAN}'' into $\mathcal{Q}$;
            \STATE Increment total signals sent;
        \ENDIF
        \STATE Sleep for $\Delta t$;
    \ENDWHILE

\STATE \textcolor{blue!40!white}{// \textbf{Thread 2: Main thread}}
\WHILE{simulation is running}
    \IF{new episode starts}
        \STATE Clear $\mathcal{Q}$ and reset total signals sent to $0$;
        \STATE Remove previous human models from $\mathcal{E}$;
    \ENDIF
   
        \STATE // \textbf{Agent handles signals before observation}
        \WHILE{not $\mathcal{Q}.\texttt{empty}()$}
            \STATE Dequeue signal from $\mathcal{Q}$;
            \STATE $t \leftarrow (\text{total signals processed}) \bmod N$ \COMMENT{Compute current frame index};
            \STATE Remove previous human models from $\mathcal{E}$;
            \FOR{each human motion $h \in \mathbf{H}$}
                \STATE Retrieve motion category, translation, and rotation of $h$ at frame $t$;
                \STATE Load template $\tau_h$ into $\mathcal{T}$;
                \STATE Add human $o_h$ to $\mathcal{E}$ using template $\tau_h$;
                \STATE Set translation and rotation of $o_h$;
            \ENDFOR
            \STATE Increment total signals processed;
        \ENDWHILE
        \STATE Agent observes environment and makes decision;

\ENDWHILE

\end{algorithmic}
\end{algorithm}

\subsection{Real-Time~Human~Rendering}
\label{Appx.real-time-rendering}

We integrate dynamic human models into simulation through a multi-threaded pipeline inspired by \emph{Producer--Consumer} principles and Java-style signaling (Algorithm~\ref{alg:realistic_human_rendering}).~This setup enables agents to observe and respond to human motions in real time, facilitating adaptable navigation policies.

\noindent \textbf{System~Initialization.}~We begin by loading the environment \(\mathcal{E}\), the set of human motions \(\mathbf{H}\), and an object template manager \(\mathcal{T}\) that handles 3D model templates efficiently.

\noindent \textbf{Signal~Sender~Thread (Thread~1).}~At intervals \(\Delta t\), Thread~1 places “refresh” signals into a queue \(\mathcal{Q}\). If \(\mathcal{Q}\) is full, it pauses until earlier signals are processed, preventing data overload. This thread models a continuous stream of human motion updates at a fixed frequency.

\noindent \textbf{Main~Simulation~Thread (Thread~2).}~When the agent is about to act, Thread~2 checks \(\mathcal{Q}\) for pending refresh signals.~It calculates the current frame index \(t\) as \(\text{(signals\_processed} \bmod N)\), where \(N\) is the total length of the human motion sequence. Template manager \(\mathcal{T}\) then removes outdated models and loads frame~\(t\) into the environment, adjusting each figure’s position and orientation.

\noindent \textbf{Synchronization~\&~Consistency.}~We refresh human models immediately before the agent’s perception step, ensuring it observes the latest motion state. Upon starting a new episode, \(\mathcal{Q}\) is cleared, and signal counters reset, so human motions revert to frame~0, maintaining consistency across episodes.~This real-time process keeps human activities synchronized with agent’s action cycle, creating dynamic scenes where agents must adapt to changing bystander locations and behaviors.
\vspace{-0.3cm}


\begin{figure*}[!t]
\setlength{\abovecaptionskip}{3pt}
\setlength{\belowcaptionskip}{3pt}
    \centering
    \includegraphics[width=0.95\linewidth]{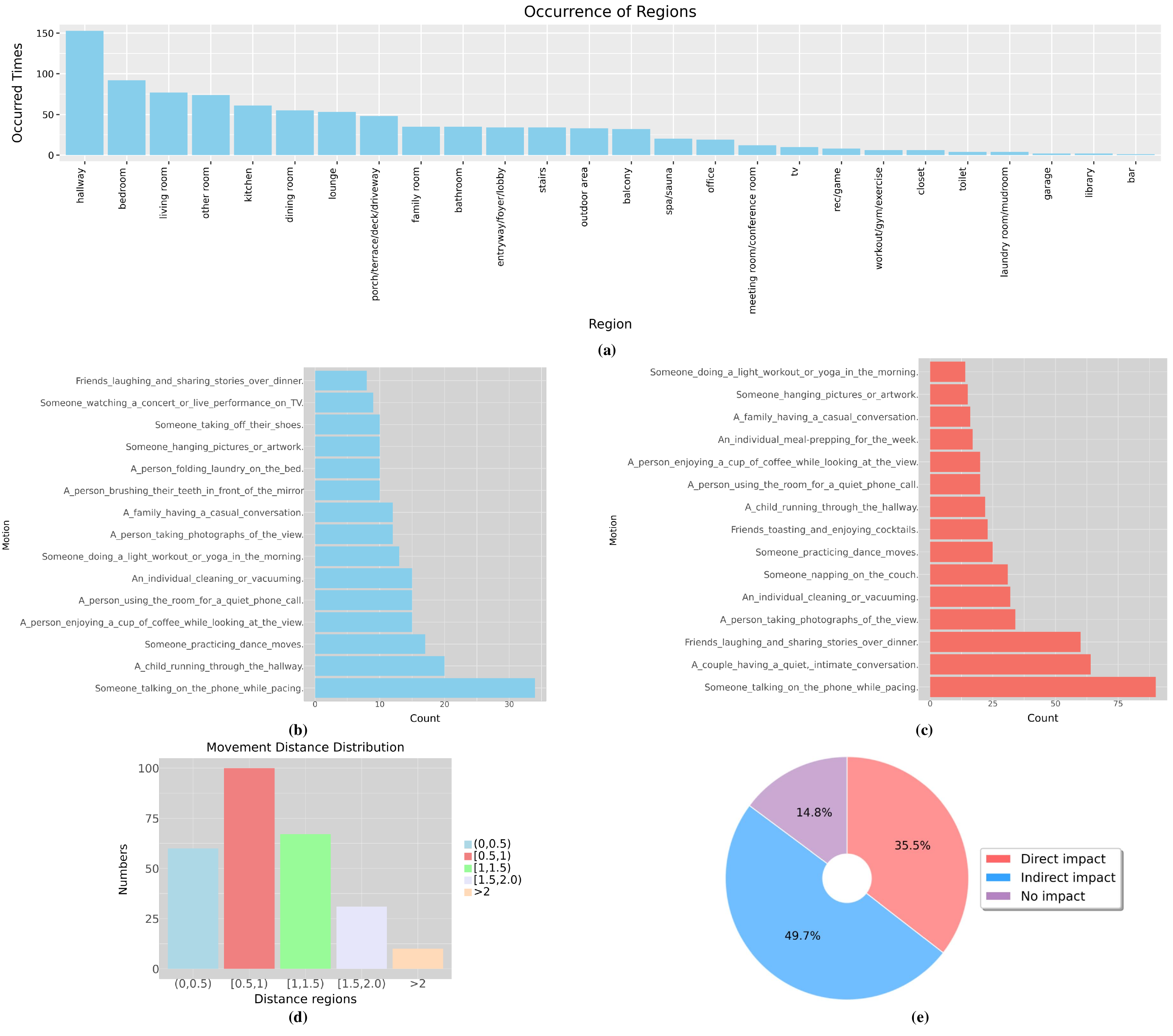}
    \caption{Statistics on human annotations in HA-VLN simulator: (a) Distribution of humans by 26 region types; (b) Top 15 motions without multi-human interaction \& movement enrichment; (c) Top 15 motions with enrichment;  (d) Distribution of human trajectory lengths (in meters); (e) Impact of human presence on environment, categorized as direct, indirect, and no impact. (Zoom in to view)}
    \vspace{-0.3cm}
    \label{fig:app_motion_all}
\end{figure*}

\subsection{API~Design}
\label{Appx.api-design}
\noindent \textbf{Discrete~Environment (DE).}~In our discrete setting, all agent and human positions are tracked via a real-time navigational graph displayed in a 2D top-down view. Each human’s activity is stored as a tuple \(\langle p_h, d_{agent}, \theta_{relative}, a_{status}\rangle\), where \(p_h\) is the human’s 2D coordinate, \(d_{agent}\) is the distance to the agent, \(\theta_{relative}\) is the relative orientation, and \(a_{status}\) indicates activity state. This representation supports efficient, simultaneous tracking of multiple humans in a discrete viewpoint space.  

\noindent \textit{Multi-Entity Detection \& Tracking.}~We employ object detection on each discrete panorama to identify humans, assigning unique IDs for continuous monitoring throughout the navigation process. By linking recognized human poses to specific graph nodes, we anchor their activities to well-defined spatial references.  

\noindent \textit{User~Interface.}~A specialized UI presents a bird’s-eye view of the 2D graph, allowing researchers to visualize, annotate, and adjust human behaviors in real time. This interface significantly streamlines data annotation and analysis for discrete human-aware navigation research.

\noindent \textbf{Continuous~Environment (CE).}~Our API in continuous mode mainly focuses on three components: \emph{(1)~Human~Activity~Monitoring}, \emph{(2)~Environmental~Perception}, and \emph{(3)~Navigation~Support}.

\begin{figure}[t]
\setlength{\abovecaptionskip}{0pt} 
\setlength{\belowcaptionskip}{0pt} 
\centering
\includegraphics[width=0.6\textwidth]{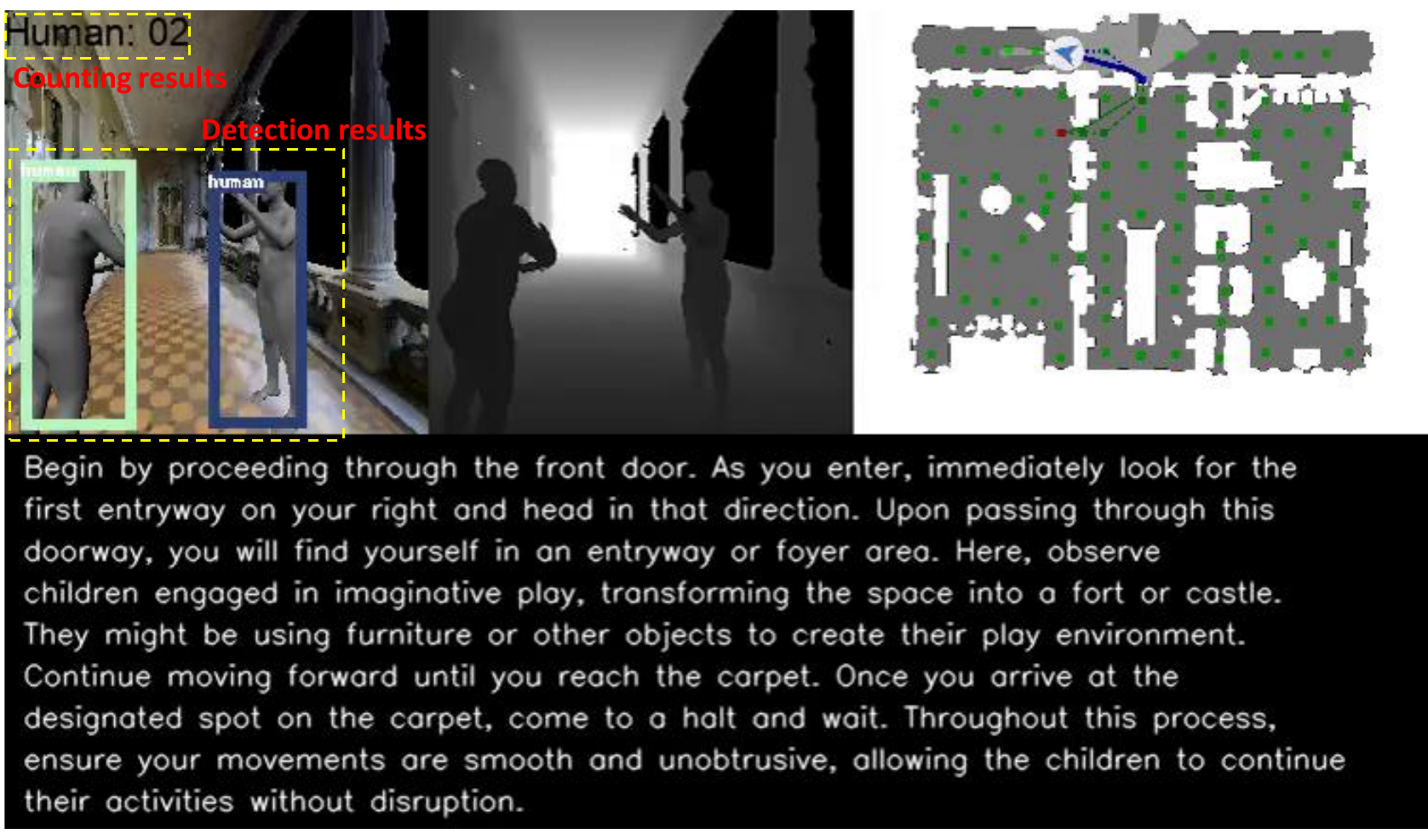}
\caption{The visualization of Human Counting.}
\label{fig:counting}
\vspace{-0.4cm}
\end{figure}

\noindent \textbf{(1)~Human~Activity~Monitoring.}~We track and analyze human activity in real time as in Sec.~\ref{sec:havln-sim}. When collisions occur, the agent reverts to its prior position, and we identify whether the obstacle is human or an inanimate object. For human collisions, we log the coordinates and motion state to inform potential reward-shaping strategies. Distance and orientation estimates derive from agent–human coordinate data. For instance, we employ the Grounding-DINO~\citep{liu2024grounding} detector on RGB inputs with the prompt \emph{“human”} to count individuals.~Fig.~\ref{fig:counting} illustrates how human detection bounding boxes enable real-time counting.

\noindent \textbf{(2)~Environmental~Perception.}~We maintain a dynamic scene graph comprising static elements (e.g., buildings, furniture) and moving humans. The agent continuously updates this graph by fusing positional changes and human motion data in its vicinity. This ensures real-time awareness of human activities for downstream decisions.

\noindent \textbf{(3)~Navigation~Support.}~An A*-based planner computes candidate trajectories while accounting for both dynamic humans and static obstacles. During execution, we monitor any divergence between the agent’s chosen route and the planner’s recommended path. This method highlights human-centric obstacles and informs the agent’s short-term re-planning steps.
Our unified API supports real-time detection, tracking, and socially compliant navigation decisions in both \emph{discrete} and \emph{continuous} modes. It simplifies multi-human scene management, ensures intuitive collision handling, and provides robust path-planning assistance, together forming a foundation for advanced human-aware navigation algorithms.

\subsection{Human~Activities~Annotation~Analysis}
\label{Appx.human-motion-statistics}
\noindent \textbf{Human~Distribution~by~Region.}~Fig.~\ref{fig:app_motion_all}(a) illustrates the distribution of 910 humans across 26 region types in 90 buildings, averaging about nine individuals per building. Even though each person moves independently, this distribution ensures robust and dynamic multi-human interactions, closely mirroring real-world scenarios.

\noindent \textbf{Motion~Frequency~Analysis.}~Figs.~\ref{fig:app_motion_all}(b)--(c) compare the 15 most frequent motions before and after multi-human enrichments. While the total number of motions increases, we also embed additional movement patterns and group interactions into existing actions. For instance, “\emph{talking on the phone while pacing}” may now involve extended pacing distances or layered scenarios like “\emph{a couple having a quiet conversation}” or “\emph{friends sharing stories over dinner}.”

\noindent \textbf{Movement~Distance~Analysis.}~Fig.~\ref{fig:app_motion_all}(d) displays the distribution of trajectory lengths for actively moving humans. Specifically, 22.4\% cover distances up to 0.5\,m, 37.3\% reach 0.5--1\,m, 25.0\% span 1--1.5\,m, 11.6\% extend 1.5--2\,m, and the remaining 3.7\% exceed 2\,m. This wide range reflects the diverse indoor and outdoor behaviors encompassed in the dataset.

\noindent \textbf{Human~Impact~Analysis.}~As shown in Fig.~\ref{fig:app_motion_all}(e), humans exert a notable influence on navigation paths: 35.5\% of the 16,844 paths in HA-VLN physically intersect with human motion, while 49.7\% of viewpoints are indirectly affected (i.e., humans are visible along the route). These statistics underline the importance of accounting for human presence and movement trajectories when designing real-world navigation agents.

All datasets used in this work (HAPS 1.0~\citep{li2024human}, R2R-CE~\citep{krantz2020beyond}) are publicly available research datasets released under appropriate licenses. During dataset construction, we checked that the included visual and textual data do not contain names, faces, or other information that could uniquely identify individuals. Offensive or harmful content was not observed in the sources used. Thus, no anonymization beyond the original dataset release was required.

\begin{table*}[!ht]
\centering
\caption{\small \textbf{Instruction Samples from the HA-R2R Dataset.} Text in \textcolor{mypurple2}{purple} highlights \emph{human-related actions/movements}, 
while text in \textcolor{blue}{blue} indicates explicit \emph{agent-human interaction} cues. These examples illustrate how HA-R2R integrates dynamic human considerations and social awareness into navigation instructions.}
\vspace{-0.1in}
\begin{tabular}{|p{1\textwidth}|}
\hline
1. Exit the library and turn left. As you proceed straight ahead, you will enter the bedroom, \textcolor{mypurple2}{where you can observe a person actively searching for a lost item, perhaps checking under the bed or inside drawers}. Continue moving forward, \textcolor{blue}{ensuring you do not disturb his search}. As you pass by, \textcolor{mypurple2}{you might see a family engaged in a casual conversation on the porch or terrace}, \textcolor{blue}{be careful not to bump into them}. Maintain your course until you reach the closet. Stop just outside the closet and await further instructions. \\
\hline
2. Begin your path on the left side of the dining room, \textcolor{mypurple2}{where a group of friends is gathered around a table, enjoying dinner and exchanging stories with laughter}. As you move across this area, \textcolor{blue}{be cautious not to disturb their gathering}. The dining room features a large table and chairs. Proceed through the doorway that leads out of the dining room. Upon entering the hallway, continue straight and then make a left turn. As you walk down this corridor, you might notice framed pictures along the walls. The sound of laughter and conversation from the dining room may still be audible as you move further away. Continue down the hallway until you reach the entrance of the office. Here, \textcolor{mypurple2}{you will observe a person engaged in taking photographs, likely focusing on capturing the view from a window or an interesting aspect of the room}. Stop at this point, ensuring you are positioned at the entrance without obstructing the photographer's activity. \\
\hline
3. Starting in the living room, \textcolor{mypurple2}{you can observe an individual practicing dance moves, possibly trying out new steps}. As you proceed straight ahead, \textcolor{mypurple2}{you will pass by couches where a couple is engaged in a quiet, intimate conversation, speaking softly to maintain their privacy}. Continue moving forward, ensuring you navigate around any furniture or obstacles in your path. As you transition into the hallway, \textcolor{mypurple2}{notice another couple enjoying a date night at the bar, perhaps sharing drinks and laughter}. \textcolor{blue}{Maintain a steady course without disturbing them}, keeping to the right side of the hallway. Upon reaching the end of your path, you will find yourself back in the living room. Here, \textcolor{mypurple2}{a person is checking their appearance in a hallway mirror, possibly adjusting their attire or hair}. Stop by the right candle mounted on the wall, ensuring you are positioned without blocking any pathways. \\
\hline
4. Begin by leaving the room and turning to your right. Proceed down the hallway, be careful of any human activity or objects along the way. As you continue, look for the first doorway on your right. Enter through this doorway and advance towards the shelves. Once you reach the vicinity of the shelves, come to a halt and wait there. During this movement, avoid any obstacles or disruptions in the environment. \\
\hline
\end{tabular}
\vspace{-0.1in}
\label{appx.tab:har2r-instructions-examples}
\end{table*}

\section{Agent~Details}
\label{Appx.agent_all}

\subsection{HA-R2R~Instruction~Examples}
\label{Appx.har2r-ue-all}
Table~\ref{appx.tab:har2r-instructions-examples} illustrates four sample instructions from the \emph{Human-Aware Room-to-Room} (HA-R2R) dataset. These examples encompass multiple scenarios: multi-human interactions (e.g., 1, 2, 3), direct agent--human encounters (e.g., 1, 2, 3), situations with four or more bystanders (e.g., 3), and paths devoid of humans (e.g., 4). Together, they demonstrate how HA-R2R challenges an agent with diverse human-aligned instructions.

\subsection{HA-R2R~Instruction~Generation}
\label{Appx.har2r-ue-prompt-method}
\noindent To create enriched instructions for HA-R2R, we use ChatGPT-4o and LLaMA-3-8B-Instruct to expand upon R2R-CE’s original textual data. Our strategy involves a carefully crafted few-shot prompt, combining a \emph{system prompt} (Listing~\ref{list3:enrichment_prompt}) and \emph{few-shot examples} (Listing~\ref{list4:enrichment-few-shot-example}).  

\noindent \textbf{Prompt~Structure.}~The system prompt lays out guidelines for generating instructions that emphasize social context. It encourages mentioning human activities and interactions relevant to navigation paths. Few-shot examples then illustrate desired format, including references to human behavior (e.g., “\emph{someone quietly making a phone call; keep your voice down as you proceed}”), positional references, and object interactions.

\noindent \textbf{Iterative~Refinement.}~In early trials, the models sometimes produced extraneous or subjective content, lacking sufficient detail on human activities. We iteratively refined the system prompt and examples, clarifying the need for neutral tone, accuracy, and contextual alignment with human-related scenarios. In each round, we analyzed model outputs, identified discrepancies, and adjusted examples to showcase more detailed, coherent, and socially aware instructions.
This process guided ChatGPT-4o and LLaMA-3-8B-Instruct toward generating instructions that fully integrate human-centric elements such as bystander activities, spatial cues, and subtle behavioral recommendations. The final HA-R2R instructions thus reflect enriched scene descriptions where agents must account for diverse, real-world nuances involving human presence.

\begin{figure*}[!t]
    \begin{minipage}{.53\textwidth}
        \begin{lstlisting}[caption={Format of our LLM prompt for HA-R2R instruction generation.}, label={list3:enrichment_prompt}, numbers=none, xleftmargin=0pt]
"system": "You are tasked with generating an objective and detailed description of a path and the related scene information based on the given input. Focus solely on observable entities and actions in the scene, avoiding descriptions of smell, atmosphere, or subjective feelings. Ensure clarity and precision in object and human activity descriptions."

"examples": ...  //examples in Listing 4 

"user": Input:
[instruction_text]: \"...\";
[start]: \"...\";
[during]: \"...\";
[end]: \"...\";

Please provide an objective, step-by-step description of the path, focusing on the objects and human activities observed at each stage of the navigation (beginning, during, and end). Expand your description by including character actions, objective descriptions of objects, and details of the movement process. Your description should maintain a neutral tone and avoid embellishments. Please simulate the robot's movement along the path, while incorporating possible connections or interactions between the robot, objects, and human activities. 

Output: ...  //generation
        \end{lstlisting}
    \end{minipage}%
    \hspace{.05\textwidth}
    \begin{minipage}{.43\textwidth}
        \begin{lstlisting}[style=rightstyle, caption={Few-shot examples for HA-R2R instruction generation.}, label={list4:enrichment-few-shot-example}, numbers=none, xleftmargin=0pt]
# Example 1:
Input:
[instruction_text]: \"Go through the doorway to the hall.  Go down the hall past the landing.  Take the first door on the right and stop in the doorway.  \";
[start]: \"other room:A person using the room for a quiet phone call. \";
[during]: \"other room:A person using the room for a quiet phone call. \";
[end]: \"hallway:An individual cleaning or vacuuming. \";

Output:
You should start from the room. Before you walk through the doorway into the hallway, you will notice someone quietly making a phone call, so please remain quiet as you move. Then, proceed down the hallway past the landing. As you continue through the corridor, you will see a responsible cleaner using a vacuum to tidy up. Finally, your destination is at the end of the hallway, enter the first door on the right and stop in the doorway.
        \end{lstlisting}
    \end{minipage}
\vspace{-0.1in}
\end{figure*}

\subsection{HA-R2R~Data~Analysis}
\label{Appx.har2r-ue-statistics}
\begin{figure*}
\setlength{\abovecaptionskip}{1pt}
\setlength{\belowcaptionskip}{-1pt}
    \centering
    \includegraphics[width=1\linewidth]{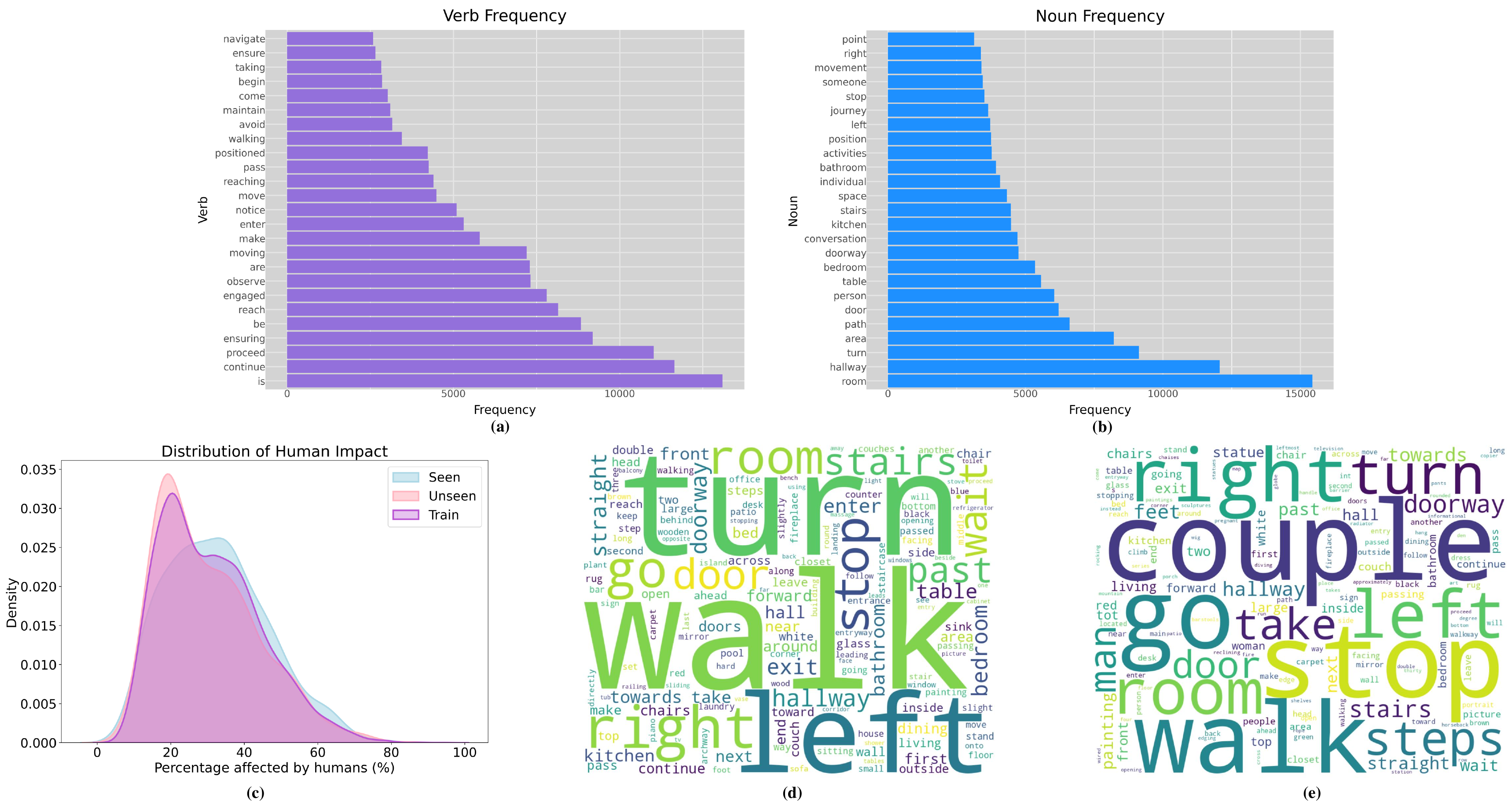}
    \caption{\small
    \textbf{Statistics for HA-R2R Dataset.}
    (a) Verb frequency distribution for all instructions.
    (b) Noun frequency distribution for all instructions.
    (c) Distribution of human impact within HA-R2R.
    (d) Word cloud of instructions not aligned with human activities.
    (e) Word cloud of instructions explicitly involving human actions. 
    Larger font size indicates higher frequency or proportion in the HA-R2R dataset.
    }
    \vspace{-0.1in}
    \label{fig:appx_all_instruction}
\end{figure*}

\noindent \textbf{Word~Frequency~Analysis.}  
We conduct a word frequency study on HA-R2R to gauge its capacity for representing realistic, human-centric scenarios.  
Figs.~\ref{fig:appx_all_instruction}(a)~and~(b) illustrate frequently used nouns and verbs, confirming the dataset’s focus on both spatial navigation and social interactions.

\emph{Nouns.}~The five most common nouns are \emph{room}, \emph{hallway}, \emph{turn}, \emph{area}, and \emph{path}, with \emph{room} alone appearing over 15{,}000 times. Other notable terms (\emph{person}, \emph{doorway}, \emph{kitchen}) highlight spatial complexity and social elements such as \emph{conversation}, \emph{activities}, and \emph{someone}.

\emph{Verbs.}~The five most frequent verbs, \emph{is}, \emph{continue}, \emph{proceed}, \emph{ensuring}, \emph{be}, reveal an action-oriented narrative, while additional terms (\emph{engaged}, \emph{observe}, \emph{notice}, \emph{avoid}, \emph{maintain}) underscore instructions geared toward social awareness and precise route-following.

\noindent \textbf{Human~Impact~Analysis.}  
Fig.~\ref{fig:appx_all_instruction}(c) shows that most instructions contain 20--60\% human-related content, reflecting the dataset’s emphasis on people in everyday scenes. Comparisons of word clouds in Figs.~\ref{fig:appx_all_instruction}(d)~and~(e) confirm that while both human-aligned and non-human segments use common navigational verbs (\emph{walk, left, right}), instructions involving humans introduce additional social context (\emph{couple, man, painting}). This integration of interpersonal cues elevates HA-R2R beyond simple route directives, better mirroring real-world navigation challenges in human-filled environments.

\subsection{Visual~and~Depth~Embeddings}
\label{appx:vd}
Following VLN-CE~\citep{krantz2020beyond}, we employ parallel streams to process RGB and depth images. Each viewpoint produces a set of features from two specialized ResNet-50 models:
\begin{enumerate}
[leftmargin=1.1em,itemsep=2pt,topsep=2pt]
    \item \textbf{RGB~Features.} Let \(\{v^{rgb}_1, v^{rgb}_2, \dots, v^{rgb}_k\}\), where \(v^{rgb}_i \in \mathbb{R}^{2048}\), be outputs of a ResNet-50 pretrained on ImageNet.
    \item \textbf{Depth~Features.} Let \(\{v^d_1, v^d_2, \dots, v^d_k\}\), where \(v^d_i \in \mathbb{R}^{128}\), be outputs of another ResNet-50 pretrained on Gibson-4+~\citep{xia2018gibson} and MP3D for point-goal navigation.
\end{enumerate}

We fuse these two feature streams along with a directional encoding \(d_i\) indicating spatial orientation:
\begin{equation}
v_i 
=\bigl[\;v^{rgb}_i\,W_{\mathrm{rgb}};\;\,v^d_i\,W_{\mathrm{depth}};\;\,d_i\;\bigr]\,W_{\mathrm{merge}},
\end{equation}
where \(W_{\mathrm{rgb}}\), \(W_{\mathrm{depth}}\), and \(W_{\mathrm{merge}}\) are learnable projection matrices with ReLU activation. The directional encoding \(d_i\) is constructed by repeating \((\cos\theta^i_t,\;\sin\theta^i_t)\) 32 times, where \(\theta^i_t\) measures the relative heading offset of the agent.
The fused embedding \(v_i \in \mathbb{R}^d\) is either 512 or 768 dimensions, matching the requirements of our \textbf{HA-VLN-CMA} or \textbf{HA-VLN-VL} agent. Both ResNet backbones remain fixed during training, ensuring consistent and stable representations from the RGB and depth channels throughout the learning process.

\subsection{Text~Embeddings}
\label{appx:te}
For the \textbf{HA-VLN-VL} agent, we utilize text embeddings from \emph{PREVALENT}~\citep{hao2020towards}, which was pre-trained on 6.58M image–text–action triplets, thereby capturing broad contextual cues for navigation. Conversely, the \textbf{HA-VLN-CMA} agent adopts embeddings from BERT~\citep{devlin2018bert}, also widely used for its strong language representations. Formally, let \(\ell = \{w_1,\dots,w_n\}\) be a sequence of tokens representing the instruction. Each token \(w_i\) is mapped to a one-hot vector \(e_i \in \mathbb{R}^V\), where \(V\) is the vocabulary size. An embedding matrix \(E \in \mathbb{R}^{V \times d}\) then projects \(e_i\) into a continuous \(d\)-dimensional space:
\begin{equation}
x_i = E^\top\,e_i,\quad x_i \in \mathbb{R}^d.
\end{equation}
In this manner, each discrete token \(w_i\) is transformed into a trainable embedding \(x_i\), forming the foundation of the model’s linguistic understanding.

\begin{figure*}[t]
\setlength{\abovecaptionskip}{1pt}
\setlength{\belowcaptionskip}{-1pt}
    \centering
    \includegraphics[width=1\linewidth]{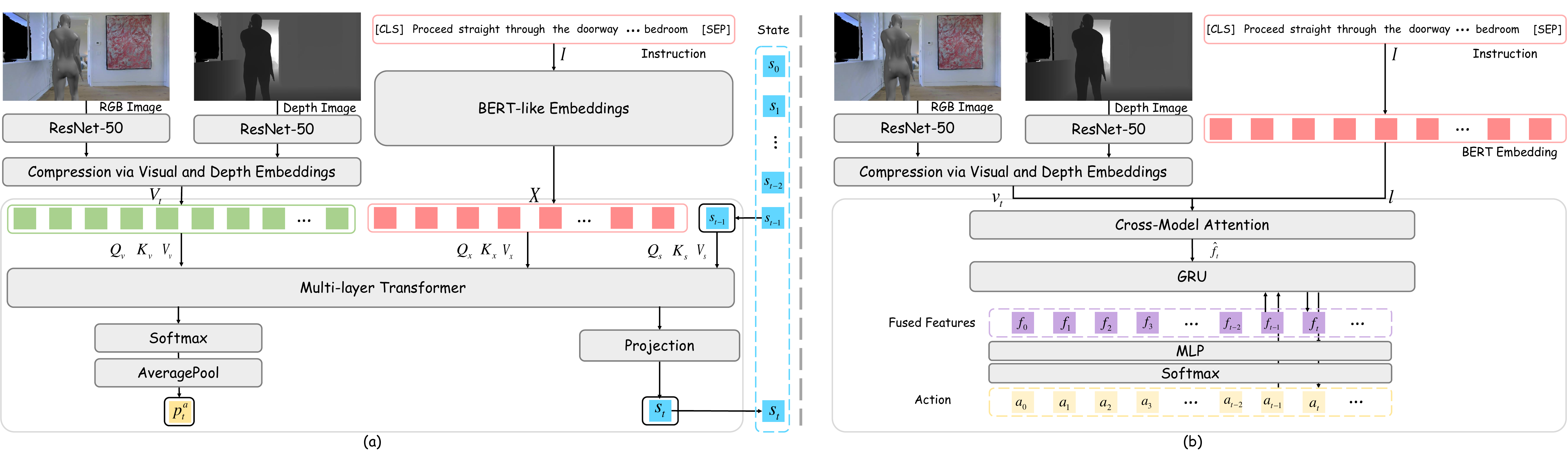}
    \caption{\small
\textbf{Network Structures.}
(a)~\textbf{HA-VLN-VL} adopts a BERT-like transformer with a specialized state token. 
RGB and depth inputs are compressed by ResNet-50 and concatenated, while instruction tokens feed a BERT-like encoder. 
A multi-layer transformer computes cross-modal attention, producing per-step action probabilities via average-pooling and a final projection. 
In both architectures, continuous or discrete commands are then derived for navigation based on the agent’s policy output.
(b)~\textbf{HA-VLN-CMA} employs a cross-modal attention (CMA) module combined with a GRU policy. 
RGB and depth images are first processed by two ResNet-50 encoders and fused into a single feature stream, which attends to the instruction tokens; 
the fused features are then fed into a GRU and MLP to predict actions.
}
    \label{fig:appx_network}
    \vspace{-0.1in}
\end{figure*}

\begin{figure*}[t]
\setlength{\abovecaptionskip}{1pt}
\setlength{\belowcaptionskip}{-1pt}
    \centering
    \includegraphics[width=1\linewidth]{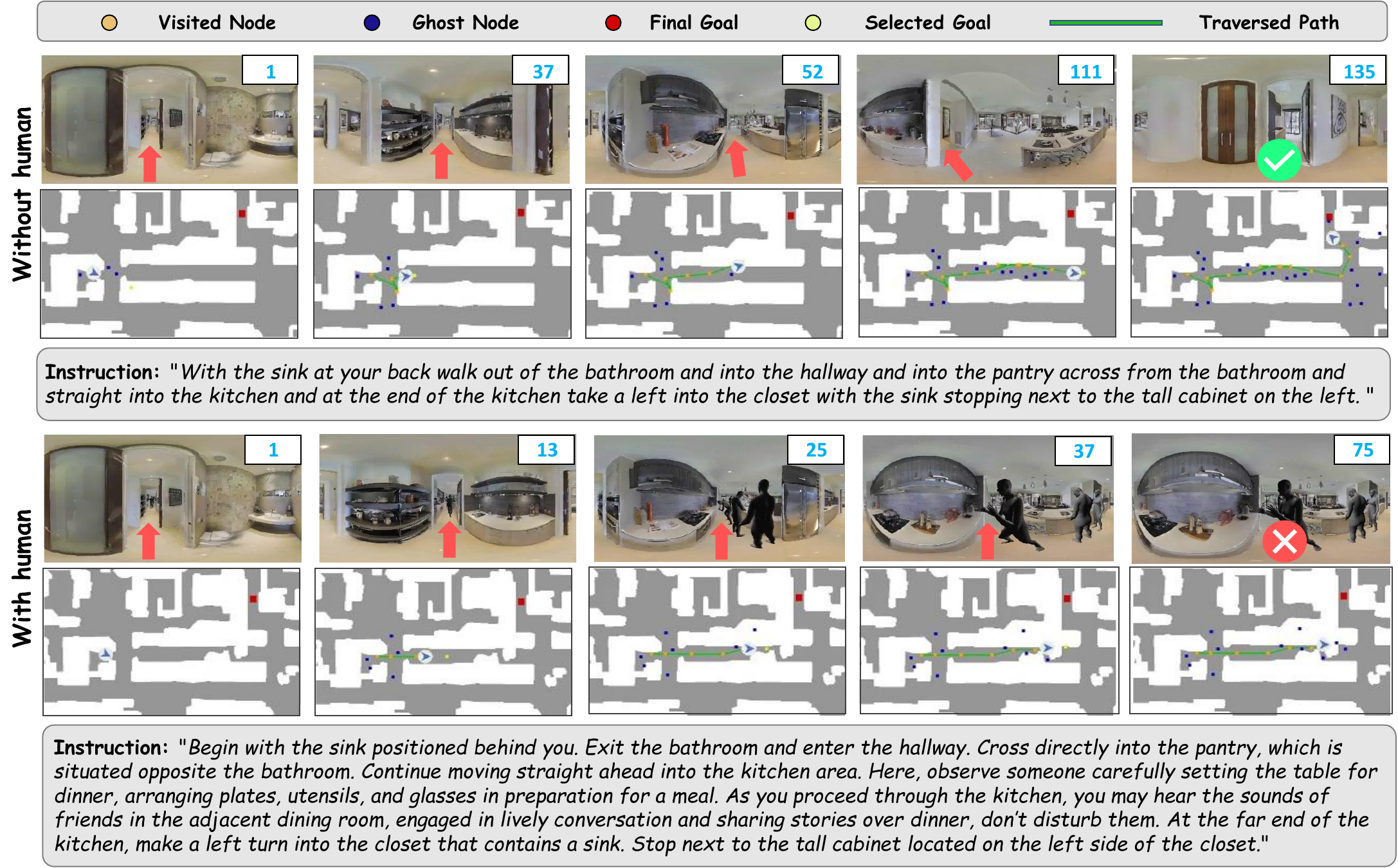}
    \caption{\small
    \textbf{Trajectory Comparison Under Human vs.\ No-Human Conditions.}
    We illustrate the same episode’s trajectories predicted by BEVBert~\citep{an2023bevbert}, trained on VLN-CE, in scenarios with (bottom) and without (top) human presence.  
    In the top row, no bystanders are present, and the agent follows its instructions with minimal collision risk. 
    In the bottom row, bystanders and human-aligned cues lead to altered motion decisions, sometimes creating additional collision challenges.
  }
  \vspace{-0.1in}
    \label{fig:appx_visual_1}
\end{figure*}

\subsection{HA-VLN-VL~Structure}
\label{Appx.agent-vl}

\noindent \textbf{Model~Overview.}  
\textbf{HA-VLN-VL} adopts a BERT-like architecture inspired by Recurrent~VLN-BERT~\citep{Hong_2021_CVPR}, extending it to handle human-aware navigation. At each timestep \(t\), the model receives the previous state \(s_{t-1}\), language tokens \(X\), and fused RGB–depth visual features \(V_t\) (Sec.~\ref{appx:vd}). It outputs an updated state \(s_t\) and an action distribution \(p_t^a\):
\begin{equation}
s_t,\; p_t^a = \mathrm{HA\text{-}VLN\text{-}VL}\bigl(s_{t-1},\,X,\,V_t\bigr).
\end{equation}

\noindent \textbf{State~Token.}  
In line with BERT conventions, the model maintains a \emph{state token} \(s_t\) that encapsulates the agent’s internal context. Initially, \(s_0\) is set to embedding of \texttt{[CLS]} token. At each step, the state token is updated by appending agent’s previously executed action \(a_t\) and projecting resulting vector:
\begin{equation}
s_t = [\,s'_t;\,a_t\,]\;W_s,
\end{equation}
where \(s'_t\) is the final Transformer-layer output, and \(W_s\) is a learnable projection matrix.

\noindent \textbf{Visual~Attention.}  
To decide the next action, we compute attention scores between \(s_t\) and the set of visual tokens \(V_t=\{v_1, v_2,\dots,v_n\}\), covering navigable directions plus a “stop” option:
\begin{equation}
A_{s,v}^t=\mathrm{Softmax}\!\Bigl(\frac{Q_s\,K_v^\top}{\sqrt{d_h}}\Bigr),
\end{equation}
where \(Q_s\) is derived from \(s_t\) and \(K_v\) from \(v_i\in V_t\). The model then aggregates these attention scores via an average-pooling step:
\begin{equation}
p_t^a = \mathrm{\overline{AveragePool}}\bigl(A_{s,v'}^t\bigr),
\end{equation}
yielding an action distribution over possible moves. The agent selects:
\begin{equation}
a_t = \arg\max\bigl(p_t^a\bigr).
\end{equation}

\noindent \textbf{Training~Objective.}  
\textbf{HA-VLN-VL} is optimized through a combination of \emph{supervised imitation learning}—to mimic ground-truth trajectories—and optional \emph{reinforcement learning}, which rewards safe and efficient paths. As depicted in Fig.~\ref{fig:appx_network}(a), the model continuously refines its understanding of language instructions and visual cues, offering robust and socially aware navigation.

\subsection{HA-VLN-CMA~Structure}
\label{Appx.agent_cma}
\noindent \textbf{Architecture~Overview.}
\textbf{HA-VLN-CMA} is a dual-stream visual-language agent featuring \emph{Cross-Modal Attention (CMA)} and a recurrent decoder for navigation in human-populated scenarios (see Fig.~\ref{fig:appx_network}(b)).  
It processes two visual channels—RGB and Depth—alongside language instructions, then outputs an action at each time step.

\noindent \textbf{Dual-Stream~Visual~Encoding.}
Following Sec.~\ref{appx:vd}, each observation \(o_t\) is split into:
\begin{equation}
v^{\mathrm{rgb}}_t = \mathrm{ResNet}^{\mathrm{rgb}}(o_t), 
\quad
v^{\mathrm{d}}_t = \mathrm{ResNet}^{\mathrm{depth}}(o_t),
\end{equation}
where \(\mathrm{ResNet}^{\mathrm{rgb}}\) and \(\mathrm{ResNet}^{\mathrm{depth}}\) are separate backbones for RGB and Depth, respectively. The fused feature representation is
\begin{equation}
v_i = \bigl[v^{\mathrm{rgb}}_i W_{\mathrm{rgb}};\; v^d_i W_{\mathrm{depth}};\; d_i\bigr]\,W_{\mathrm{merge}},
\end{equation}
where \(W_{\mathrm{rgb}}\), \(W_{\mathrm{depth}}\), and \(W_{\mathrm{merge}}\) are projection matrices, and \(d_i\) is a direction encoding (Sec.~\ref{appx:vd}).

\noindent \textbf{Language~Encoder.}~Textual instructions \(\{w_1, \dots, w_T\}\) are transformed into contextual embeddings:
\begin{equation}
l = \mathrm{BERT}(w_1, \dots, w_T),
\end{equation}
where these embeddings capture the semantic structure of the instruction and serve as input to the cross-modal module.

\noindent \textbf{Cross-Modal~Attention~\&~Recurrent~Decoding.}
At time step~\(t\), we attend to the language features using multi-head attention:
\begin{equation}
\hat{f}_t = \mathrm{MultiHeadAttn}(v_t,\;l),
\end{equation}
where \(\mathrm{Attention}(Q,K,V)=\mathrm{softmax}\!\bigl(\frac{QK^\top}{\sqrt{d_k}}\bigr)V.\)
Multi-head attention helps handle lengthy and detailed instructions by learning multiple representations in parallel. Next, we combine the resulting multimodal embeddings with the previous action \(a_{t-1}\) in a GRU-based decoder:
\begin{equation}
f_t = \mathrm{GRU}\bigl(\,[\,(\,v_t,l),\,a_{t-1}\,],\,f_{t-1}\bigr),
\end{equation}
where \(f_{t-1}\) is the previous hidden state. Finally, an MLP outputs the action distribution:
\begin{equation}
a_t = \mathrm{softmax}(\mathrm{MLP}(f_t)),
\end{equation}
where \(\mathrm{MLP}(f_t)=W_a\,f_t+b_a\), and \(a_t\) is sampled from \(P(a_t|f_t)\).

\noindent \textbf{Training~Objectives.}
\textbf{HA-VLN-CMA} is trained end-to-end with a mixture of imitation learning (to mimic GT paths) and reinforcement learning (to encourage collision-free, socially compliant navigation). By learning from both paradigms, the agent refines its ability to balance path efficiency and safe distancing in human-populated environments.

\section{Experiments \& Leaderboard~Details}
\label{Appx.exp-all}

\subsection{Evaluation~Metrics}
\label{Appx.exp-eval-metrics}

We adopt a two-tier evaluation protocol for \emph{HA-VLN 2.0}, measuring both \emph{perception} (human awareness) and \emph{navigation} (task completion). Perception metrics track how effectively the agent detects and responds to dynamic humans, while navigation metrics assess overall performance.

\noindent \textbf{Total~Collision~Rate~(TCR).}  
Given the strong impact of human activities around critical nodes (viewpoints) along the navigation path, we manage dynamic humans to ensure precise measurement. For navigation instance \(i\), let \(A^c_i\) be the set of critical human activities the agent will encounter in ground-truth path (e.g., passing close to moving humans or crowded regions). We define:
\begin{equation}
\mathrm{TCR} = \frac{\sum_{i=1}^L (\,c_i - |\,A^c_i|\,)}{L},
\end{equation}
where \(c_i\) denotes the set of time steps with collisions that counts collisions within 1\,m of a human. TCR quantifies how often collisions occur in human-occupied zones.

\noindent \textbf{Collision~Rate~(CR).}  
CR is the fraction of navigation instances incurring at least one collision, conditioned on the fraction \(\beta\) of instructions influenced by humans:
\begin{equation}
\mathrm{CR} = \frac{\sum_{i=1}^L \min\bigl(c_i - |A^c_i|,\,1\bigr)}{\beta L}.
\end{equation}
Unlike TCR, CR highlights whether a collision occurred at all, offering insight into safety over entire trajectories.

\noindent \textbf{Navigation~Error~(NE).}  
NE is the mean distance between agent’s final position and intended target:
\begin{equation}
\mathrm{NE} = \frac{\sum_{i=1}^L d_i}{L},
\end{equation}
where \(d_i\) is agent–target distance at episode end.

\noindent \textbf{Success~Rate~(SR).}  
SR measures the ratio of episodes completed with zero collisions, and checks if the agent stops sufficiently close to the goal~\citep{anderson2018vision}, we provide the equation for the collision check part here:
\begin{equation}
\mathrm{SR} = \frac{\sum_{i=1}^L \mathbb{I}\bigl(c_i - |A^c_i| = 0\bigr)}{L},
\end{equation}
where \(\mathbb{I}\) is 1 if the agent avoids collisions, and 0 otherwise.

\subsection{Ground~Truth~Path~Annotation}
\label{Appx.exp-sota-havlnce}

In HA-VLN-CE, the agent must reach within 3\,m of target while minimizing collisions. To label ground-truth paths, we use an A*-based heuristic search that identifies the shortest viable route, dynamically re-planning when obstacles block progress.

\subsection{Further Discussion on Step Size}

In Table~\ref{tab:step_size_comb}, a 1.0\,m step was treated as four 0.25\,m sub-steps, and a 2.25\,m step as nine 0.25\,m sub-steps, with collisions checked after each sub-step.  
When evaluated on the val\_unseen split, BEVBert agent fails to navigate effectively with both 1.0\,m and 2.25\,m step sizes (SR drops to zero).

\begin{table}[htbp]
\tiny
  \centering
  \caption{\small  
    \textbf{Impact of Step Size Combination on Navigation.} In this experiment, we treat 1m step as four 0.25m steps, and 2.25m step as nine 0.25m steps. In this case, collisions are detected every 0.25m. We show results for \textbf{BEVBert}~\citep{an2023bevbert} on unseen validation.
  }
  \label{tab:step_size_comb}
  \vspace{-0.1in}
  \resizebox{0.5\linewidth}{!}{%
    \begin{tabular}{lcccc}
      \hline
      \textbf{Step Size} & \textbf{NE}$\downarrow$ & \textbf{TCR}$\downarrow$ & \textbf{CR}$\downarrow$ & \textbf{SR}$\uparrow$ \\
      \hline
      1.00 & 6.85 & 26.97 & 0.94 & 0.004 \\
      2.25 & 8.79 & 112.78 & 0.97 & 0.000 \\
      \hline
    \end{tabular}
  }
\end{table}

\subsection{Visualization~of~Navigation}
\label{Appx.exp-visual-nav}

Figs.~\ref{fig:appx_visual_1} \&~\ref{fig:appx_visual_2} illustrate trajectories predicted by \textbf{BEVBert}~\citep{an2023bevbert} (trained on VLN-CE) and \textbf{HA-VLN-CMA}\(^*\), which showcases success and failure in human-filled or empty environments.

\noindent \textbf{Failures~with~Human~Crossing.}  
In Fig.~\ref{fig:appx_visual_1}, the agent performs well when no bystanders are present. Yet in a human-populated setting, it fails to adjust at step~37 when a volunteer crosses its path, leading to collision.

\noindent \textbf{Collision~vs.~Avoidance.}  
Fig.~\ref{fig:appx_visual_2} similarly shows two scenarios. At step~39 in the top pane, a direct approach by a bystander overwhelms the agent, causing a collision. In the bottom pane at step~22, the agent successfully deviates upon sensing a person nearby, avoiding any collision altogether.
These visualizations confirm that dynamic human presence greatly complicates navigation, highlighting the need for robust social-aware models.


\mbox{}
\begin{figure}[!t]
\centering
\includegraphics[width=\linewidth]{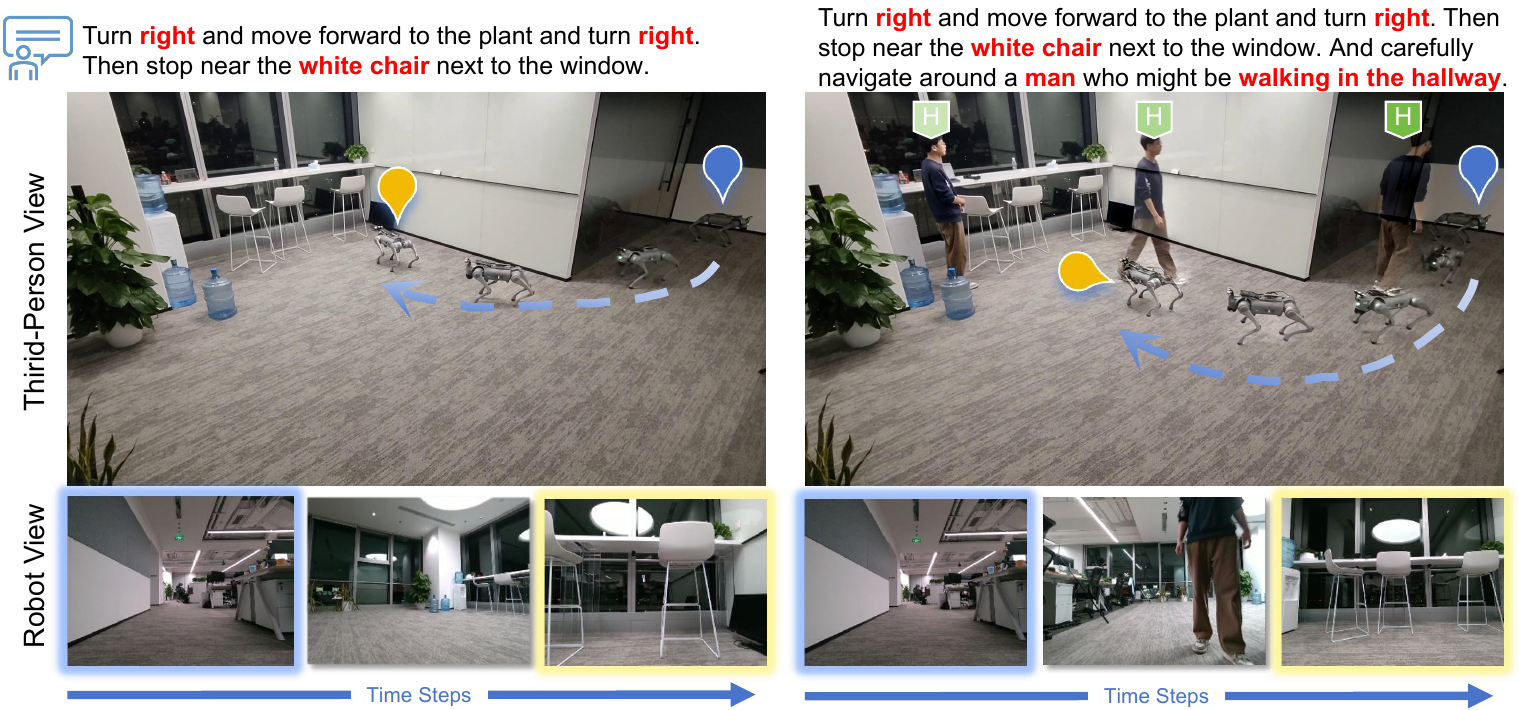} 
\caption{\small
    \textbf{Navigation success in an office} (\emph{left}: no humans, \emph{right}: with humans).
    \emph{Top}: The given instruction for the robot.
    \emph{Middle}: A third-person view of the robot’s path.
    \emph{Bottom}: The robot’s selected view.
}
\label{fig:real_val_office_success}
\end{figure}
\mbox{}
\begin{figure}[!t]
\centering
\includegraphics[width=0.6\linewidth]{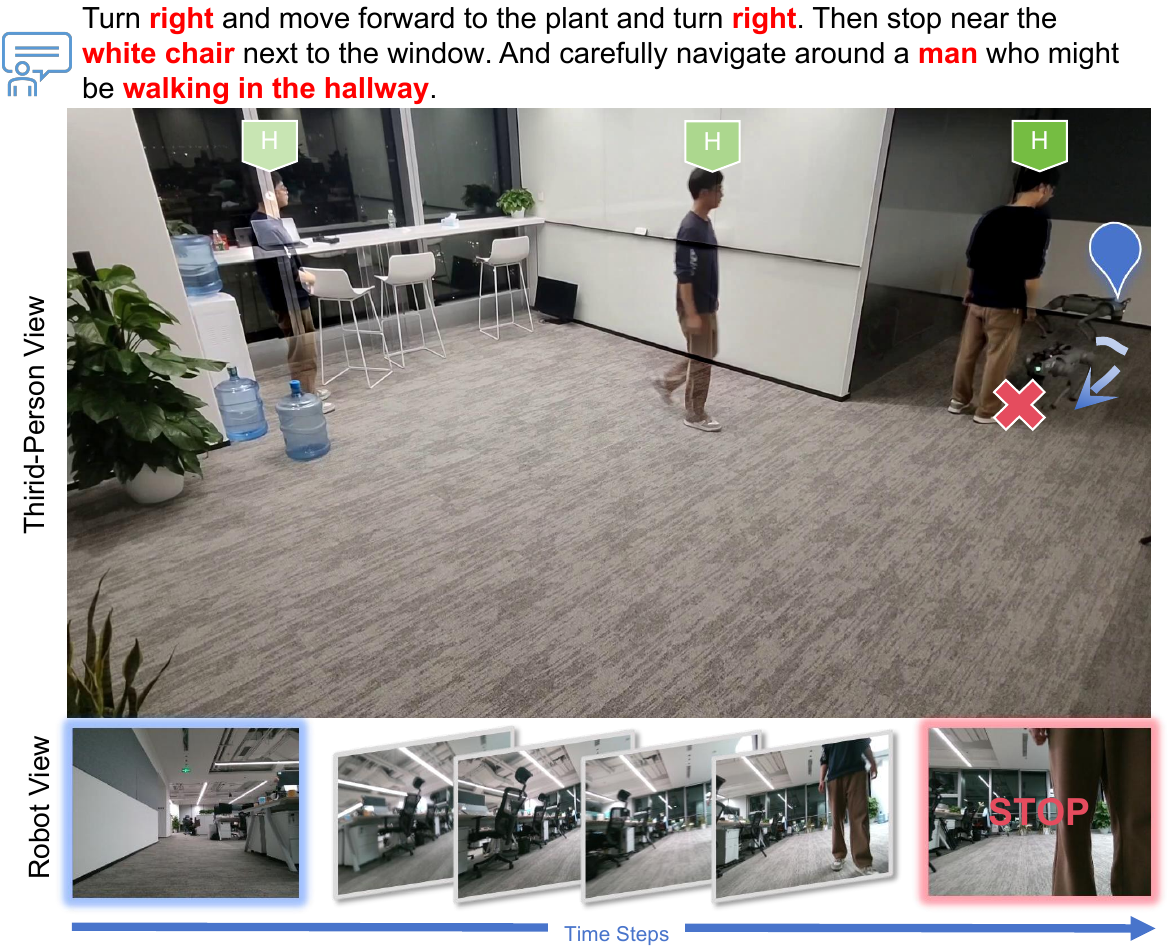}
\caption{\small
    \textbf{Navigation failure in an office setting.}
    A volunteer abruptly changes position, causing robot to collide mid-path.
    This highlights the difficulty of adapting to sudden human movement in confined workspaces.
}
\label{fig:real_val_office_fail}
\end{figure}
\mbox{}
\begin{figure}[!t]
\centering
\includegraphics[width=\linewidth]{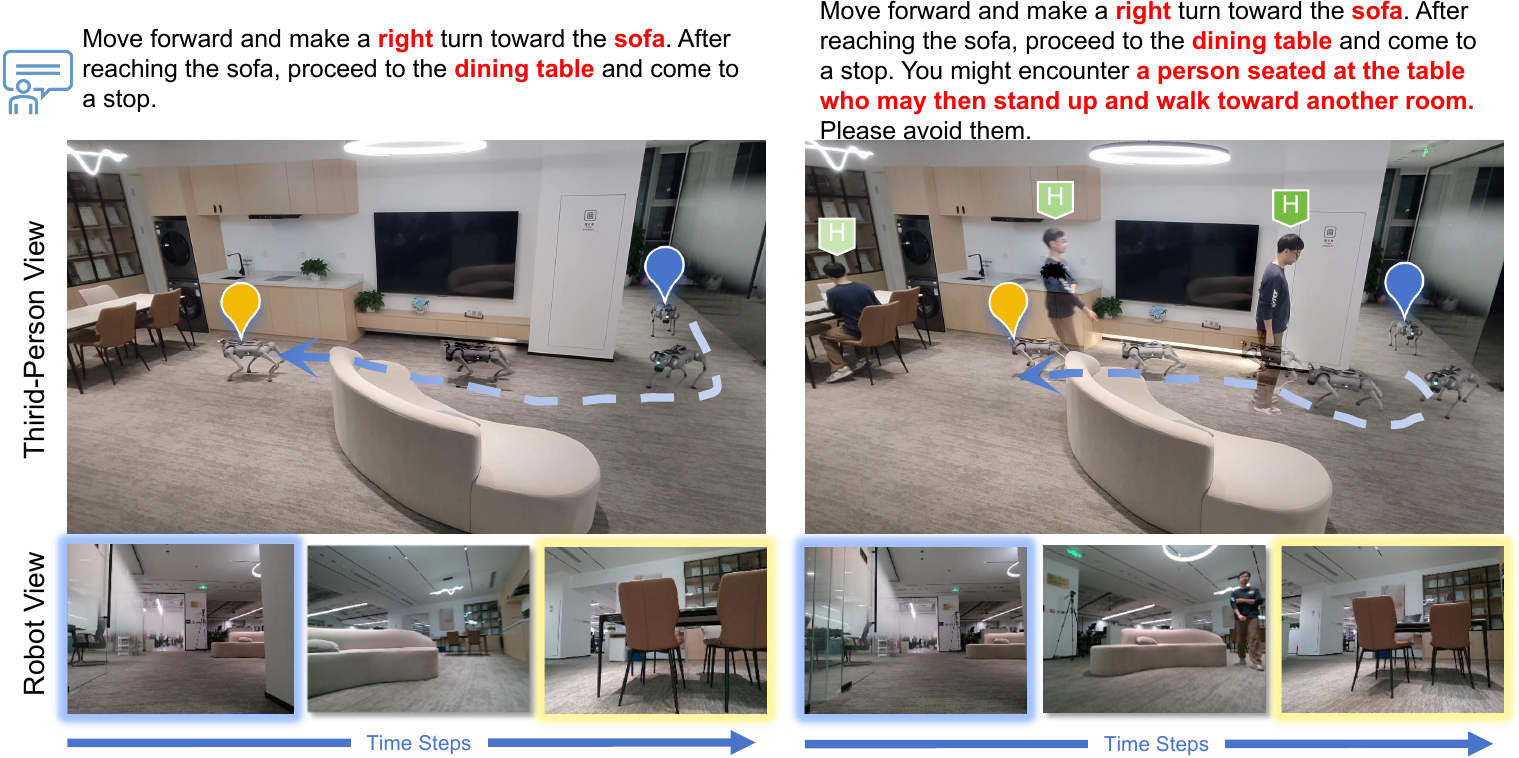}
\caption{\small
    \textbf{Navigation success in a living room} (\emph{left}: no bystanders, \emph{right}: with bystanders).
    The robot follows instructions toward the sofa and dining area,
    keeping safe distances while navigating around volunteers.
}
\label{fig:real_val_livingroom_success}
\end{figure}
\mbox{}
\begin{figure}[!t]
\setlength{\abovecaptionskip}{-1pt}
\setlength{\belowcaptionskip}{-1pt}
\centering
\includegraphics[width=0.6\linewidth]{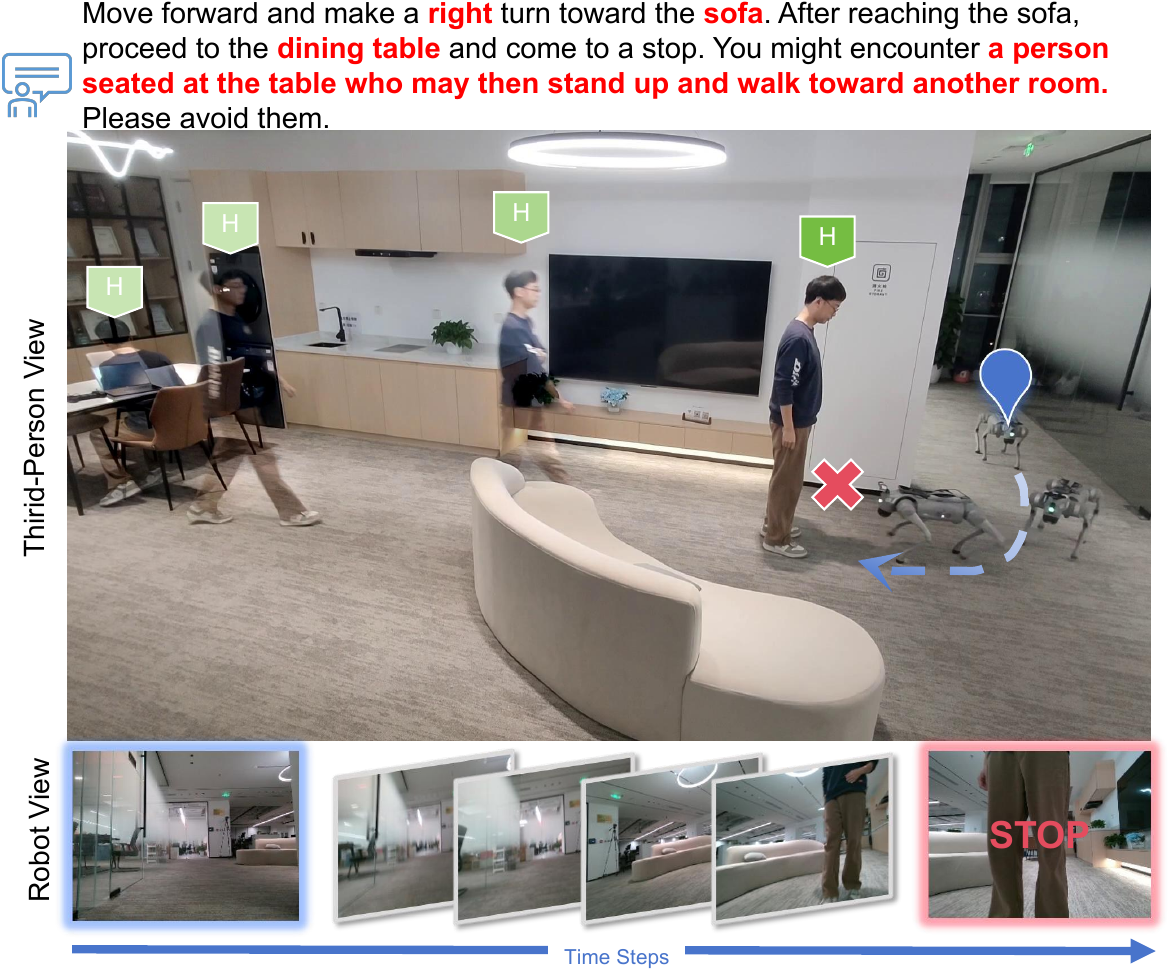}
\caption{\small
    \textbf{Navigation failure in a living room with multiple bystanders.}
    Attempting to move beyond sofa toward a dining area, the robot collides with a volunteer who abruptly stands and shifts position. This underscores how unpredictable human motion
    can disrupt agent’s intended path, requiring rapid re-planning.
}
\label{fig:real_val_livingroom_fail}
\end{figure}
\mbox{}
\begin{figure}[!t]
\setlength{\abovecaptionskip}{2pt}
\setlength{\belowcaptionskip}{2pt}
\centering
\includegraphics[width=\linewidth]{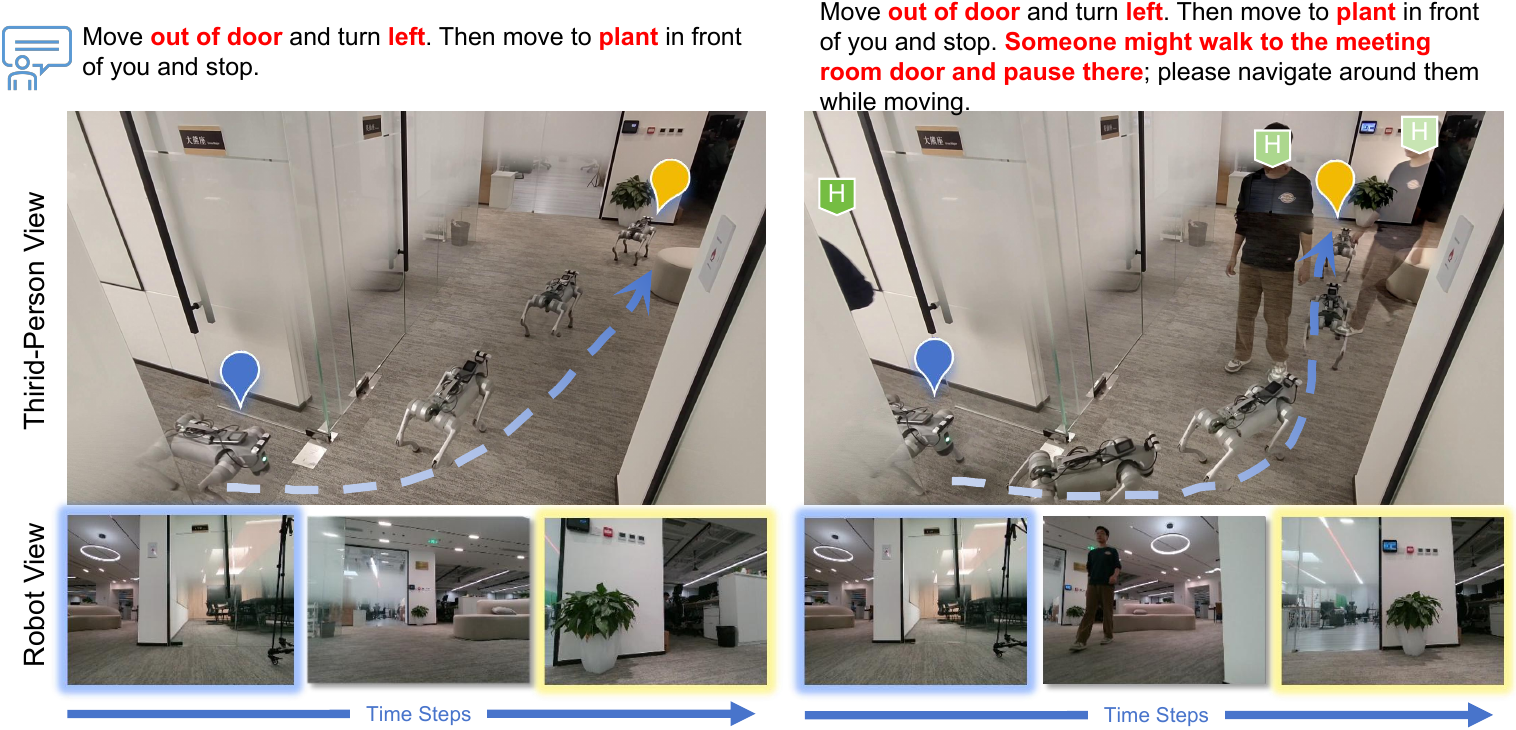}
\caption{\small
    \textbf{Navigation success in a hallway} (\emph{left}: no bystanders, \emph{right}: with bystanders).
    When volunteers appear, the robot halts or deviates to avoid collisions,
    showcasing adaptive behavior in a constrained corridor.
}
\label{fig:real_val_hallway_success}
\end{figure}
\mbox{}
\begin{figure}[!t]
\setlength{\abovecaptionskip}{-1pt}
\setlength{\belowcaptionskip}{-1pt}
\centering
\includegraphics[width=0.6\linewidth]{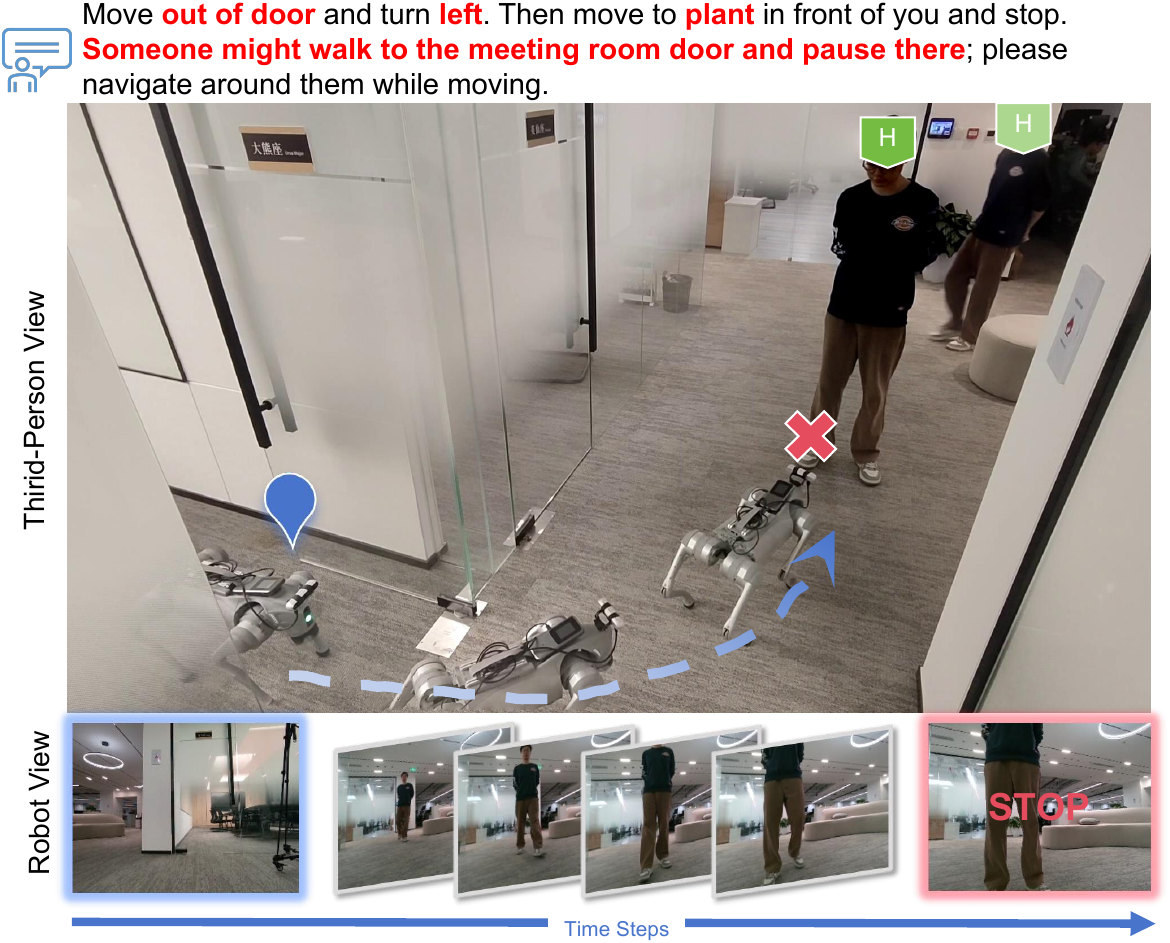}
\caption{\small
    \textbf{Navigation failure in a hallway.}~A volunteer’s sudden positional change causes a mid-path collision and mission failure,
    reflecting the challenge of unpredictable human movement, even in comparatively open corridors.
}
\label{fig:real_val_hallway_fail}
\end{figure}
\mbox{}

\subsection{Implementation Clarification}

Agents in Table~\ref{tab:comparision_a} employ different observation modalities: HA-VLN-CMA, HA-VLN-VL, BEVBert, and ETPNav receive panoramic RGBD observations at each location, while NaVid~\citep{zhang2024navid} and Navila~\citep{cheng2025navila} receive frontal RGB observations. Due to architectural differences, we report only zero-shot performance for NaVid and Navila. Specifically, these VLM-based methods were originally trained on R2R-CE alongside large-scale auxiliary datasets for cross-dataset generalization. Retraining them from scratch solely on our HA-R2R would discard their pretrained visual-language representations and significantly degrade performance, as they rely on extensive multimodal pretraining to align language instructions with visual observations. Therefore, we evaluate their zero-shot transfer capabilities to fairly assess their generalization to our task setting.

\subsection{Leaderboard Tools}

The test partition of HA-R2R contains 3,408 instructions across 18 withheld buildings and intentionally emphasizes multi-human routes. To assess performance on this challenging test split, we host leaderboards for HA-R2R-DE and HA-R2R-CE benchmarks, evaluating both collision-related metrics (TCR, CR) and navigation metrics (NE, SR). We prepare an interactive interface shown in Fig.~\ref{fig:interactive interface plus real world exp} (a), where participants can explore the simulator from nine different views to examine all the annotated human motions and the surrounding environments. This allows them to gain a deeper understanding of the challenging dynamic scenarios we provide. Submissions may include agent code or trajectories, providing reproducible, server-side evaluations and setting a new benchmark for human-centric, dynamic VLN research.

\subsection{Validation~on~Real-World~Robots}
\label{Appx.exp-real-world}
To deploy our navigation agents on physical hardware, the robot is equipped with an \emph{NVIDIA Jetson NX} for AI inference and a \emph{Raspberry Pi~4B} for motion control. The Jetson handles core navigation computations (receiving camera images and inferring action commands), while the Pi executes high-level movement directives such as \emph{turn left} or \emph{move forward}. We set a minimum step size of 0.25\,m and a rotation increment of 15\,degrees. An onboard IMU continuously monitors the robot’s orientation and position, ensuring movement commands align with issued directives.

\begin{table}[!t]

  \centering
\caption{\textbf{Navigation success rate across different region layouts} with (w/) and without (w/o) human presence. Each result is averaged over 30 episodes across 3 instances of each region type.}
  \label{tab:nav_performance}
  \vspace{-0.1in}
  \renewcommand{\arraystretch}{1.1} 
\resizebox{\linewidth}{!}{
\begin{tabular}{lcccccccccc}
\hline
& \multicolumn{2}{c}{\textbf{Living Room}} & \multicolumn{2}{c}{\textbf{Office}} & \multicolumn{2}{c}{\textbf{Hallway}} & \multicolumn{2}{c}{\textbf{Lobby}} & \multicolumn{2}{c}{\textbf{ALL}} \\
\cmidrule(lr){2-3} \cmidrule(lr){4-5} \cmidrule(lr){6-7} \cmidrule(lr){8-9} \cmidrule(lr){10-11}
\textbf{Methods} & w/o & w/ & w/o & w/ & w/o & w/ & w/o & w/ & w/o & w/ \\
\hline
HA-VLN-CMA-Base (trained on VLN-CE) & 0.23 & 0.08 & 0.26 & 0.08 & 0.30 & 0.13 & 0.24 & 0.07 & 0.26 & 0.09 \\
HA-VLN-VL (trained on VLN-CE) & 0.38 & 0.11 & 0.38 & 0.10 & 0.47 & 0.17 & 0.38 & 0.10 & 0.40 & 0.12 \\
HA-VLN-CMA-Base (trained on HA-VLN) & 0.24 & 0.13 & 0.24 & 0.13 & 0.29 & 0.20 & 0.23 & 0.13 & 0.25 & 0.15 \\
HA-VLN-VL (trained on HA-VLN) & 0.42 & 0.17 & 0.43 & 0.17 & \textbf{0.49} & \textbf{0.20} & 0.43 & 0.17 & 0.44 & 0.18 \\
\hline
\end{tabular}}
\vspace{-0.15in}
\end{table} 

\noindent \textbf{Setup.}~Our evaluations use a \emph{Unitree~GO2-EDU} quadruped, featuring the \emph{Intel~Realsense~D435i} camera providing RGB imagery and a \emph{3D~LiDAR} below camera for detection.~IMU refines positional and orientational control, enabling consistent motions. The quadruped rotates to get the panoramic view at each step. We evaluate our agents in four types of everyday indoor environments (each with three instances), including \emph{office}, \emph{living~room}, \emph{hallway}, and \emph{lobby}, under two conditions: \emph{(i)}~w/o human presence (no bystanders) and \emph{(ii)}~w/ human presence (2-4 free-moving volunteers).~This setup simulates realistic indoor traffic patterns and partial observability.

\noindent \textbf{Observations.}~As illustrated in Fig.~\ref{fig:interactive interface plus real world exp}, the robot frequently pauses or yields to avoid oncoming pedestrians. In the absence of bystanders, it navigates smoothly (Fig.~\ref{fig:real_val_office_success}), but collisions arise in cramped corridors or when crowds converge suddenly (Fig.~\ref{fig:real_val_office_fail}). We observe similar patterns in living-room environments (Figs.~\ref{fig:real_val_livingroom_success}--\ref{fig:real_val_livingroom_fail}) and hallways (Fig.~\ref{fig:real_val_hallway_success}).  

Table~\ref{tab:nav_performance} shows the average \textbf{NSR} (Navigation Success Rate) across 30 trials in each instance. While human presence invariably lowers \textbf{NSR}, HA-VLN-VL consistently outperforms HA-VLN-CMA-Base, demonstrating stronger adaptability to dynamic motion. Also, Table~\ref{tab:nav_performance} shows agents trained on HA-VLN achieve higher \textbf{NSR} (0.18 vs.~0.12) than VLN-CE, demonstrating HA-R2R’s sim-to-real gain under realistic conditions.~Still, partial observability and abrupt group formations remain challenging, especially in narrow passages or at congested junctions. Here further details performance under varying crowd densities. 

\noindent \textbf{Visual~Demonstrations.}~Figs.~\ref{fig:real_val_office_success}, \ref{fig:real_val_livingroom_success}, and \ref{fig:real_val_hallway_success} show the robot traversing distinct indoor environments, offices, living rooms, and hallways, guided by natural-language instructions. In Fig.~\ref{fig:interactive interface plus real world exp}, the robot navigates around multiple people, leveraging camera inputs to avoid collisions through minor path adjustments.
Although the agent typically succeeds in reaching its destination, collisions remain possible when bystanders change their trajectories unexpectedly. Figs.~\ref{fig:real_val_office_fail}, \ref{fig:real_val_livingroom_fail}, and \ref{fig:real_val_hallway_fail} illustrate such scenarios, highlighting real-time challenges in unpredictable, human-inhabited spaces.~More demos on our project webpage, further illustrate robot’s performance and underscore how human-aware training aids sim-to-real transfer in dynamic indoor environments.

\noindent \textbf{Insights.}~These~experiments~confirm that simulation-trained, multi-human navigation policies can indeed transfer to physical robots.~However, further refinement in collision forecasting and reactive control is needed to handle unpredictable human behavior in tight indoor settings.

\section{Ethics Statement}

In this study, no human subjects or animal experimentation was involved. All datasets used, including HAPS 2.0, HA-R2R, HA-VLN simulator annotation data, were sourced in compliance with relevant usage guidelines, ensuring no violation of privacy. We have taken care to avoid any biases or discriminatory outcomes in our research process. No personally identifiable information was used, and no experiments were conducted that could raise privacy or security concerns.~We are committed to maintaining transparency and integrity throughout the research process.

\section{Use of LLMs}
Large Language Models (LLMs) were only used to aid in sentence rephrasing and grammar checking of the manuscript.